\documentclass[10pt,twocolumn,letterpaper]{article}

\usepackage[pagenumbers]{iccv} 

\usepackage{graphicx}
\usepackage{booktabs}
\usepackage{multirow}
\usepackage{pifont}
\usepackage{caption}
\usepackage{xcolor}
\usepackage{colortbl}
\usepackage[percent]{overpic}

\usepackage[accsupp]{axessibility}  

\definecolor{firstcolor}{HTML}{BDE6CD}
\definecolor{secondcolor}{HTML}{E2EEBC}
\definecolor{thirdcolor}{HTML}{FFF8C5}

\definecolor{LightYellow}{rgb}{1,1,0.7}

\newcommand{\fst}[1]{\cellcolor{firstcolor}\bfseries #1}
\newcommand{\snd}[1]{\cellcolor{secondcolor} #1}
\newcommand{\trd}[1]{\cellcolor{thirdcolor}#1}

\usepackage[table,xcdraw]{xcolor}
\usepackage{arydshln}
\usepackage{xcolor}

\definecolor{First}{HTML}{BDE6CD}
\definecolor{Second}{HTML}{E2EEBC}
\definecolor{Third}{HTML}{FFF8C5}

\newcommand{\net}{FlowSeek}

\newcommand{\bR}{\mathbf{R}}

\newcommand{\bT}{\mathbf{T}}

\makeatletter
\DeclareRobustCommand\onedot{\futurelet\@let@token\@onedot}
\def\@onedot{\ifx\@let@token.\else.\null\fi\xspace}
\def\eg{e.g\onedot}

\makeatother

\newcommand{\xdownarrow}[1]{%
  {\left\downarrow\vbox to #1{}\right.\kern-\nulldelimiterspace}
}

\newcommand{\xuparrow}[1]{%
  {\left\uparrow\vbox to #1{}\right.\kern-\nulldelimiterspace}
}

\definecolor{iccvblue}{rgb}{0.21,0.49,0.74}
\usepackage[pagebackref,breaklinks,colorlinks,allcolors=iccvblue]{hyperref}
\usepackage{graphicx}
\usepackage[export]{adjustbox}

\title{\net: Optical Flow Made Easier with Depth Foundation Models \\ and Motion Bases}

\author{Matteo Poggi$^{1,2}$\\
\small $^{1}$Department of Computer Science and Engineering (DISI)\\
\and
Fabio Tosi$^{1}$\\
\small $^{2}$Advanced Research Center on Electronic System (ARCES)\\
}

\author{Matteo Poggi \hspace{2cm} Fabio Tosi\\
University of Bologna, Italy \\
\texttt{Project page:} \url{https://flowseek25.github.io/} \\
}

\begin{document}

\twocolumn[{
\renewcommand\twocolumn[1][]{#1}
\maketitle
\begin{center} 
    \vspace{-0.6cm}
    \vspace{0.2cm}
    \centering
    \begin{overpic}[trim=0cm 0cm 23.2cm 0cm, clip, width=\linewidth]{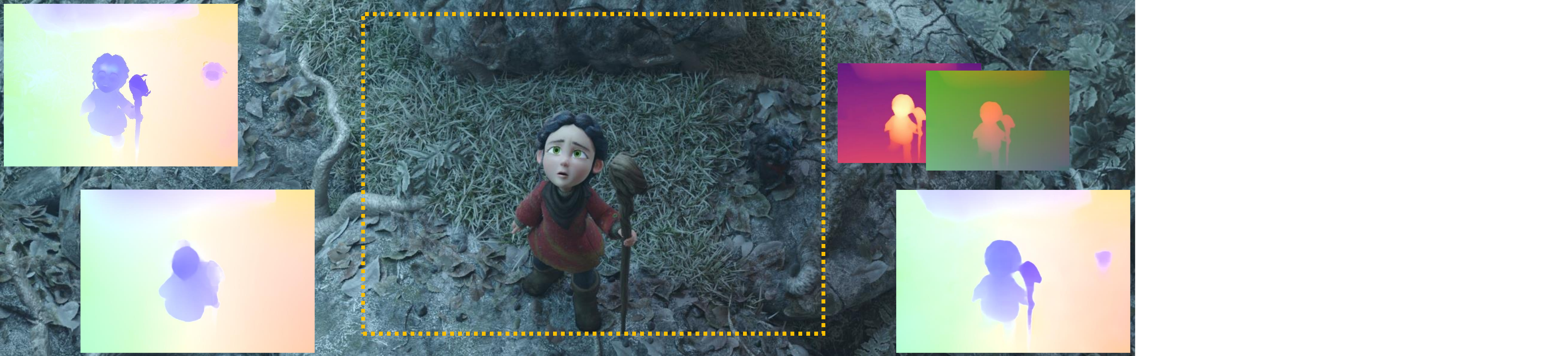}
    \put (1,29) {\textcolor{purple}{\textbf{\texttt{Ground-Truth}}}}
    \put (8,13) {\textcolor{purple}{\textbf{\texttt{SEA-RAFT (L)}}}}
    \put (80,13) {\textcolor{purple}{\textbf{\texttt{\net{} (L)}}}}
    \put (74,26) {\textcolor{yellow}{\footnotesize\textbf{\texttt{Foundation Model Priors}}}}
    \end{overpic}\vspace{-0.3cm}
    \captionof{figure}{\textbf{\net{} in Action.} State-of-the-art optical flow models struggle at generalizing across different domains, with a lack of fine details in their predictions. \net{} achieves superior generalization by exploiting the strong priors from depth foundation models.} 
    \label{fig:teaser}
\end{center}
}
]

\begin{abstract}
We present FlowSeek, a novel framework for optical flow requiring minimal hardware resources for training. FlowSeek marries the latest advances on the design space of optical flow networks with cutting-edge single-image depth foundation models and classical low-dimensional motion parametrization, implementing a compact, yet accurate architecture. FlowSeek is trained on a single consumer-grade GPU, a hardware budget about 8× lower compared to most recent methods, and still achieves superior cross-dataset generalization on Sintel Final and KITTI, with a relative improvement of 10 and 15\% over the previous state-of-the-art SEA-RAFT, as well as on Spring and LayeredFlow datasets.  
\end{abstract}
    
\section{Introduction}
\label{sec:intro}

Optical flow \cite{horn1981determining} is one of the classical problems in computer vision, reaching almost half a century of history. It consists of estimating the 2D motion fields connecting pixels across two or multiple frames in videos, and is the foundation of several higher-level tasks such as action recognition \cite{sun2018optical}, video interpolation \cite{xu2019quadratic, liu2020video, huang2020rife}, or 4D synthesis and reconstruction \cite{Wu_2024_CVPR,TiNeuVox}. In the last decade, this field has been reshaped by the advent of deep learning \cite{dosovitskiy2015flownet}, which has pushed the research community towards the development of end-to-end deep architectures \cite{ilg2017flownet2,ranjan2017optical,sun2018pwc,sun2019models,hui18liteflownet,hui20liteflownet2,hui20liteflownet3,teed2020raft} to replace the hand-crafted algorithms \cite{brox2004high,brox2009large,sun2010secrets,chen2016full,leordeanu2013locally,revaud2015epicflow,li2016fast,hu2016efficient,Hu_2017_CVPR} developed before. Throughout the years, several design strategies have been proposed, including coarse-to-fine architectures \cite{ranjan2017optical,sun2018pwc,hui18liteflownet,hui20liteflownet2,hui20liteflownet3}, 4D convolutions \cite{yang2019volumetric,wang2020displacement} or recurrent neural networks \cite{teed2020raft}, with the latter becoming the dominant \cite{ sui2022craft, huang2022flowformer, shi2023flowformer++, zhao2022global, lu2023transflow, wang2024sea} following RAFT \cite{teed2020raft}.

Associated with architectural design, two other factors are paramount for the development of accurate flow models: i) having access to vast amounts of training data, annotated with high-quality flow ground-truth labels, and ii) the availability of a substantial hardware budget, that is, multiple high-end GPUs, for training complex architectures and pushing their performance to their best \cite{ sui2022craft, huang2022flowformer, shi2023flowformer++, zhao2022global, lu2023transflow, wang2024sea}. Dependence on both is common to any computer vision task based on deep learning, often making the difference between a suboptimal method or a state-of-the-art solution. As a result, academia and industry are racing to increase both the availability of training data and the number of GPUs to train deep models -- e.g., FlowFormer \cite{huang2022flowformer} and GMFlow \cite{xu2022gmflow} trained on 4$\times$ V100 GPUs, SEA-RAFT \cite{wang2024sea} on 8$\times$ 3090 RTX GPUs.
We feel the reliance on brute force to push the bar of progress can be a shortcut for low-hanging fruits, yet may preclude pursuing further methodological advances.
Furthermore, in the long run, over-reliance on hardware capabilities can make research inaccessible to groups that do not have sufficient budgets to compete with larger laboratories or companies.

Therefore, we argue that progress can be pursued even with a lower hardware budget, as evidenced by recent news from adjacent research fields such as natural language processing, where the DeepSeek model \cite{guo2025deepseek} proved unprecedented performance after being trained on a fraction of the hardware budget used by competitors \cite{gpt4o}.
We believe that similar stories can be written in computer vision by building on existing vision foundation models carefully repurposed for different tasks, aiming to \textit{recycle} the effort to training them rather than training a new solution from scratch, still with prohibitive hardware requirements. 
Recent examples involve the fine-tuning of pre-trained image generation models \cite{rombach2022high} for tasks such as depth estimation \cite{ke2024repurposing} or optical flow itself \cite{saxena2023surprising}, or embedding single-image depth foundation models \cite{yang2024depth} within deep stereo \cite{bartolomei2024stereo,wen2025stereo,cheng2025monster,jiang2025defom} or multi-view stereo \cite{izquierdo2025mvsanywhere} architectures. We believe a path similar to the latter can be taken for optical flow. 

In this paper, we introduce \textbf{\net{}}, a novel deep architecture for optical flow estimation designed at the intersection of three worlds. Indeed, \net{} harmonizes i) recent advances in the design space of optical flow networks \cite{wang2024sea}, ii) cutting-edge depth foundation models \cite{yang2025depth}, pre-trained on millions-scale datasets, and iii) low-dimensional motion parametrization \cite{heeger1992subspace} from the classical
computer vision literature. 
By connecting these components at the opposites of a 30-year time span, \net{} implements a compact solution for optical flow that can be trained on a single consumer-scale GPU, yet achieve state-of-the-art accuracy and fine-grained details -- as shown in Fig. \ref{fig:teaser}, with models trained on TartanAir \cite{wang2020tartanair}, FlyingChairs \cite{dosovitskiy2015flownet},  FlyingThings3D \cite{mayer2016large} and tested on Spring \cite{mehl2023spring}.

Our contributions can be summarized as follows:

\begin{itemize}
    \item We introduce \net, the first optical flow model that integrates a pre-trained depth foundation model. 

    \item We explore different design strategies to best exploit the prior knowledge of the foundation model for the optical flow estimation task.

    \item We develop several variants of \net, implementing different trade-offs between accuracy and efficiency, yet maintaining the single-GPU requirement at training time.
\end{itemize}

\section{Related Work}

\textbf{Optical Flow.} Optical flow estimation has evolved from classical approaches that treated it as an optimization problem \cite{horn1981determining, black1993framework, zach2007duality} to modern deep learning methods. The field was revolutionized by FlowNet \cite{dosovitskiy2015flownet, ilg2017flownet2}, which first formulated flow estimation as a supervised learning problem and later introduced stacked architectures to improve accuracy. SpyNet \cite{ranjan2017optical} combined classical spatial pyramid concepts with deep learning, while PWC-Net \cite{sun2018pwc} advanced the field by incorporating warping and cost volumes. Later work focused on efficiency through lightweight architectures \cite{hui18liteflownet, hui20liteflownet2} and improved volumetric processing \cite{yang2019volumetric}. In particular, PWC-Net+ \cite{sun2019models} demonstrated the importance of training protocols beyond architectural choices.

A major breakthrough came with RAFT \cite{teed2020raft}, which established a new paradigm through iterative refinement and multi-scale cost volumes, with notable follow-up work \cite{sun2022disentangling} disentangling the contributions of architecture and training.  This has led to many architectural advances: SEA-RAFT \cite{wang2024sea} enhanced accuracy through mixture-of-Laplace loss and rigid-motion pre-training, while several approaches explored efficient architectures \cite{morimitsu2024recurrent, deng2023explicit} and high-resolution estimation \cite{zheng2022dip, jahedi2024ms, jahedi2024ccmr}. The emergence of transformer architectures led to further significant improvements, with FlowFormer \cite{huang2022flowformer}, FlowFormer++ \cite{shi2023flowformer++}, CRAFT \cite{sui2022craft}, GMFlowNet \cite{zhao2022global}, and efficient high-resolution approaches \cite{leroy2023win} leveraging various forms of attention for global context. Other notable developments have addressed specific challenges like occlusion handling and spatial affinity \cite{jiang2021learning, luo2022learning, sun2022skflow}, while novel directions include unifying flow with stereo and depth estimation \cite{xu2022gmflow, xu2023unifying} and leveraging geometric matching pretraining \cite{dong2023rethinking}. Among emerging approaches, diffusion models \cite{saxena2023surprising} demonstrated surprising effectiveness without task-specific designs.

While our focus is on supervised learning, the field has seen parallel evolution in other directions. Temporal information has been explored by multi-frame approaches \cite{shi2023videoflow, cho2024flowtrack, wang2023tracking}, while data scarcity has been addressed by unsupervised learning \cite{jonschkowski2020matters, liu2020learning} and joint learning with related tasks \cite{jiao2021effiscene, liu2020flow2stereo}. The challenge of training data has inspired various solutions, from automatic generation and unlabelled video synthesis \cite{sun2021autoflow, han2022realflow} to novel approaches such as single-image flow synthesis \cite{aleotti2021learning} and augmentation strategies \cite{jeong2023distractflow}.

\begin{figure*}[t]
    \centering
    \begin{overpic}[width=1.\textwidth]{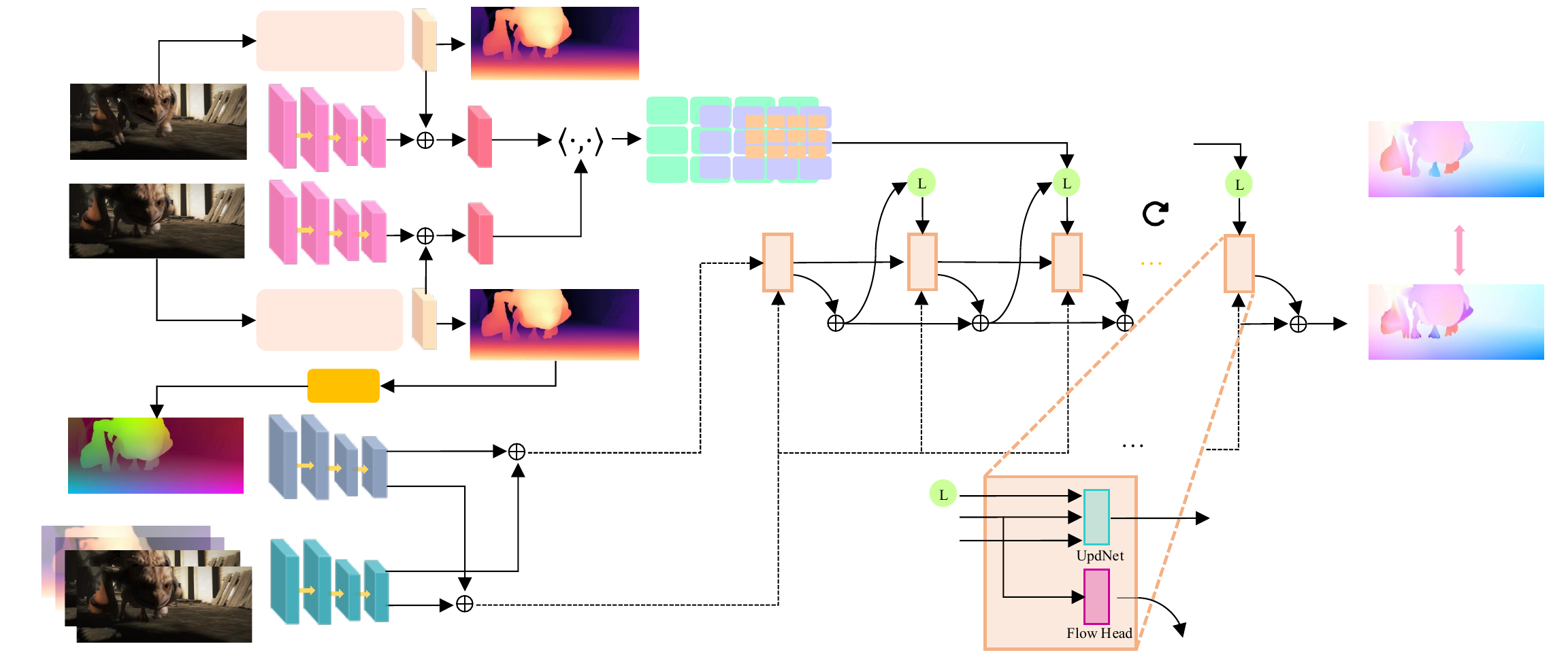}
    \put (2.5,34) {\tiny{$\mathbf{I}_0$}}
    \put (2.5,27.6) {\tiny{$\mathbf{I}_1$}}
    \put (27.5,36.5) {\tiny{$\mathbf{\Phi}_0$}}
    \put (27.5,23.5) {\tiny{$\mathbf{\Phi}_1$}}
    \put (24.7,34.2) {\tiny{$\mathbf{F}_0$}}
    \put (24.7,28) {\tiny{$\mathbf{F}_1$}}
    \put (32,34) {\tiny{$\mathbf{F}_0^\mathbf{\Phi}$}}
    \put (32,27.8) {\tiny{$\mathbf{F}_1^\mathbf{\Phi}$}}
    \put (41.2,39) {\tiny{$\mathbf{D}_0$}}
    \put (41.2,21) {\tiny{$\mathbf{D}_1$}}
    \put (17,39) {\tiny{Depth Anything V2}}
    \put (20.5,39.9) {\includegraphics[width=0.014\textwidth]{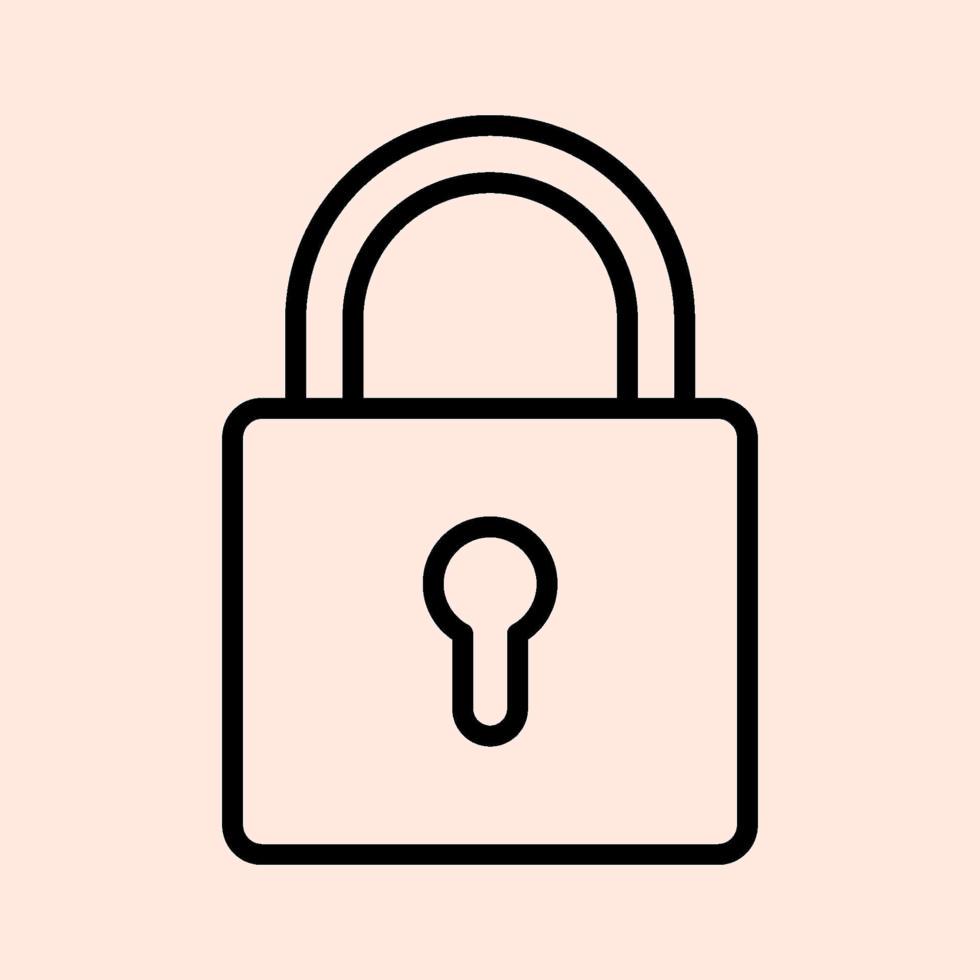}}
    \put (17,21.3) {\tiny{Depth Anything V2}}
    \put (20.5,22.1) {\includegraphics[width=0.014\textwidth]{images/lock.jpg}}
    \put (17.8,24.2) {\tiny{Feature Extractor}}
    \put (18.5,8.9) {\tiny{Base Network}}
    \put (18.2,0.9) {\tiny{Context Network}}
    \put (42.2,29.3) {\tiny{4D Correlation Volumes}}
    \put (90.3,28.5) {\tiny{Ground Truth}}
    \put (88.3,18.1) {\tiny{Estimated Optical Flow}}
    \put (20.7,17.1) {\tiny{Bases}}
    \put (8.5,9.6) {\tiny{$\mathcal{B}_\text{motion}$}}
    \put (5,-0.2) {\tiny{$\mathbf{I}_0 (\oplus \mathbf{D}_0) \oplus \mathbf{I}_1 (\oplus \mathbf{D}_1$)}}
    \put (25,6.2) {\tiny{$\mathbf{H}^0$}}
    \put (25,4) {\tiny{$\mathbf{C}$}}
    \put (25,14) {\tiny{$\mathbf{H}_\mathcal{B}$}}
    \put (25,11.5) {\tiny{$\mathbf{B}$}}
    
    \put (59.3,8.8) {\tiny{$\mathbf{H}^i$}}
    \put (77.7,8.8) {\tiny{$\mathbf{H}^{i+1}$}}
    \put (59.5,7.5) {\tiny{$\mathbf{C}$}}
    \put (75.7,4)  {\tiny{$\mathcal{F}^j$}}
    \put (72.5,30) {\tiny{$\times N$}}

    \end{overpic}\vspace{-0.2cm}
    \caption{\textbf{Architecture Overview.} Our proposed \net{} architecture processes a pair of images $\mathbf{I}_0, \mathbf{I}_1$ through parallel paths: a shared-weight feature extractor produces $\mathbf{F}_0, \mathbf{F}_1$, while a depth foundation model (\eg Depth Anything v2 \cite{yang2025depth}) estimates depth maps $\mathbf{D}_0, \mathbf{D}_1$ and features $\mathbf{\Phi}_0, \mathbf{\Phi}_1$. These are combined to obtain enriched features $\mathbf{F}_0^\mathbf{\Phi}, \mathbf{F}_1^\mathbf{\Phi}$ for building 4D correlation volumes $\{\mathbf{V}^s\}_s^S$. A Base Network extracts motion features $\mathbf{B}, \mathbf{H}_\mathcal{B}$ from geometric bases $\mathcal{B}_\text{motion}$, while a Context Network processes images (and optionally depths) to obtain context features $\mathbf{C}, \mathbf{H}^0$. Flow $\mathcal{F}^j$ is iteratively refined through an UpdNet processing correlation lookups (L),  hidden states $\mathbf{H}^i$ and context, with a FlowHead predicting updated at each step. Training relies on log-likelihood maximization over Laplacian mixture distributions. }\vspace{-0.3cm}
    \label{fig:framework}
\end{figure*}

\textbf{Motion and Flow Bases}. Classical approaches to motion estimation leverage the fact that camera motion induces six-dimensional linear subspace of possible flow fields \cite{heeger1992subspace}. This was leveraged by \cite{Wulff2015} who introduced a learned PCA basis for efficient sparse-to-dense flow estimation. Recent methods have expanded these concepts: \cite{ye2021motion, liu2022unsupervised} used flow bases for unsupervised homography estimation, Bowen et al. \cite{Bowen20223dv} proposed learning scene representations through flow subspaces, while Safadoust and Güney \cite{safadoust2023ICCV} demonstrated their utility for multi-object discovery through depth-aware flow decomposition. We build upon this line of work by combining motion bases with modern foundation models for depth estimation to bootstrap flow estimation. 

\textbf{Vision Foundation Models (VFMs) for Geometry.} VFMs have greatly advanced computer vision through large-scale pre-training. While models like CLIP \cite{radford2021learning} have demonstrated strong capabilities in image-level tasks, and DINO \cite{caron2021emerging} in dense representation learning, their application to geometric vision problems has expanded rapidly across 3D reconstruction \cite{wang2024dust3r, zhang2024monst3r}, pose estimation \cite{wen2024foundationpose}, and depth prediction. In depth estimation, models like Depth Anything \cite{yang2024depth} and its successor Depth Anything v2 \cite{yang2025depth} have achieved state-of-the-art results through synthetic data training and knowledge distillation, alongside other VFMs for monocular depth estimation \cite{ranftl2020towards, Ranftl2021, bochkovskii2024depth, ke2024repurposing}. Similarly, stereo matching has also benefited from VFMs, thanks to specialized adapters \cite{bartolomei2024stereo, zhou2025all, li2024roadformer, cheng2025monster, wen2025stereo, jiang2025defom} that mitigate the domain gap between pre-trained ViT features and geometric matching requirements. Despite these advances across geometric tasks, the potential of VFMs, particularly their rich semantic and geometric understanding, remains largely unexplored in the context of optical flow estimation. Our work addresses this gap by leveraging the prior knowledge infused in depth foundation models for flow estimation.

\section{Method Overview}

\subsection{Optical Flow Backbone}

\net{} is built on top of an optical flow backbone. Following the latest trends, we 
start over SEA-RAFT \cite{wang2024sea}, a model assembling four basic modules, as shown in Fig. \ref{fig:framework}.

\textbf{Features Extractor.} The first step performed by any flow backbone involves extracting meaningful features from input images. For this purpose, either a convolutional neural network (CNN) or a Vision Transformer (ViT) can be employed. We deploy a classical CNN from the ResNet family \cite{he2016deep} as our $\texttt{FeatNet}$, which we initialize with ImageNet pre-trained weights. This network processes both images $\mathbf{I}_0$ and $\mathbf{I}_1$ to generate dense feature maps $\mathbf{F}_0$ and $\mathbf{F}_1$:

\begin{equation}
    \mathbf{F}_0 = \texttt{FeatNet}(\mathbf{I}_0), \hspace{1cm} \mathbf{F}_1 = \texttt{FeatNet}(\mathbf{I}_1)    
\end{equation}
 at $\frac{1}{8}$ of the input resolution and with $K$ channels.

\textbf{All-Pair Correlation Volume.} These features are used to measure the per-pixel similarity between all possible candidate matches across the two images by building a 4D all-pair correlation volume \cite{teed2020raft}. For any pair of pixels in $\mathbf{I}_0$ and $\mathbf{I}_1$, respectively at coordinates $ij$ and $uv$, their correlation is computed as the dot product between the extracted features $\mathbf{F}_0$ and $\mathbf{F}_1$. A correlation volume pyramid $\{ \mathbf{V}^s \}_s^S$ is then built: 

\begin{equation}
    \mathbf{V}^s(ijuv) = \sum_{k=0}^{K} \mathbf{F}_0(ijk) \cdot \mathbf{F}^s_1(uvk)
\end{equation}
with $\mathbf{F}^s_1$ being features $\mathbf{F}_1$ downsampled through an $\texttt{AvgPool}$ layer with stride $s$ for different scales $S$.

\textbf{Context Network.} In addition to the $\texttt{FeatNet}$ extracting $\mathbf{F}_0$, $\mathbf{F}_1$, we deploy a further module, a $\texttt{ContexNet}$, to extract contextual features that guide the iterative flow estimation process. Specifically, following \cite{wang2024sea}, this model processes both images to extract context features $\mathbf{C}$, as well as the initial hidden state $\mathbf{H}^0$ to bootstrap the iterative flow estimation process:
\begin{equation}
    \mathbf{C}, \mathbf{H}^0 = \texttt{ContexNet}(\mathbf{I}_0 \oplus \mathbf{I}_1)
\end{equation}
with $\oplus$ being the concatenation operator.

\textbf{Flow Head.} Optical flow $\mathbf{\Delta}_\mathcal{F}$ is iteratively estimated from the hidden state $\mathbf{H}$. Specifically, a shallow $\texttt{FlowHead}$ predicts an initial flow $\mathbf{\Delta}_\mathcal{F}^0$ from $\mathbf{H}^0$, and progressively refines it through residual updates $\mathbf{\Delta}_\mathcal{F}^i$. At iteration $j$, the current flow estimate  $\mathcal{F}^j$ is defined as:

\begin{equation}
   \mathcal{F}^j = \sum_{i=0}^j \mathbf{\Delta}_\mathcal{F}^i = \sum_{i=0}^j \texttt{FlowHead}(\mathbf{H}^i)
\end{equation}
where $\mathbf{H}^i$ is the hidden state at iteration $i$. For efficiency, the $\texttt{FlowHead}$ predicts residuals at $\frac{1}{8}$ resolution, which are then upsampled using convex upsampling \cite{teed2020raft}.

\textbf{Update Operator.} Progressive refinement is performed by recurrently updating the hidden state $\mathbf{H}$ with an $\texttt{UpdNet}$, a recurrent network processing context feature $\mathbf{C}$, the current hidden state $\mathbf{H}^i$ and correlation scores retrieved from the pyramid $\{\mathbf{V}^s\}_s^S$ by means of a look-up operation based on current flow estimate $\mathbf{\Delta}_\mathcal{F}$ and a radius $r$, predicting an updated hidden state $\mathbf{H}^{i+1}$:

\begin{equation}
    \mathbf{H}^{i+1} = \texttt{UpdNet}(\mathbf{C}, \mathbf{H}^{i}, \texttt{LookUp}(\{\mathbf{V}^s\}_s^S, \mathbf{\Delta}_\mathcal{F}^i, r) )
\end{equation}

\subsection{Depth Foundation Model}

With the increasing availability of deep models trained on web-scale datasets, we aim to transfer the knowledge of such foundation models to the optical flow task. 
Specifically, we select candidates from the adjacent literature concerning single-image depth estimation \cite{Ranftl2021,yang2024depth,yang2025depth}, given the relationship between 3D geometry and induced optical flow on images -- i.e., given a fixed motion of the camera or of a subject in the scene, pixels move in the images proportionally to the inverse depth of the 3D points they represent.

Given a depth foundation model $\texttt{FoundModel}$, we use it to predict inverse depth maps $\mathbf{D}_0, \mathbf{D}_1$ for both images $\mathbf{I}_0, \mathbf{I}_1$. During inference, we also retain the very last features produced by the decoder before depth regression, $\mathbf{\Phi}_0$ and $\mathbf{\Phi}_1$, since they are strongly correlated with it \cite{wen2025stereo,jiang2025defom}:

\begin{equation}
\begin{split}
    \mathbf{\Phi}_0, \mathbf{D}_0 ={} &\texttt{FoundModel}(\mathbf{I}_0), \\
    \mathbf{\Phi}_1, \mathbf{D}_1 ={} &\texttt{FoundModel}(\mathbf{I}_1)  
\end{split}
\end{equation}
Then, we enrich the original features $\mathbf{F}_0, \mathbf{F}_1$ extracted by the flow backbone by concatenating them with those obtained from the depth foundation model. This is carried out after processing the latter with a shallow $\texttt{BottNeck}$ network \cite{wen2025stereo}, composed of three 3$\times$3 convolutional layers with stride 2 to downsample resolution to $\frac{1}{8}$ -- the same resolution at which the flow backbone returns features $\mathbf{F}_0, \mathbf{F}_1$:

\begin{equation}
\begin{split}
    \mathbf{F}_0^\mathbf{\Phi} ={} &\texttt{FeatNet}(\mathbf{I}_0) \oplus \texttt{BottNeck}(\mathbf{\Phi}_0), \\ 
    \mathbf{F}_1^\mathbf{\Phi} ={} &\texttt{FeatNet}(\mathbf{I}_1) \oplus \texttt{BottNeck}(\mathbf{\Phi}_1) 
\end{split}
\end{equation}
Enriched features  $\mathbf{F}_0^\mathbf{\Phi}, \mathbf{F}_1^\mathbf{\Phi}$ are then used to build the correlation volumes pyramid. Concurrently, $\mathbf{D}_0, \mathbf{D}_1$ can also be forwarded to the $\texttt{ContextNet}$ together with the images to extract stronger contextual and hidden state features:

\begin{equation}
    \mathbf{C}, \mathbf{H}^0 = \texttt{ContexNet}(\mathbf{I}_0 \oplus \mathbf{D}_0 \oplus \mathbf{I}_1 \oplus \mathbf{D}_1)
\end{equation}

\setlength\tabcolsep{.2em}
\begin{table*}[t]
    \centering
    \resizebox{1.0\linewidth}{!}{
    \renewcommand{\tabcolsep}{10pt}
    \begin{tabular}{lccccccccccr}
    \toprule
        \multirow{2}{*}{Model Name} & \multicolumn{3}{c}{Priors} & \multirow{2}{*}{Iters.} & \multicolumn{2}{c}{Backbones} & \multicolumn{2}{c}{TartanAir (val)} & \multicolumn{2}{c}{KITTI 2012 (train)} &\multirow{2}{*}{\#MACs} \\
    \cmidrule(l{0.5ex}r{0.5ex}){2-4}\cmidrule(l{0.5ex}r{0.5ex}){6-7}\cmidrule(l{0.5ex}r{0.5ex}){8-9}\cmidrule(l{0.5ex}r{0.5ex}){10-11}
        & $\mathbf{\Phi}_{0,1}$ & $\mathbf{D}_{0,1}$ & \texttt{BaseNet} & & Optical Flow & Depth & EPE & 1px & Fl-EPE & Fl-All \\ 
    \midrule
         SEA-RAFT (S) & & & & 4 & SEA-RAFT (S) & & 1.38 & 6.24 & 1.94 & 6.31 & 284.7G\\
         SEA-RAFT (S*) & & & & 12 & SEA-RAFT (S) & & 1.28 & 5.83 & 1.86 & 6.16 & 452.9G \\
         SEA-RAFT (M) & & & & 4 & SEA-RAFT (M) & & 1.35 & 6.13 & 1.91 & 5.93 & 486.9G\\
         SEA-RAFT (L) & & & & 12 & SEA-RAFT (M) & & 1.30 & 5.93 & 1.79 & 5.99 & 655.1G\\
         \midrule 
         \textcolor{purple}{\multirow{6}{*}{\net{} (T)}} & $\checkmark$ & & & 4 & \multirow{7}{*}{SEA-RAFT (S)} & \multirow{7}{*}{Depth Any. v2 (S)} & 1.30 & 5.88 & 1.76 & 5.69 & 435.0G\\
         & & $\checkmark$ & & 4 &  &  & 1.15 & 5.36 & 1.42 & 4.67 & 435.4G\\
         & $\checkmark$ & $\checkmark$ & & 4 &  &  & 1.11 & 5.20 & 1.43 & 4.70 & 400.2G\\
         & & & $\checkmark$ & 4 &  &  & 1.04 & 4.79 & 1.31 & 4.23 & 659.5G\\
         \rowcolor{LightYellow}
         & $\checkmark$ & & $\checkmark$ & 4 &  &  & 
         1.03 & 4.72 & 1.30 & 4.16 & 694.7G\\
         & $\checkmark$ & $\checkmark$& $\checkmark$ & 4 &  &  & 1.03 & 4.82 & 
         1.29 & 4.19 & 694.7G\\
         \midrule
         \textcolor{orange}{\net{} (S)} & $\checkmark$ & $\checkmark$& $\checkmark$ & 12 & SEA-RAFT (S) & Depth Any. v2 (S) & 0.99 & 4.50 & 1.20 & 3.95 & 1241.9G \\
         \textcolor{orange}{\net{} (M)} & $\checkmark$ & $\checkmark$& $\checkmark$ & 4 & \multirow{2}{*}{SEA-RAFT (M)} & \multirow{2}{*}{Depth Any. v2 (B)} & 0.90 & 4.15 & 1.35 & 4.37 & 1312.2G \\
         \textcolor{orange}{\net{} (L)} & $\checkmark$ & $\checkmark$& $\checkmark$ & 12 &  &  & 0.85 & 3.87 & 1.25 & 4.12 & 1859.4G \\
         \midrule
         \textcolor{teal}{\multirow{3}{*}{\net{} (T)}} & $\checkmark$ & $\checkmark$& $\checkmark$ & 4 & \multirow{3}{*}{SEA-RAFT (S)} & DPT-Hybrid & 1.08 & 5.10 & 1.52 & 5.19 & 865.6G\\
         & $\checkmark$ & $\checkmark$& $\checkmark$ & 4 & & Depth Any. v1 (S) & 1.04 & 5.01 & 1.44 & 4.82 & 694.7G\\
         & $\checkmark$ & $\checkmark$& $\checkmark$ & 4 & & Depth Any. v2 (S) & 1.03 & 4.72 & 1.30 & 4.16 & 694.7G\\
         \midrule 
         \textcolor{blue}{CRAFT} & & & & 12 & \multirow{2}{*}{CRAFT} & \multirow{4}{*}{Depth Any. v2 (S)} & 1.77 & 8.31 & 2.17 & 9.03 & 315.6G\\
         \textcolor{blue}{CRAFT (\net)} & & & $\checkmark$ & 12 & & & 1.39 & 7.00 & 1.62 & 7.22 & 423.6G\\
         \textcolor{blue}{FlowFormer} & & & & - & \multirow{2}{*}{FlowFormer} & & 1.63 & 7.57 & 2.67 & 9.13 & 974.6G\\
         \textcolor{blue}{FlowFormer (\net)} & & & $\checkmark$ & - & & & 1.30 & 6.06 & 1.54 & 6.36 & 1587.6G \\
    \bottomrule
    \end{tabular}
    }\vspace{-0.3cm}
    \caption{\textbf{Ablation and Generality Studies.} We ablate different \textbf{\textcolor{purple}{priors combinations}}, \textbf{\textcolor{orange}{model sizes}}, \textbf{\textcolor{teal}{depth foundation models}}, and \textbf{\textcolor{blue}{optical flow backbones}} on TartanAir and KITTI 2012. The impact is measured against baseline SEA-RAFT models, reported at the top. All the models are trained for 100K steps on TartanAir \cite{wang2020tartanair}, using a single RTX 3090 GPU. (S)* means SEA-RAFT (S) running 12 iterations.}
    \label{tab:ablations}
    \vspace{-0.3cm}
\end{table*}

\begin{table}[]
    \centering
    \resizebox{1.0\linewidth}{!}{
    \renewcommand{\tabcolsep}{4pt}
    \begin{tabular}{cc}
    \begin{tabular}{lcccc}
    \\
    \toprule    
    \multirow{2}{*}{\texttt{BaseNet} inputs} & \multicolumn{2}{c}{TartanAir (val)} & \multicolumn{2}{c}{KITTI 2012 (train)}  \\ 
    \cmidrule(l{0.5ex}r{0.5ex}){2-3}
    \cmidrule(l{0.5ex}r{0.5ex}){4-5}
    & EPE & 1px & Fl-EPE & Fl-All \\ 
    \midrule
    $\mathbf{D}_{0}$ & 1.05 & 4.79 & \bf 1.27 & 4.21 \\
    $\mathcal{B}_\text{motion}$ & \bf 1.03 & \bf 4.72 & 1.30 & \bf 4.16 \\
    \midrule
    \end{tabular}&
    \begin{tabular}{cccc}
    \multicolumn{4}{c}{KITTI 2015 (val)}  \\
    \toprule    
    \multicolumn{2}{c}{static} & \multicolumn{2}{c}{dynamic}  \\ 
    \cmidrule(l{0.5ex}r{0.5ex}){1-2}
    \cmidrule(l{0.5ex}r{0.5ex}){3-4}
    Fl-EPE & Fl-All & Fl-EPE & Fl-All \\ 
    \midrule
    1.21 & 2.07 & \bf 2.49 & \bf 8.61 \\
    \bf 1.06 & \bf 1.96 & 2.60 & 9.07 \\
    \midrule
    \end{tabular} \\
    \end{tabular}}\vspace{-0.3cm}
    \caption{\textbf{Ablation Study -- different inputs to the \texttt{BaseNet}.} Model: \net{} (T). On the right: models fine-tuned on the first 160 images of KITTI 2015 training set, evaluated on the other 40.}
    \label{tab:depth_vs_bases}\vspace{-0.3cm}
\end{table}

\subsection{Low-Dimensional Motion Priors}

We choose to cast the foundation model priors into a form more suitable for our model. 
Specifically, it is well established \cite{heeger1992subspace} that for a static scene with known depth, the space of possible optical flow fields can be reduced to a linear combination of six basis vectors, corresponding to the six degrees of freedom in 3D motion:

\begin{equation}
    \label{eq:six_basis}
    \mathcal{B}_\text{motion} = \{\mathbf{\Delta}_{\bT x}, \mathbf{\Delta}_{\bT y}, \mathbf{\Delta}_{\bT z}, \mathbf{\Delta}_{\bR x}, \mathbf{\Delta}_{\bR y}, \mathbf{\Delta}_{\bR z}\}
\end{equation}

These six bases are categorized into two types: three model the translational component of motion and depend on inverse depth $\mathbf{D}_0$, while the other three represent rotational components. The six bases are defined as:

\begin{eqnarray}
\label{eqn:basis-camera}
        &\mathbf{\Delta}_{\bT x} = \begin{bmatrix}
              f_x~\mathbf{D}_0 \\
             0
          \end{bmatrix}, 
        &\mathbf{\Delta}_{\bR x} =  \begin{bmatrix}
              {f_y}^{-1}~\bar{\mathbf{U}}~\bar{\mathbf{V}} \\
             f_y + {f_y}^{-1}~\bar{\mathbf{V}}^2
          \end{bmatrix} \nonumber \\
        \nonumber \\ %
        &\mathbf{\Delta}_{\bT y} = \begin{bmatrix}
              0 \\
             f_y~\mathbf{D}_0
          \end{bmatrix}, 
        &\mathbf{\Delta}_{\bR y} = \begin{bmatrix}
            f_x + {f_x}^{-1}~\bar{\mathbf{U}}^2 \\
            {f_x}^{-1}~\bar{\mathbf{U}}~\bar{\mathbf{V}}
        \end{bmatrix} \nonumber \\
        \nonumber \\ %
        &\mathbf{\Delta}_{\bT z} = \begin{bmatrix}
             -\bar{\mathbf{U}}~\mathbf{D}_0 \\
             -\bar{\mathbf{V}}~\mathbf{D}_0
          \end{bmatrix},
        &\mathbf{\Delta}_{\bR z} = \begin{bmatrix}
             f_x~{f_y}^{-1}~\bar{\mathbf{V}} \\
             -f_y~{f_x}^{-1}~\bar{\mathbf{U}}
          \end{bmatrix}  
\end{eqnarray}
where $f_x,f_y$ represent the camera focal length, and $~\bar{\mathbf{U}},~\bar{\mathbf{V}}$ are pixel coordinates grids normalized according to the principal point $(c_x,c_y)$, which we can assume to be at the center of the image. This formulation, however, would require explicit knowledge of camera focal length. 
As the bases are combined linearly, we can arbitrarily scale them and eliminate the focal length requirement \cite{Bowen20223dv,safadoust2023ICCV}. Specifically, by assuming $f_x = f_y$, we can remove the focal length from basis $\mathbf{\Delta}_{\bR z}$, while we can split $\mathbf{\Delta}_{\bR x}$ and $\mathbf{\Delta}_{\bR y}$ and redefine them as a linear combination of two sub-bases:
\begin{eqnarray}
    \mathbf{\Delta}_{\bR^1 x} =  \begin{bmatrix}
        0 \\ 1 
    \end{bmatrix} & \mathbf{\Delta}_{\bR^2 x} = \begin{bmatrix}
        \bar{\mathbf{U}}~\bar{\mathbf{V}} \\ \bar{\mathbf{V}}^2
    \end{bmatrix} \nonumber \\
    \mathbf{\Delta}_{\bR^1 y} =  \begin{bmatrix}
        1 \\ 0 
    \end{bmatrix} & \mathbf{\Delta}_{\bR^2 y} = \begin{bmatrix}
        \bar{\mathbf{U}}^2 \\ \bar{\mathbf{U}}~\bar{\mathbf{V}}
    \end{bmatrix}
\end{eqnarray}
Therefore, we define a set of eight bases by knowing $\mathbf{D}_0$:
 \begin{equation}
 \label{eq:eight-basis}
 \begin{split}
 \mathcal{B}_\text{motion}=\{&\mathbf{\Delta}_{\bT x}, \mathbf{\Delta}_{\bT y}, \mathbf{\Delta}_{\bT z}, \mathbf{\Delta}_{\bR^1 x}, \\ &\mathbf{\Delta}_{\bR^2 x}, \mathbf{\Delta}_{\bR^1 y}, \mathbf{\Delta}_{\bR^2 y}, \mathbf{\Delta}_{\bR z} \}
 \end{split}
 \end{equation}
as a prior for the space of possible flows. 

Although this representation holds only for rigid motions, whether from camera movement alone or from a single independently moving object in the scene, we argue it can provide an initial guess to the flow model, which will further refine it through successive iterations. 
Purposely, we introduce an additional sub-module, a $\texttt{BasesNet}$, responsible for extracting dense features from the set of bases:

\begin{equation}
    \mathbf{B}, \mathbf{H}_\mathcal{B} = \texttt{BasesNet}(\mathcal{B}_\text{motion})
\end{equation}
These features are concatenated with the original context and hidden state features $\mathbf{C}, \mathbf{H}^0$, which are used throughout the iterative flow estimation process described earlier.

\subsection{Supervision}

Following SEA-RAFT \cite{wang2024sea}, we model the output optical flow as a mixture of two Laplace distributions. Accordingly, each flow update $\mathbf{\Delta}_\mathcal{F}^i$ is defined as:

\begin{equation}
    \mathbf{\Delta}_\mathcal{F}^i = \alpha^i \cdot \frac{e^{-\frac{|x-\mu^i|}{e^{\beta_1}}}}{2e^{\beta_1}} + (1 - \alpha^i) \cdot \frac{e^{-\frac{|x-\mu^i|}{e^{\beta_2}}}}{2e^{\beta_2^i}}
\end{equation}
 where $\alpha^i,\beta_2^i,\mu^i$ are the six parameters defining the mixture (two for each flow coordinate) predicted by the $\texttt{FlowHead}$ at iteration $i$, and $\beta_1$ is fixed to 0.
Finally, optical flow predictions $\mathcal{F}^j$ at each iteration $j$ are supervised through negative log-likelihood minimization \cite{wang2024sea}:

 \begin{equation}
    \mathcal{L}_\mathcal{F} = \sum_{j=0}^{\texttt{iters}} \gamma^{N-j} (-\log \mathcal{F}^j)
\end{equation}

\section{Experimental Results}

\subsection{Implementation Details}

\net{} is implemented on top of SEA-RAFT \cite{wang2024sea} codebase. Specifically, the $\texttt{FeatNet}$, $\texttt{ContextNet}$, and $\texttt{BasesNet}$ are implemented by a subset of layers of either a ResNet-18 or a ResNet-34 \cite{he2016deep}: depending on the choice, the original SEA-RAFT model comes in two variants, respectively \textit{small} (S) or \textit{medium} (M). The number of iterative updates $\texttt{iters}$ is usually set to 4, with a third SEA-RAFT variant running 12 iterations on top of the (M) model -- namely, SEA-RAFT \textit{large} (L).
We select Depth Anything v2 \cite{yang2025depth} as the depth foundation model for our experiments, either in its \textit{small} (S) or \textit{base} (B) variants. 
By playing with different backbone sizes and iterations, we implement the \textit{tiny} (T) and \textit{small} (S) variants using ResNet-18, Depth Anything v2 (S) and setting $\texttt{iters}$ to 4 or 12 respectively; we also implement \textit{medium} (M) and \textit{large} (L) ones, by selecting ResNet-34, Depth Anything v2 (B) and setting $\texttt{iters}$ to 4 or 12 respectively. 

\setlength\tabcolsep{.6em}
\begin{table}[t]
    \centering   
    \resizebox{\linewidth}{!}{
    \renewcommand{\tabcolsep}{2pt}
    \begin{tabular}{llrrrr}
    \toprule
    \multirow{2}{*}{Extra Data} & \multirow{2}{*}{Method} & \multicolumn{2}{c}{Sintel} & \multicolumn{2}{c}{KITTI 2015}\\
    \cmidrule(l{0.5ex}r{0.5ex}){3-4}\cmidrule(l{0.5ex}r{0.5ex}){5-6}
        &  & Clean$\downarrow$ & Final$\downarrow$ & Fl-EPE$\downarrow$ & Fl-all$\downarrow$ \\ 
    \midrule
        & PWC-Net~\cite{sun2018pwc} & 2.55 & 3.93 & 10.4 & 33.7\\
        & RAFT~\cite{teed2020raft} & 1.43 & 2.71 & 5.04 & 17.4\\
        & GMA~\cite{jiang2021learning} & 1.30 & 2.74 & 4.69 & 17.1\\
        & SKFlow~\cite{sun2022skflow} & 1.22 & 2.46 & 4.27 & 15.5\\
        & FlowFormer~\cite{huang2022flowformer} & \snd 1.01 & {2.40} & 4.09$^{\dagger}$ & 14.7$^{\dagger}$\\
        & DIP~\cite{zheng2022dip} & 1.30 & 2.82 & 4.29 & 13.7\\
        & EMD-L~\cite{deng2023explicit} & \fst {0.88} & 2.55 & 4.12 & 13.5\\
        & CRAFT~\cite{sui2022craft} & 1.27 & 2.79 & 4.88 & 17.5\\
        & RPKNet~\cite{morimitsu2024recurrent} & 1.12 & 2.45 & - & 13.0\\
        & GMFlowNet~\cite{zhao2022global} & 1.14 & 2.71 & 4.24 & 15.4\\
        \hdashline
        & \multirow{2}{*}{SEA-RAFT (S)~\cite{wang2024sea}} &1.27 & 4.32 & 4.61 & 15.8\\
        & & 1.28* & 3.02* & 5.10* & 16.5*\\
        \hdashline
        & \multirow{2}{*}{{SEA-RAFT (M)~\cite{wang2024sea}}} &1.21 & 4.04 & 4.29 & 14.2\\
        &  &1.30* & 3.09* & 5.30* & 15.8*\\
        \hdashline
        & \multirow{2}{*}{SEA-RAFT (L)~\cite{wang2024sea}} &1.19 & 4.11 & {3.62} & {12.9}\\       
        & &1.21* & 3.08* & {4.37}* & {14.3}*\\
        \hdashline
        & \bf \net{} (T) & 1.12 & 2.53 & 3.95 & 12.7 \\
        & \bf \net{} (S) & 1.04 & 2.43 & \trd 3.36 & \trd 11.5 \\
        & \bf \net{} (M) & 1.15 & 2.40 & 4.59 & 13.7 \\
        & \bf \net{} (L) & 1.07 & \snd 2.21 & 3.82 & 12.5 \\
        \midrule
        Tartan & GMFlow~\cite{xu2022gmflow} & - & - & 8.70 & 24.4\\
        \midrule
        \multirow{10}{*}{Tartan}  & \multirow{2}{*}{SEA-RAFT (S)~\cite{wang2024sea}} &1.27 & 3.74 & 4.43 & 15.1\\
        &  & 1.27* & 2.89* & 4.99 & 15.7\\
        \hdashline
        & \multirow{2}{*}{SEA-RAFT (M)~\cite{wang2024sea}} & 1.27 & 3.85 & 4.30 & 14.3 \\
        &  & 1.36* & 2.91* & 5.45* & 16.0* \\
        \hdashline
        & \multirow{2}{*}{SEA-RAFT (L)~\cite{wang2024sea}} & 1.23 & 3.37 & 3.73 & 12.7 \\
        &  & 1.22* & 2.73* & 4.21* & 13.5* \\
        \hdashline
        & \bf \net{} (T) & 1.13 & 2.48 & 4.06 & 12.2 \\
        & \bf \net{} (S) & 1.05 & 2.37 & \snd 3.32 & \fst 11.0 \\
        & \bf \net{} (M) & 1.10 & \trd 2.31 & 3.99 & 12.1 \\
        & \bf \net{} (L) & \trd 1.03 & \fst 2.18 & \fst 3.31 & \snd 11.2 \\
        \midrule\midrule
        AF $\rightarrow$ AF+T & DDVM \cite{saxena2023surprising} & 1.48 & 2.22 & 3.71 & 14.07 \\
        AF $\rightarrow$ AF+T+KU+Tartan & DDVM \cite{saxena2023surprising} &  1.24 & 2.00 & 2.19 & 7.58 \\
    \bottomrule
    \end{tabular}
    }\vspace{-0.3cm}
    \caption{\textbf{Zero-Shot Generalization -- Sintel (train) and KITTI 2015 (train).} Methods on top are trained with ``C$\rightarrow$T" schedule. 
    $^{\dagger}$ denotes tiling at test time. * denotes model trained with one GPU. 
    }
    \label{tab:SK-train}
    \vspace{-0.3cm}
\end{table}

\begin{figure*}[t]
    \centering
    \renewcommand{\tabcolsep}{1pt}
    \begin{tabular}{cccc}
    
    \includegraphics[width=0.23\textwidth, frame]{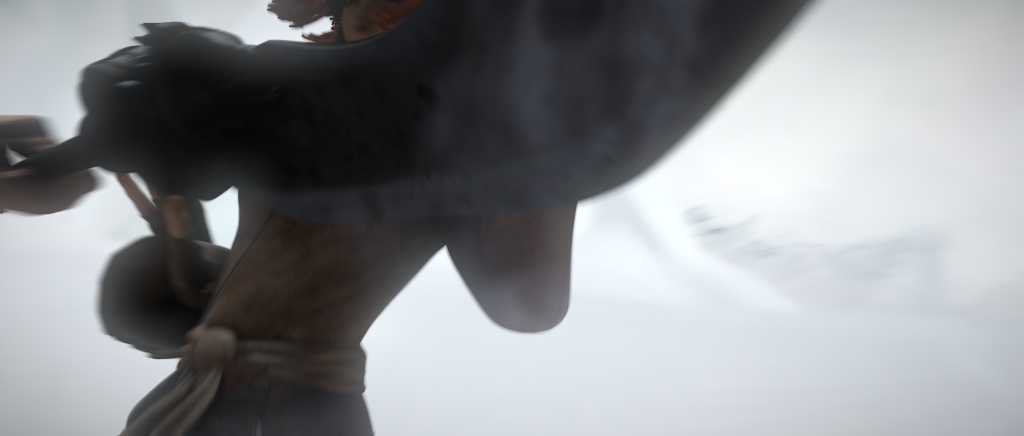} &
    \begin{overpic}[width=0.23\textwidth, frame]{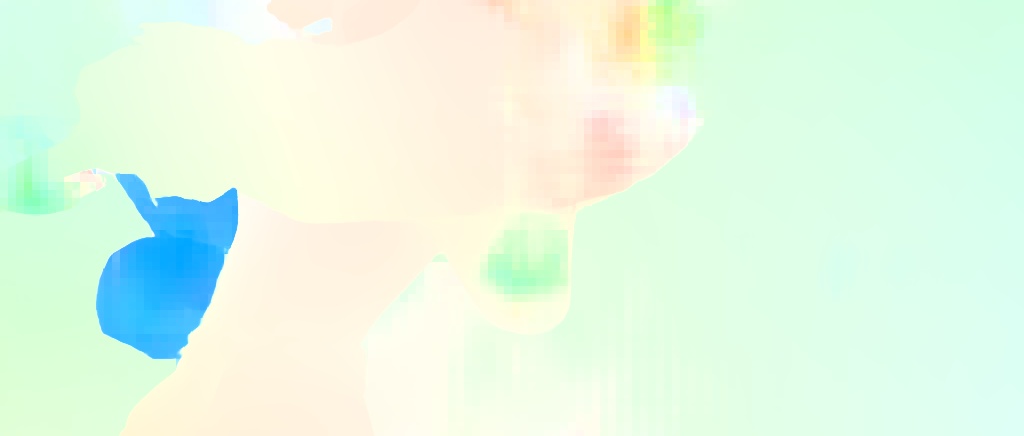}
    \put (2,35) {\textcolor{purple}{\textbf{\texttt{EPE: 23.635}}}}
    \end{overpic} &
    \begin{overpic}[width=0.23\textwidth, frame]{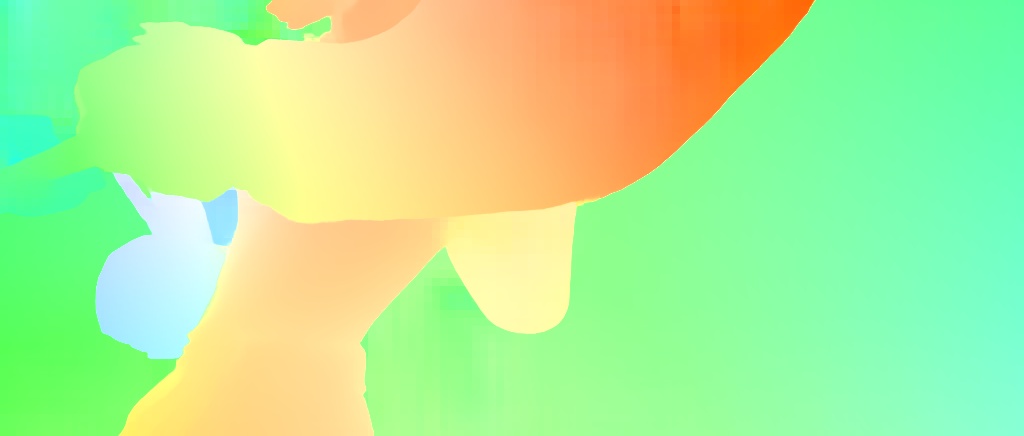}
    \put (2,35) {\textcolor{purple}{\textbf{\texttt{EPE: 4.928}}}}
    \end{overpic} &
    \begin{overpic}[width=0.23\textwidth, frame]{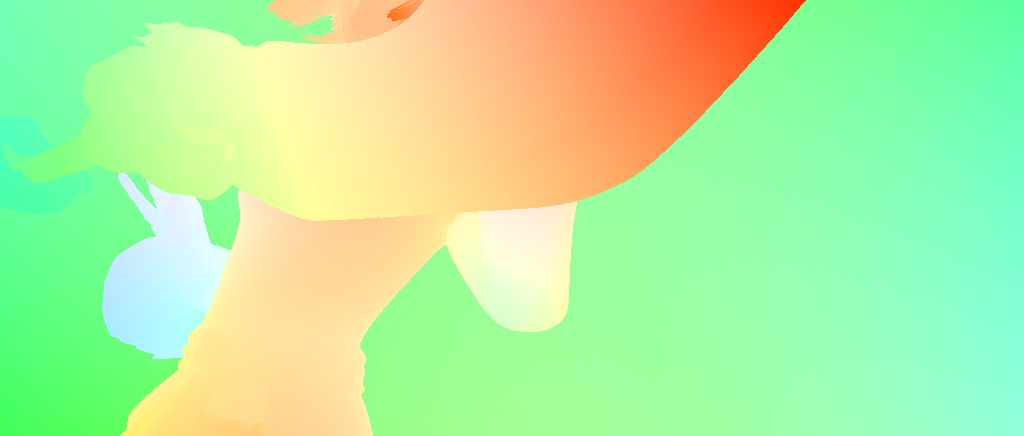}\end{overpic} \\

    \includegraphics[width=0.23\textwidth,frame]{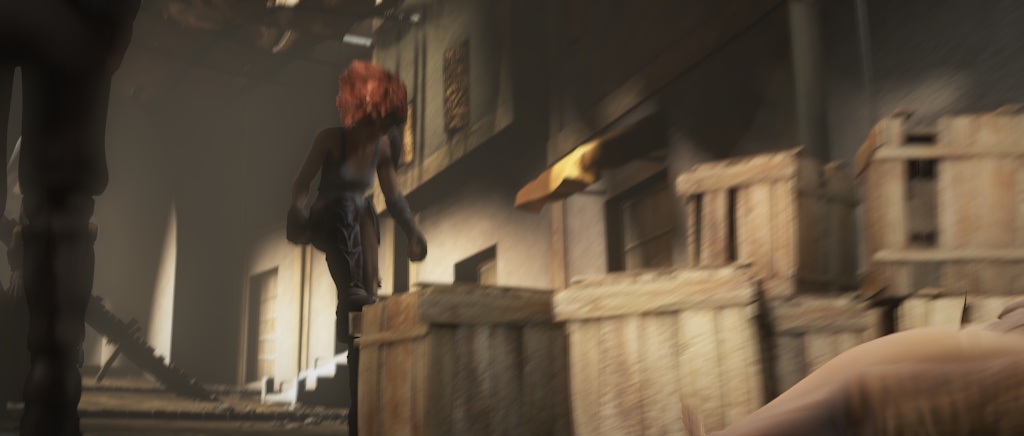} &
    \begin{overpic}[width=0.23\textwidth,frame]{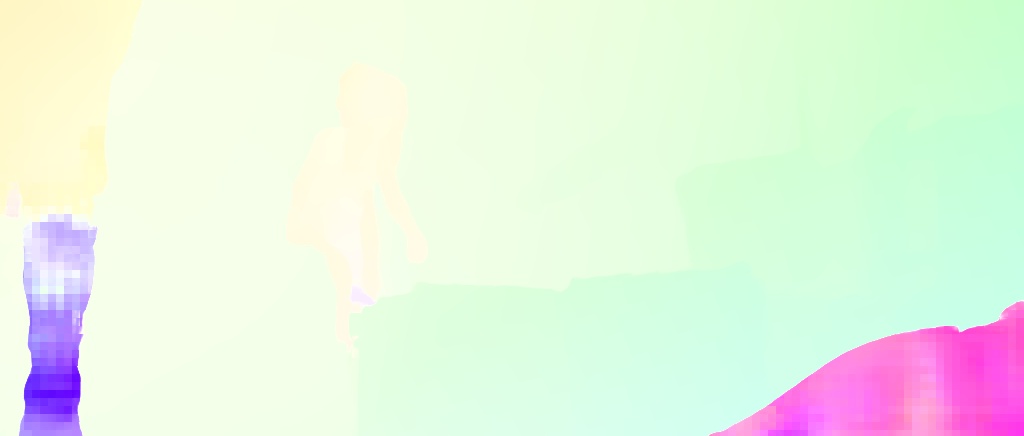}
    \put (2,35) {\textcolor{purple}{\textbf{\texttt{EPE: 7.770}}}}
    \end{overpic} &
    \begin{overpic}[width=0.23\textwidth,frame]{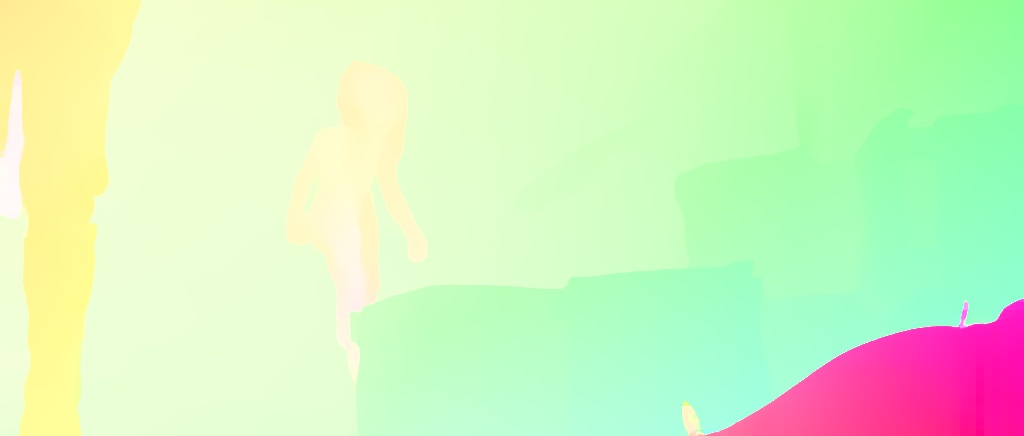}
    \put (2,35) {\textcolor{purple}{\textbf{\texttt{EPE: 2.645}}}}
    \end{overpic} &
    \begin{overpic}[width=0.23\textwidth,frame]{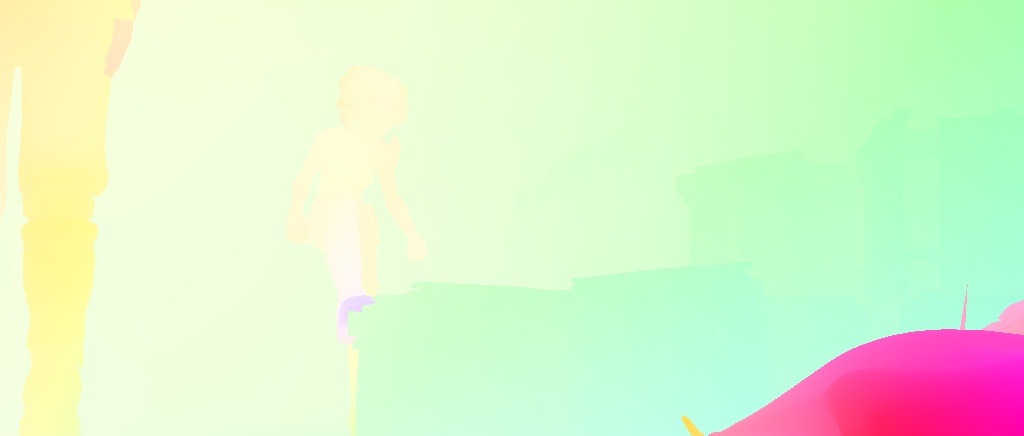}\end{overpic} \\

    \small $\mathbf{I}_0$ & \small SEA-RAFT (L) \cite{wang2024sea} & \small \bf \net{} (L) & \small Ground-truth \\

    \end{tabular}\vspace{-0.3cm}
    \caption{\textbf{Qualitative Results on Sintel \cite{butler2012naturalistic}.} From left to right: first frame, flow by SEA-RAFT (L) and \net{} (L), ground-truth flow.}\vspace{-0.3cm}
    \label{fig:sintel}
\end{figure*}

\begin{figure*}[t]
    \centering
    \renewcommand{\tabcolsep}{1pt}
    \begin{tabular}{ccc}
    
    \includegraphics[width=0.3\textwidth,frame]{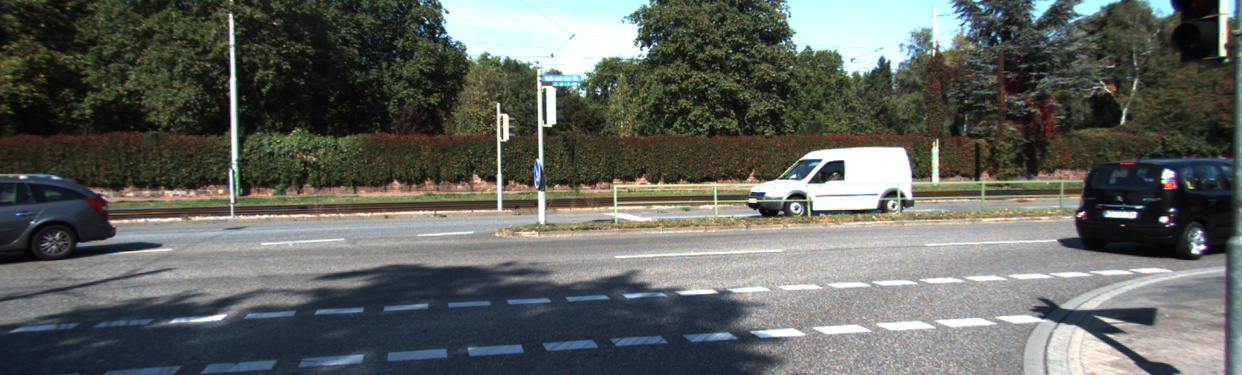} &
    \begin{overpic}[width=0.3\textwidth,frame]{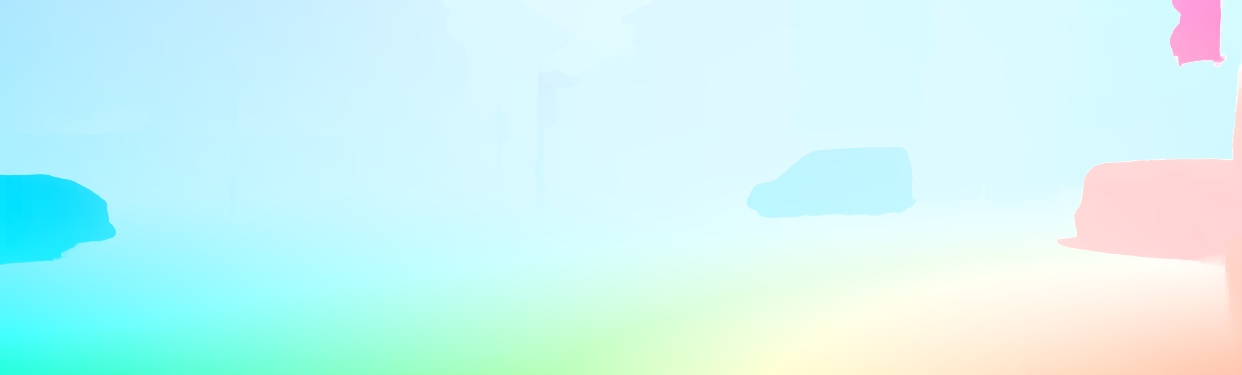}
    \put (2,25) {\textcolor{purple}{\textbf{\texttt{EPE: 1.232}}}}
    \end{overpic} &
    \begin{overpic}[width=0.3\textwidth,frame]{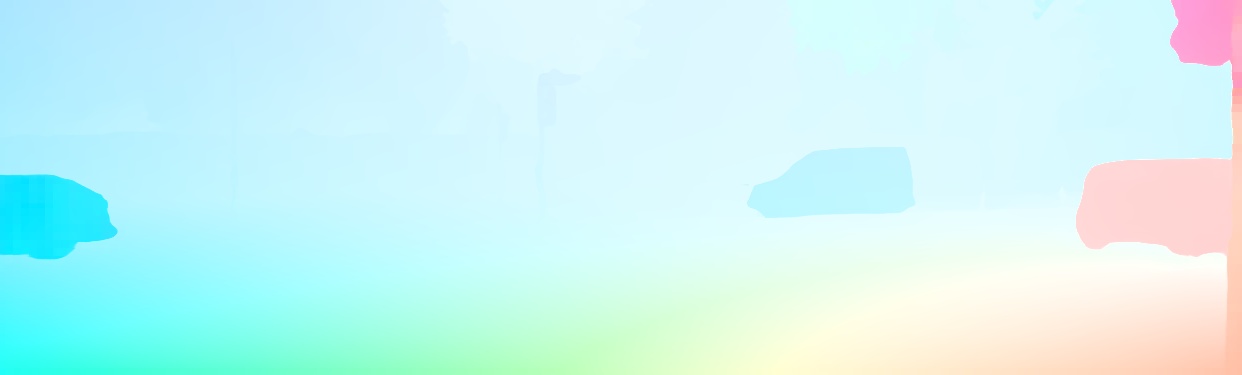} 
    \put (2,25) {\textcolor{purple}{\textbf{\texttt{EPE: 1.162}}}}
    \end{overpic}
    \\
    
    \includegraphics[width=0.3\textwidth,frame]{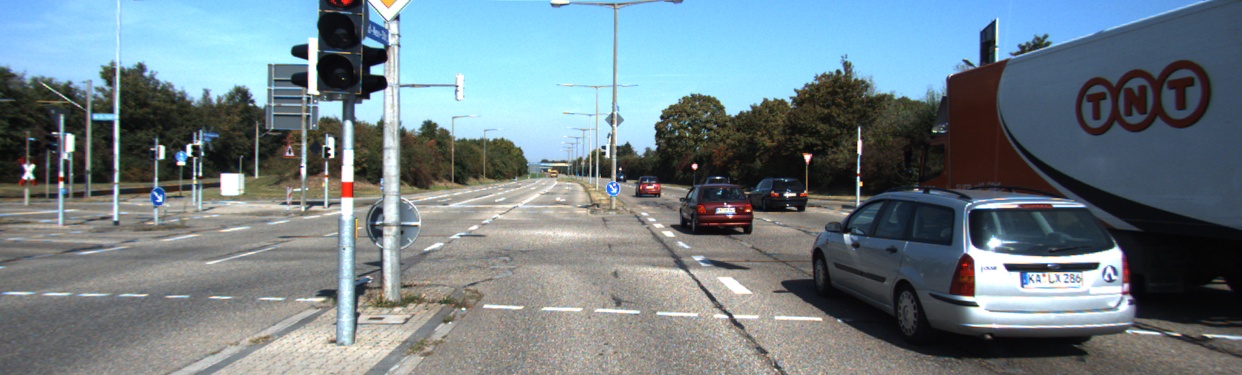} &
    \begin{overpic}[width=0.3\textwidth,frame]{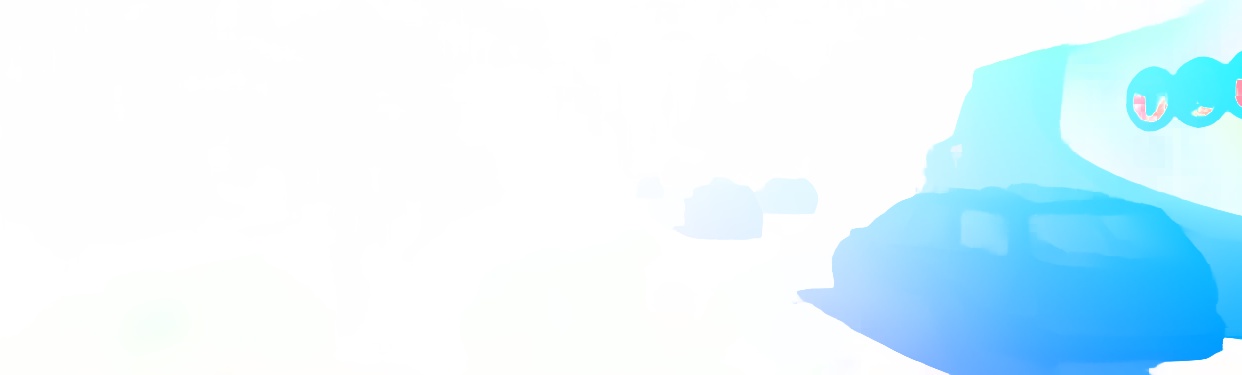}
    \put (2,25) {\textcolor{purple}{\textbf{\texttt{EPE: 1.835}}}}
    \end{overpic} &
    \begin{overpic}[width=0.3\textwidth,frame]{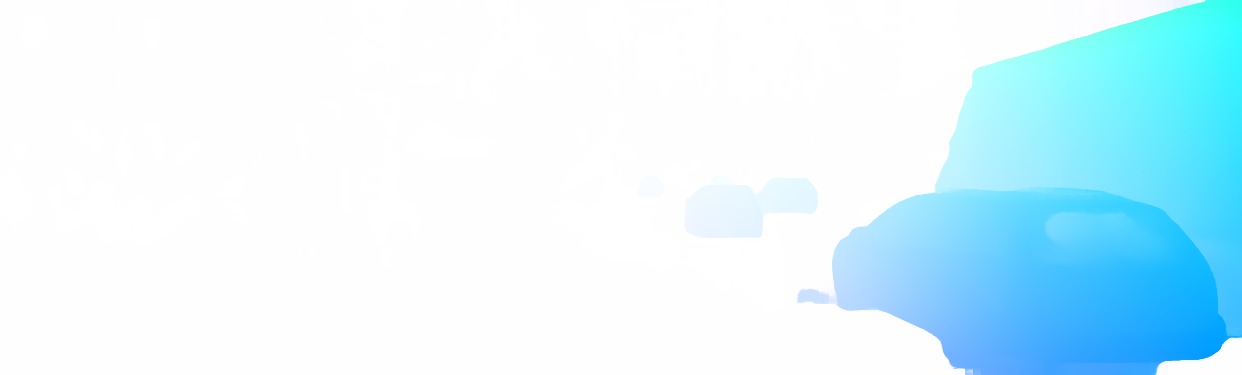}
    \put (2,25) {\textcolor{purple}{\textbf{\texttt{EPE: 1.460}}}}
    \end{overpic} \\

    \small $\mathbf{I}_0$ & \small SEA-RAFT (L) \cite{wang2024sea} & \small \bf \net{} (L) \\
    \end{tabular}\vspace{-0.3cm}
    \caption{\textbf{Qualitative Results on KITTI 2015 \cite{menze2015object}.} From left to right: first frame, flow by SEA-RAFT (L) and \net{} (L).}\vspace{-0.3cm}
    \label{fig:kitti}
\end{figure*}

\textbf{Datasets and Metrics.} Following the existing literature \cite{wang2024sea}, our experiments involve TartanAir \cite{wang2020tartanair}, FlyingChairs \cite{dosovitskiy2015flownet}, FlyingThings3D \cite{mayer2016large} and HD1K \cite{kondermann2016hci} mainly for training purposes, as well as KITTI 2012 \cite{geiger2013vision}, Spring \cite{mehl2023spring} and LayeredFlow \cite{wen2024layeredflow} mainly for evaluation purposes, with KITTI 2015 \cite{menze2015object} and Sintel \cite{butler2012naturalistic} train sets being involved in both phases. To measure the accuracy of the different flow models, we use standard metrics such as the End-Point-Error (EPE) over any dataset, with additional metrics on TartanAir and Spring (\% of pixels with error larger than 1 pixel, \textit{1px}), KITTI datasets (\% of pixels with error larger than 3 pixels or relative error higher than 5\%, \textit{Fl-All}), and LayeredFlow (\% of pixels with error larger than 1, 3 or 5 pixels, respectively \textit{1px}, \textit{3px} and \textit{5px}). 

\textbf{Training Schedule.} For ablation studies, we train the original SEA-RAFT and \net{} variants on TartanAir \cite{wang2020tartanair}, excluding the \texttt{westerndesert} and \texttt{soulcity} sequences, which we reserve for evaluation. Each training run is carried out for 100K steps, using standard hyper-parameters from SEA-RAFT codebase \cite{wang2024sea}. All models are trained on \textbf{a single RTX 3090 GPU}, with batch sizes of 6 and 4 for variants using ResNet-18 and ResNet-34. 

For generalization experiments on Sintel and KITTI 2015, \net{} follows the multi-stage training schedule outlined in \cite{wang2024sea}, again on a single RTX 3090 GPU. First, the models are trained on TartanAir \cite{wang2020tartanair} for 300K steps, maintaining the batch size as mentioned above. Then, they undergo a second stage on FlyingChairs \cite{dosovitskiy2015flownet} (C), with batch sizes of 8 and 6 for the two ResNet variants, followed by a third stage on FlyingThings3D 
\cite{mayer2016large} (T) for 120K steps, with batch 4 and 2. 
Finally, for experiments on Spring and LayeredFlow, we further fine-tune for 300K steps on a mixture of FlyingThings3D \cite{mayer2016large}, Sintel \cite{butler2012naturalistic}, KITTI \cite{menze2015object} and HD1K \cite{kondermann2016hci} (TSKH), with batch sizes 4 and 2.

\subsection{Ablation Studies and Analysis}
\label{sec:ablation}

We start our study by evaluating the effectiveness of different design choices for implementing \net. Table \ref{tab:ablations} collects the outcome of this analysis carried out on TartanAir \cite{wang2020tartanair} and KITTI 2012 \cite{geiger2013vision}, with the original SEAR-RAFT models being reported at the very top as a reference.

\textbf{\textcolor{purple}{Prior Combinations.}} The first set of experiments aim to assess the impact of the different priors provided by the depth foundation model to \net{} (T). Using either features $\mathbf{\Phi}_{0,1}$ or sending $\mathbf{D}_{0,1}$ to the \texttt{ContextNet} alone improves performance over the SEA-RAFT (S) baseline, while combining the two further decreased the error on TartanAir at the expenses of generalization. However, the $\texttt{BaseNet}$ alone yields consistently better results, proving to be the core component for optimally exploiting the depth foundation model, although with a noticeable increase in complexity. Finally, combining 
the $\texttt{BaseNet}$ with $\mathbf{\Phi}_{0,1}$ yields the absolute best results -- this configuration, highlighted in yellow, will be used from now on -- while combining the three priors slightly decreases accuracy.

\textbf{\textcolor{orange}{Model Size.}} By playing with both the ResNet type and the $\texttt{iters}$ parameter, we can implement different trade-offs between accuracy and complexity. Switching to ResNet-34 and replacing Depth Anything v2 (S) with the \textit{base} model (B) improves results on TartanAir, with minor drops on KITTI, while running more iterations consistently yields better performance. Notably, every \net{} variant outperforms its baseline counterpart reported at the top -- our (T) vs (S), our (S) vs (S*), and so on.

\textbf{\textcolor{teal}{Depth Foundation Models.}} To demonstrate the generality of our design scheme, we train \net{} (T) variants using different depth foundation models, including DPT \cite{Ranftl2021} and Depth Anything v1 \cite{yang2024depth}. We can observe that all variants substantially outperform the SEA-RAFT (S) baseline, with accuracy increasing when switching to newer models such as those in the Depth Anything series \cite{yang2024depth,yang2025depth}.
We speculate that future, more accurate foundation models could further enhance \net{} performance.

\textbf{\textcolor{blue}{Optical Flow Backbones.}} We further prove the generality of our approach by empowering two different flow backbones, respectively CRAFT \cite{sui2022craft} and FlowFormer \cite{huang2022flowformer}, with the priors extracted by a $\texttt{BaseNet}$. This addition largely improves the accuracy of both baseline model.

\textbf{Input to the \texttt{BaseNet}.} Finally, we compare the performance of FlowSeek (T) when replacing the motion bases with the original depth map as input to the 
\texttt{BaseNet}. Tab. \ref{tab:depth_vs_bases} shows how the bases yield improvements on both TartanAir and KITTI 2012. These datasets, however, mostly contain ego-motion induced by the camera. We further fine-tune both models for 10K steps on the first 160 images of the KITTI 2015 training set and evaluate on the remaining 40. We can notice a consistent improvement in static regions, at the price of a drop on moving objects.

\subsection{Zero-Shot Generalization}

We now assess the ability of \net{} to generalize across datasets compared to SEA-RAFT \cite{wang2024sea} and other methods. From now on, we will highlight the \colorbox{First}{\textbf{absolute}}, the \colorbox{Second}{second}, and the \colorbox{Third}{third} best methods in each table among those trained following the protocol defined by SEA-RAFT.

\textbf{Sintel and KITTI 2015.} Table \ref{tab:SK-train} collects results achieved on Sintel and KITTI 2015 training sets, following the evaluation protocol established by SEA-RAFT \cite{wang2024sea}. Models reported at the top were trained on ``C+T" only, while at the bottom we show the results achieved when pre-training on TartanAir is performed. For SEA-RAFT, we report both the results by the authors, as well as those reproduced by retraining on a single GPU to highlight the impact that the hardware budget has on final accuracy -- dramatically dropping on KITTI, while improving on Sintel \textit{Final}. 

Among the former category, \net{} (T) already outperforms most existing methods, including the original SEA-RAFT (S) against which it marks a consistent margin on both datasets despite the disparity in hardware budget. 
\net{} (S) further improves over (T), while \net{} (M) and (L) outperform all competitors on Sintel Final, though losing some accuracy on KITTI as previously observed in our ablation studies. We attribute this to the higher complexity of (M) and (L) variants, whose training is likely constrained on a single GPU without strong pre-training, making \net{} (S) as the best solution under this setting.

When pre-training on TartanAir, we observe moderate improvements in the SEA-RAFT models, with much larger gains achieved by \net{} variants. In particular, \net{} (L) achieves the overall best results across Sintel Final and KITTI 2015, despite the moderate hardware budget used for training -- resulting in a batch size $8\times$ smaller than its SEA-RAFT (L) counterpart.
As a reference, we also report the results achieved by DDVM \cite{saxena2023surprising}: when only AutoFlow (AF) \cite{sun2021autoflow} and T are used (a setting similar to ours), \net{} outperforms it, whereas DDVM achieves superior results when also using Kubric (KU) and TartanAir.

Figures \ref{fig:sintel} and \ref{fig:kitti} showcase qualitative results by SEA-RAFT (L) and \net{} (L) on Sintel and KITTI. 

\begin{table}[t]
    \centering
    \resizebox{1.0\linewidth}{!}{
    \renewcommand{\tabcolsep}{8pt}
    \begin{tabular}{llrrrrrr}
    \toprule
    \multirow{2}{*}{Extra Data} & \multirow{2}{*}{Method} &\multicolumn{2}{c}{Spring (train)} \\
    \cmidrule(l{0.5ex}r{0.5ex}){3-4} 
    & 
    & 1px$\downarrow$ & EPE$\downarrow$ 
    \\ 
        \midrule 
        & RAFT~\cite{teed2020raft}  & 4.788 & 0.448 \\ 
        & GMA~\cite{jiang2021learning}  & 4.763 & 0.443 \\ 
        & RPKNet~\cite{morimitsu2024recurrent}  & 4.472 & 0.416 \\ 
        & DIP~\cite{zheng2022dip}  & 4.273 & 0.463 \\ 
        & SKFlow~\cite{sun2022skflow}  & 4.521 & 0.408 \\ 
        & GMFlow~\cite{xu2022gmflow}  & 29.49 & 0.930 \\ 
        & GMFlow$+$~\cite{xu2023unifying}  & 4.292 & 0.433 \\ 
        & Flowformer~\cite{huang2022flowformer}  & 4.508 & 0.470 & \\ 
        & CRAFT~\cite{sui2022craft}  & 4.803 & 0.448 \\ 
        \midrule 
        \multirow{8}{*}{Tartan}
        & {SEA-RAFT (S)}  & 4.161 & 0.410 \\
        & {SEA-RAFT (M)} & \trd{3.888} & \snd{0.406} \\ 
        & {SEA-RAFT (L)} & \snd 3.842 & 0.426 \\ 
        & \bf \net{} (T)  & 4.111 & 0.410 \\ 
        & \bf \net{} (S)  & 4.058 & \snd 0.406 \\ 
        & \bf \net{} (M)  & 3.941 & 0.419 \\ 
        & \bf \net{} (L)  & \fst 3.838 & \fst 0.402 \\
        \midrule
        \midrule
        MegaDepth~\cite{li2018megadepth} & MatchFlow(G)~\cite{dong2023rethinking}  & 4.504 & 0.407 \\ 
        YT-VOS~\cite{xu2018youtube}& Flowformer$++$\cite{shi2023flowformer++} & 4.482 & 0.447 \\ 
        VIPER~\cite{richter2017playing}& MS-
        RAFT+~\cite{jahedi2024ms}  & {3.577} & {0.397} \\ 
    \bottomrule
    \end{tabular}
    }\vspace{-0.3cm}
    \caption{
    \textbf{Zero-Shot Generalization -- Spring.} Methods on top are trained with ``C $\rightarrow$ T $\rightarrow$ TSKH" schedule. }
    \label{tab:spring}
    \vspace{-0.3cm}
\end{table}

\textbf{Spring.} We continue our investigation on the Spring dataset \cite{mehl2023spring}, evaluating SEA-RAFT and \net{} variants after further training on ``TSKH", by dowsampling input images by a factor $2\times$ as in \cite{wang2024sea}.
Table \ref{tab:spring} collects the outcome of this evaluation. Generally, the domain gap from ``C$\rightarrow$T$\rightarrow$TSKH" to Spring is much lower compared to what occurs from ``C$\rightarrow$T" to Sintel and KITTI, with EPE values falling below 0.5. Nevertheless, we observe that \net{} (L) beats any SEA-RAFT model, despite being trained on a single GPU compared to the multiple GPUs used by its counterparts. Only MS-RAFT+ performs slightly better, though it uses 2$\times$ A100 GPUs and VIPER \cite{richter2017playing} data.

\textbf{LayeredFlow.} We conclude our evaluation by conducting a further zero-shot evaluation experiment on the recent LayeredFlow dataset \cite{wen2024layeredflow}, which features challenging transparent and reflective surfaces. Following \cite{wen2024layeredflow}, we evaluate both SEA-RAFT and \net{} ``C$\rightarrow$T$\rightarrow$TSKH" models on the validation set by down-sampling images to $\frac{1}{8}$ of their original resolution and evaluating the predictions on the \textit{first Layer} -- i.e., the surfaces closest to the camera. 
Table \ref{tab:optical_flow_first_layer} presents the results achieved by representative optical flow architectures, as well as those involved in our experiments at the bottom.
First and foremost, we note that every \net{} variant outperforms its SEA-RAFT counterpart, often by substantial margins -- e.g., \net{} (L) improves EPE on \textit{All} category by more than 2 pixels. Moreover, when compared to other existing methods, \net{} (M) and (L) achieve consistently lower errors on \textit{All} pixels. 
In the interest of space, we refer the reader to the supplementary material for detailed results on single categories -- classifying image regions into \textit{Transparent}, \textit{Reflective} and \textit{Diffuse}.

Figure \ref{fig:layeredflow} shows qualitative comparisons between the predictions by SEA-RAFT (L) and \net{} (L), with the latter recovering finer details and exhibiting fewer artifacts.

\textbf{Qualitative Results.} We refer the reader to the supplementary material for additional qualitative samples concerning the experiments discussed so far.

\begin{table}[t]
  \centering
  \renewcommand{\tabcolsep}{10pt}
  \resizebox{\linewidth}{!}{
  \begin{tabular}{l r rrrr r rrrr r rrrr r rrrr}
    \toprule
    \multirow{2}{*}{Method} && \multicolumn{4}{c}{All} 
    \\
    \cmidrule{2-21}
    && EPE$\downarrow$ & 1px$\downarrow$ & 3px$\downarrow$ & 5px$\downarrow$ 
    \\
    \midrule
    FlowNet-C \cite{dosovitskiy2015flownet} && 9.71 & 89.07 & 61.51 & 43.93 \\
    FlowNet2 \cite{ilg2017flownet2} && 10.07 & 77.56 & 54.22 & 42.13 \\
    PWC-Net \cite{sun2018pwc} && 9.49 & 74.93 & 50.47 & 39.05 \\
    GMA \cite{jiang2021learning} && 9.77 & 72.46 & 46.93 & 36.97 \\
    SKFlow \cite{sun2022skflow} && 9.86 & 72.02 & 47.44 & 36.88 \\
    CRAFT \cite{sui2022craft} && 10.36 & 72.34 & 47.54 & 37.00 \\
    GMFlow \cite{xu2022gmflow} && \trd 9.09 & 81.99 & 51.79 & 37.75 \\
    GMFlow+ \cite{xu2023unifying} && 9.46 & 82.71 & 53.14 & 39.70 \\
    FlowFormer \cite{huang2022flowformer} && 10.20 & 73.59 & 48.97 & 38.56\\ 
    RAFT \cite{teed2020raft} && 9.38 & 71.98 & 46.46 & 36.15 \\
    \midrule 
    SEA-RAFT (S) && 10.05 & 71.48 & 46.90 & 36.32 \\
    SEA-RAFT (M) && 10.17 & 69.73 & 45.94 & 34.78 \\
    SEA-RAFT (L) && 10.99 & \trd 69.46 & 45.59 & 34.78 \\
    \bf \net{} (T) && 
    \trd 9.09  & 70.82  & 43.74 & 32.36 \\
    \bf \net{} (S) && 
    9.16 & 69.99 & \trd 43.67 & \snd 31.90 \\
    \bf \net{} (M) && 
    \fst 8.30 & \fst 68.85 & \snd 41.81 & \trd 32.09 \\
    \bf \net{} (L) && 
    \fst 8.30 & \snd 68.98 & \fst 41.49 & \fst 31.64 \\
    
    \bottomrule
  \end{tabular}
  }
  \vspace{-3mm}
  \caption{\textbf{Zero-shot Generalization -- LayeredFlow (train) first layer evaluation.} Images are down-sampled by a factor 8.}\vspace{-0.3cm}
  \label{tab:optical_flow_first_layer}
\end{table}

\begin{figure}[t]
    \centering
    \renewcommand{\tabcolsep}{1pt}
    \begin{tabular}{ccc}
    
    \includegraphics[width=0.155\textwidth,frame]{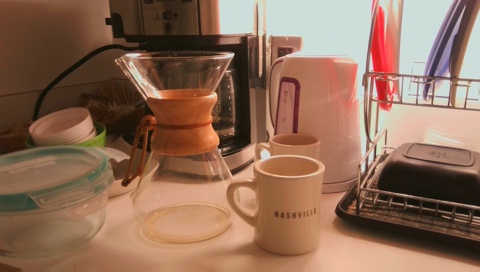} &
    \includegraphics[width=0.155\textwidth,frame]{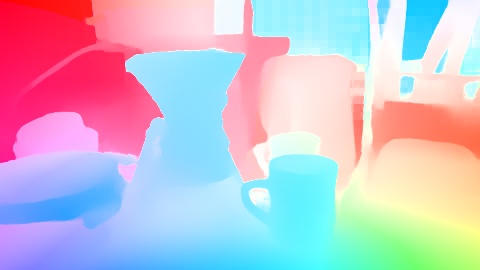} &
    \includegraphics[width=0.155\textwidth,frame]{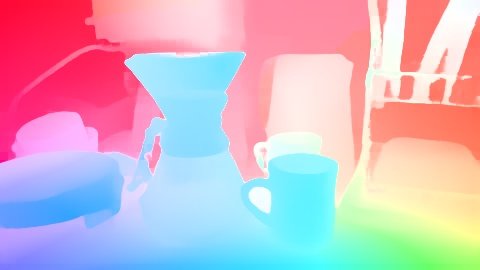} \\

    \includegraphics[width=0.155\textwidth,frame]{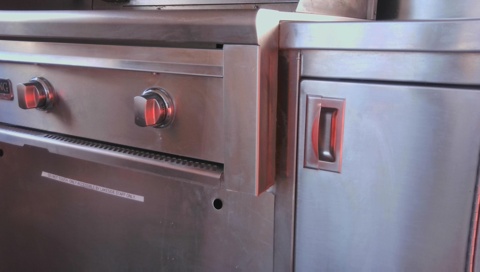} &
    \includegraphics[width=0.155\textwidth,frame]{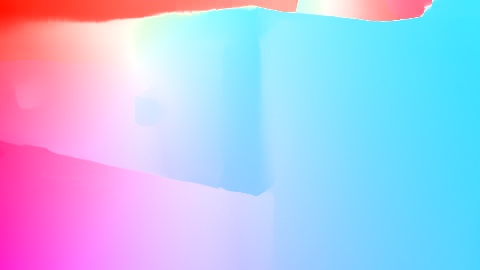} &
    \includegraphics[width=0.155\textwidth,frame]{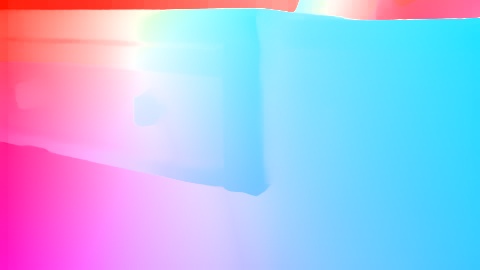} \\
    
    \small $\mathbf{I}_0$ & \small SEA-RAFT (L) \cite{wang2024sea} & \small \bf \net{} (L) \\

    \end{tabular}\vspace{-0.3cm}
    \caption{\textbf{Qualitative Results on LayeredFlow \cite{wen2024layeredflow}.} From left to right: first frame, flow by SEA-RAFT (L) and \net{} (L).}\vspace{-0.4cm}
    \label{fig:layeredflow}
\end{figure}

\section{Conclusion}

We presented \net{}, a novel architecture for optical flow that combines the latest architectural developments in the field with cutting-edge depth foundation models and low-dimensional flow parametrization from classical computer vision. The synergy of the three makes \net{} a practical model capable of achieving state-of-the-art zero-shot generalization, even when trained with as few as a single consumer-grade GPU. 
We believe \net{} can inspire further attempts to design new models trainable with minimal hardware budget in adjacent fields of computer vision. 

\textbf{Limitations.} The possibility of training \net{} comes from the availability of large, pre-trained foundation models which were likely trained with much higher hardware budget on web-scale data. However, we hope our work encourages the community to avoid training new architectures from scratch at prohibitive costs and instead reuse existing models when possible. 

\textbf{Future Work.} Training data remains another significant bottleneck for flow models. Future research will focus on this aspect, attempting to emulate the successful strategies in the depth estimation literature \cite{yang2024depth,yang2025depth}.

\small\textbf{Acknowledgment.} We thank Sadra Safadoust and Fatma Güney for the insightful discussion about motion bases.

{
    \small
    \bibliographystyle{ieeenat_fullname}
    \bibliography{main,main_old}
}

\twocolumn[{
\renewcommand\twocolumn[1][]{#1}
\maketitlesupplementary
\begin{center} 
    \renewcommand{\tabcolsep}{1pt}
    \centering
    \begin{tabular}{cccccc}

        \begin{overpic}[width=0.155\linewidth,frame]{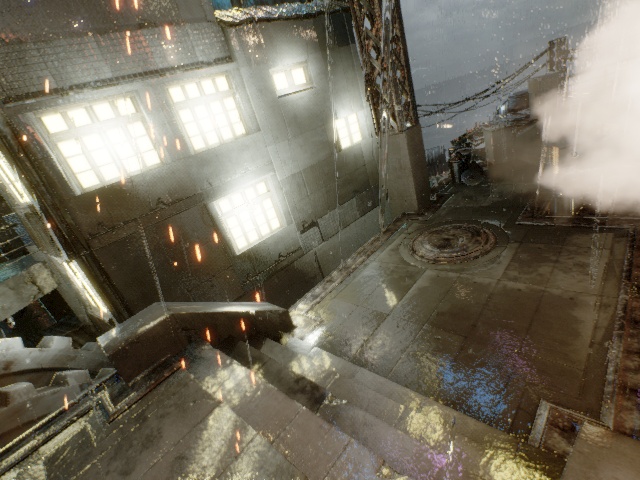}
        \end{overpic} & 
        \begin{overpic}[width=0.155\linewidth,frame]{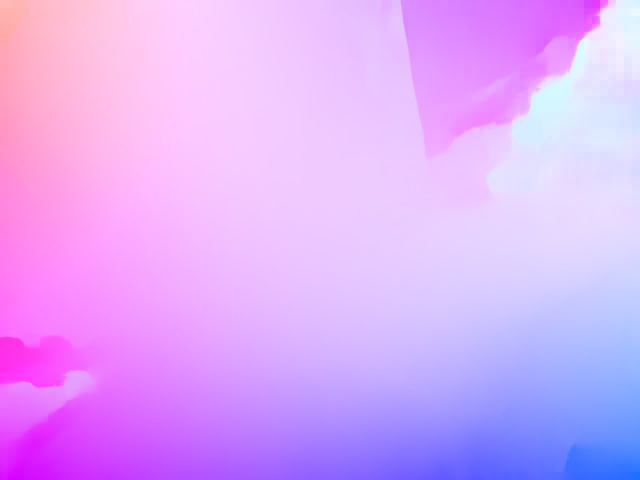}
        \put (2,65) {\footnotesize \textcolor{purple}{\textbf{\texttt{EPE: 0.487 }}}}
        \end{overpic} & 
        \begin{overpic}[width=0.155\linewidth,frame]{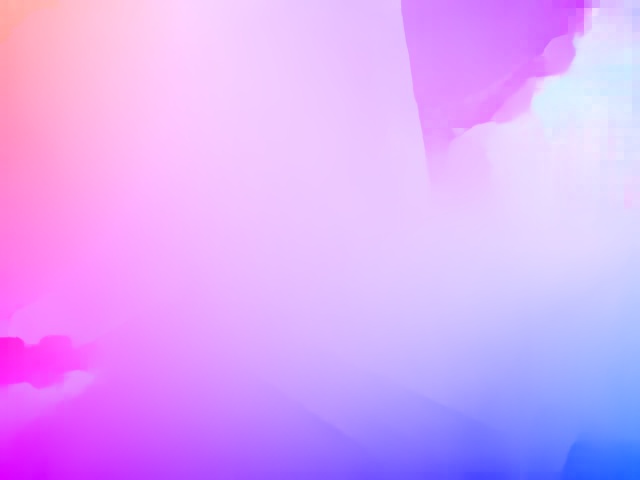}
        \put (2,65) {\footnotesize \textcolor{purple}{\textbf{\texttt{EPE: 0.459 }}}}
        \end{overpic} &
        \begin{overpic}[width=0.155\linewidth,frame]{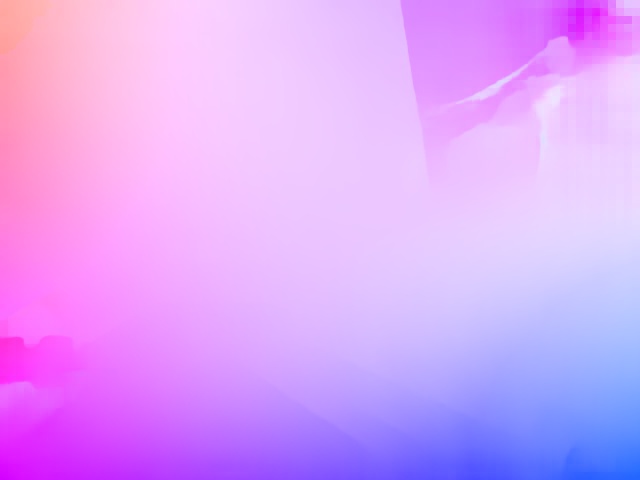}
        \put (2,65) {\footnotesize \textcolor{purple}{\textbf{\texttt{EPE: 0.290 }}}}
        \end{overpic} &
        \begin{overpic}[width=0.155\linewidth,frame]{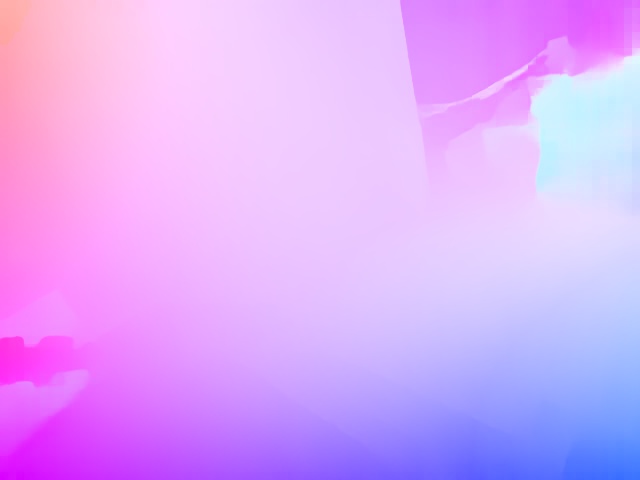}
        \put (2,65) {\footnotesize \textcolor{purple}{\textbf{\texttt{EPE: 0.406 }}}}
        \end{overpic} & 
        \begin{overpic}[width=0.155\linewidth,frame]{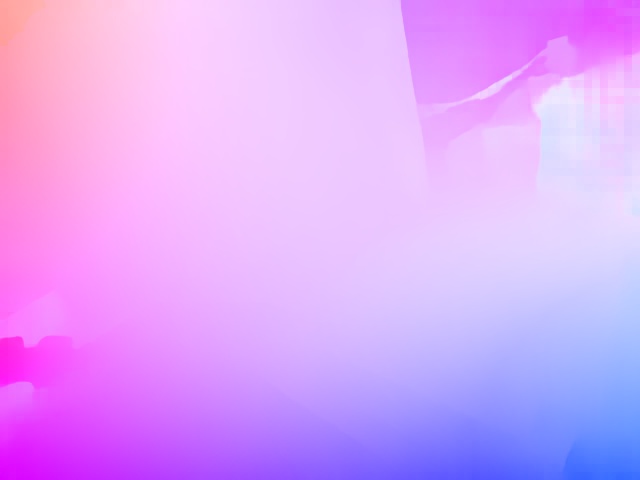}
        \put (2,65) {\footnotesize \textcolor{purple}{\textbf{\texttt{EPE: 0.272 }}}}
        \end{overpic} \\

        \begin{overpic}[width=0.155\linewidth,frame]{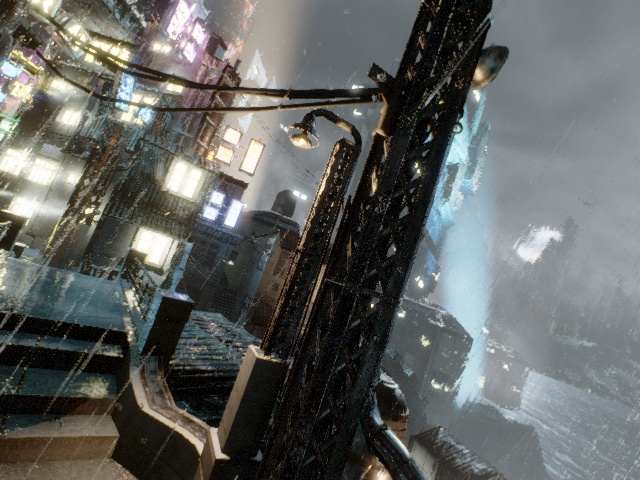}
        \end{overpic} & 
        \begin{overpic}[width=0.155\linewidth,frame]{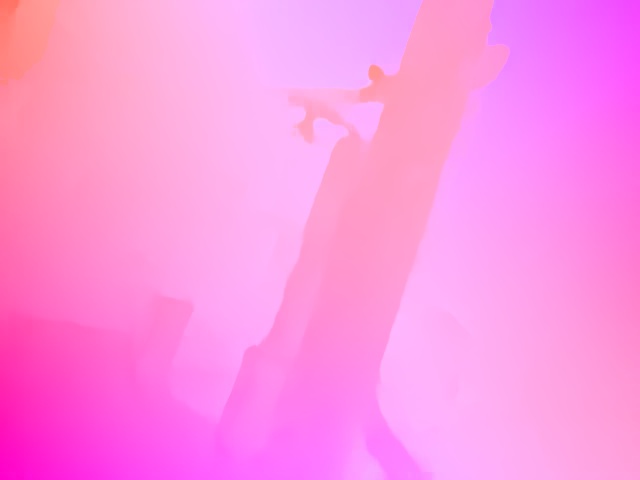}
        \put (2,65) {\footnotesize \textcolor{purple}{\textbf{\texttt{EPE: 0.527 }}}}
        \end{overpic} & 
        \begin{overpic}[width=0.155\linewidth,frame]{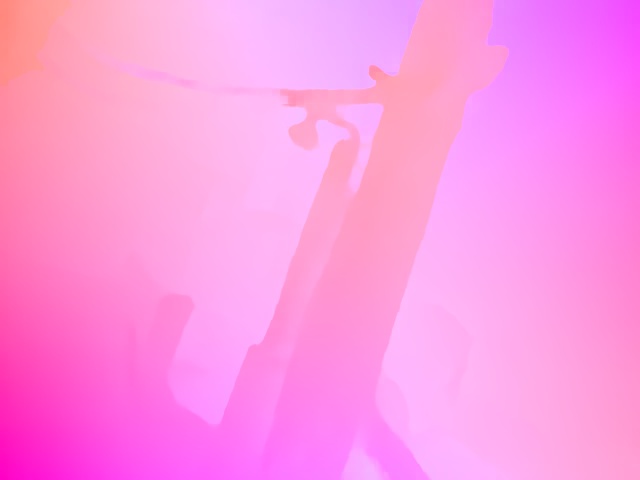}
        \put (2,65) {\footnotesize \textcolor{purple}{\textbf{\texttt{EPE: 0.520 }}}}
        \end{overpic} &
        \begin{overpic}[width=0.155\linewidth,frame]{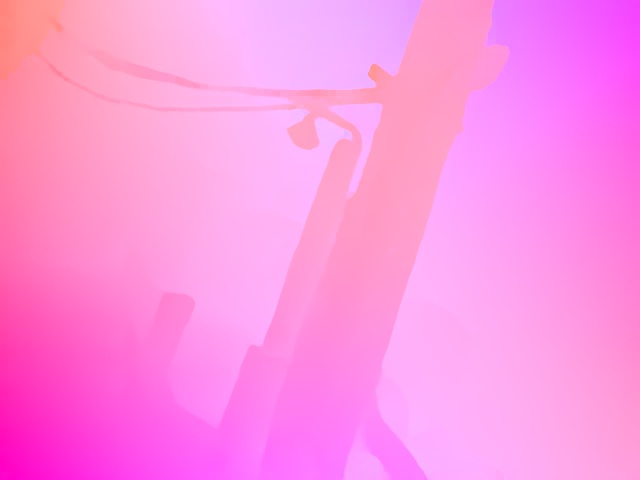}
        \put (2,65) {\footnotesize \textcolor{purple}{\textbf{\texttt{EPE: 0.496 }}}}
        \end{overpic} & 
        \begin{overpic}[width=0.155\linewidth,frame]{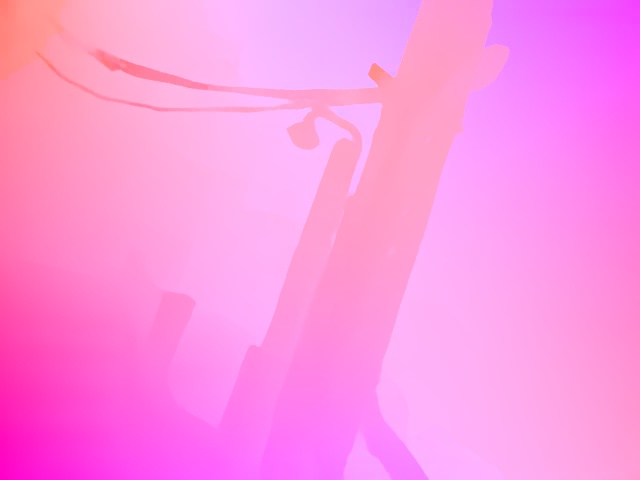}
        \put (2,65) {\footnotesize \textcolor{purple}{\textbf{\texttt{EPE: 0.409 }}}}
        \end{overpic} & 
        \begin{overpic}[width=0.155\linewidth,frame]{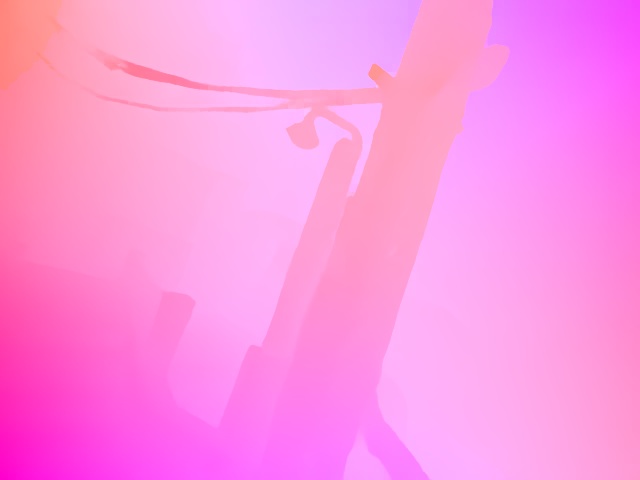}
        \put (2,65) {\footnotesize \textcolor{purple}{\textbf{\texttt{EPE: 0.380 }}}}
        \end{overpic} 
        \\

        \begin{overpic}[width=0.155\linewidth,frame]{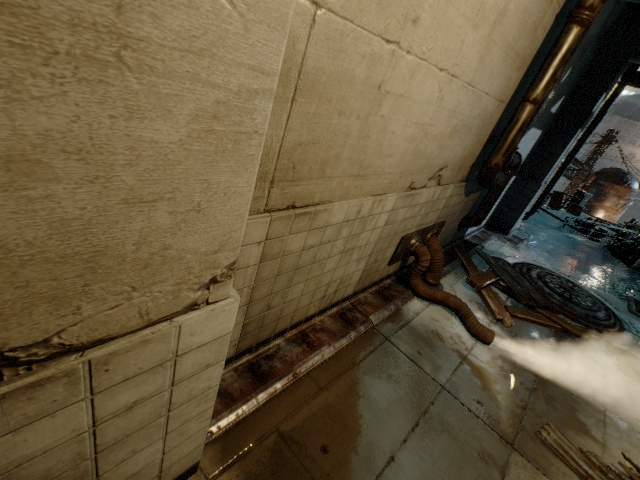}
        \end{overpic} & 
        \begin{overpic}[width=0.155\linewidth,frame]{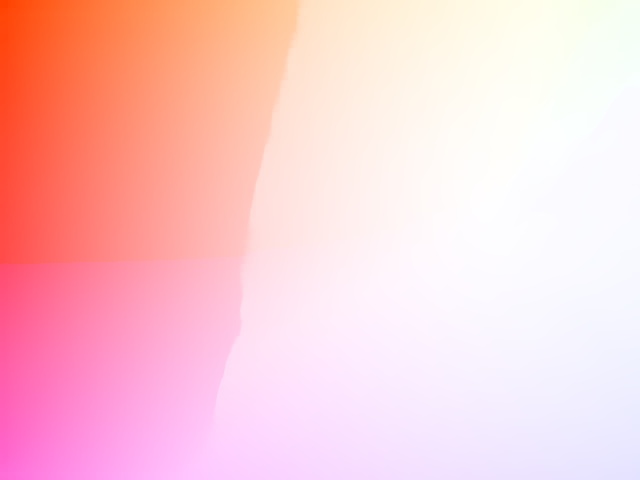}
        \put (2,65) {\footnotesize \textcolor{purple}{\textbf{\texttt{EPE: 0.327 }}}}
        \end{overpic} & 
        \begin{overpic}[width=0.155\linewidth,frame]{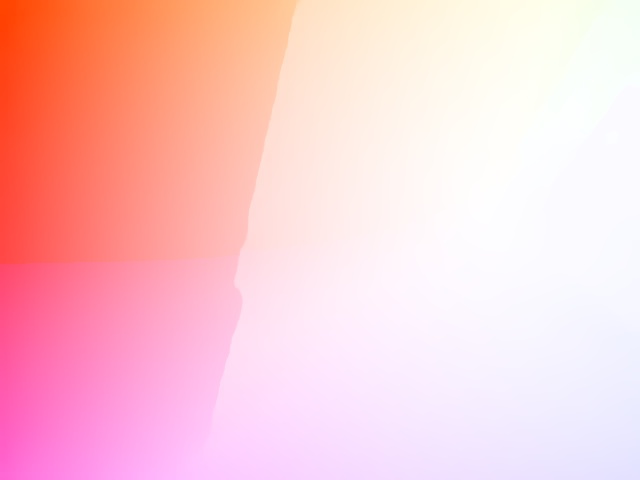}
        \put (2,65) {\footnotesize \textcolor{purple}{\textbf{\texttt{EPE: 0.277 }}}}
        \end{overpic} &
        \begin{overpic}[width=0.155\linewidth,frame]{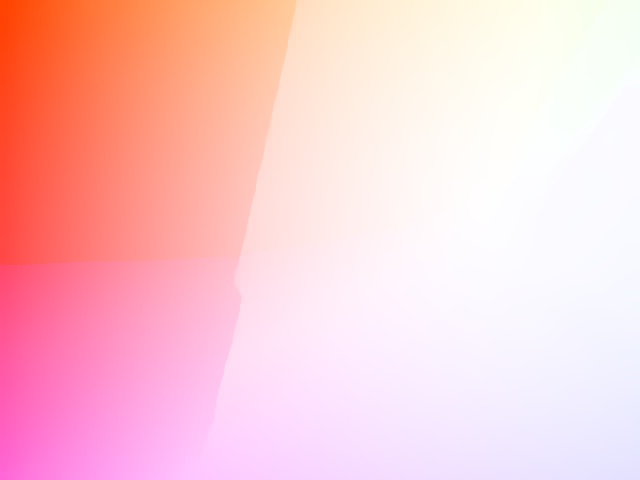}
        \put (2,65) {\footnotesize \textcolor{purple}{\textbf{\texttt{EPE: 0.257 }}}}
        \end{overpic} & 
        \begin{overpic}[width=0.155\linewidth,frame]{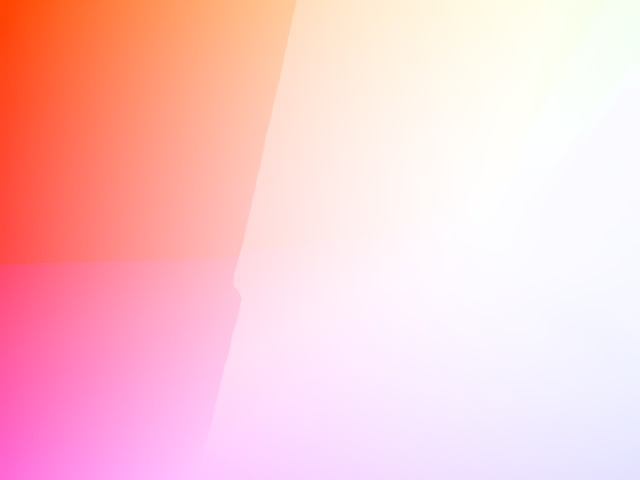}
        \put (2,65) {\footnotesize \textcolor{purple}{\textbf{\texttt{EPE: 0.240 }}}}
        \end{overpic} & 
        \begin{overpic}[width=0.155\linewidth,frame]{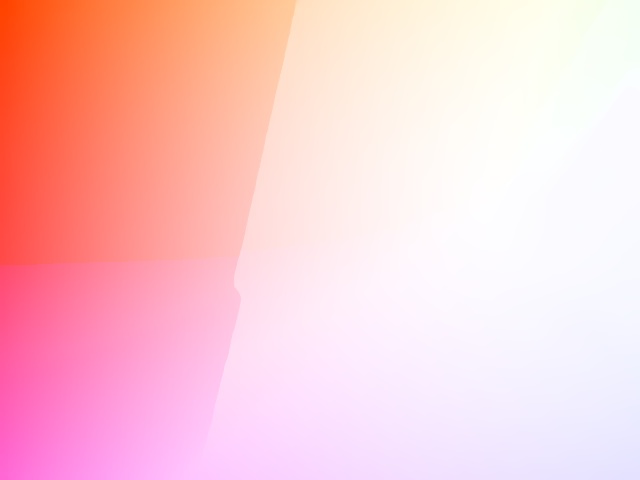}
        \put (2,65) {\footnotesize \textcolor{purple}{\textbf{\texttt{EPE: 0.226 }}}}
        \end{overpic} \\

        \begin{overpic}[width=0.155\linewidth,frame]{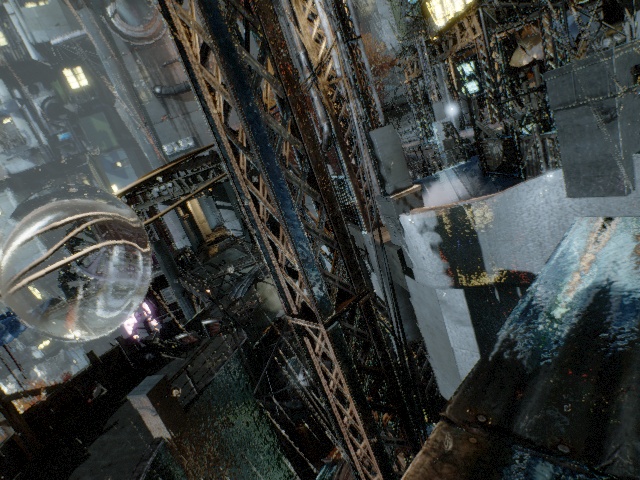}
        \end{overpic} & 
        \begin{overpic}[width=0.155\linewidth,frame]{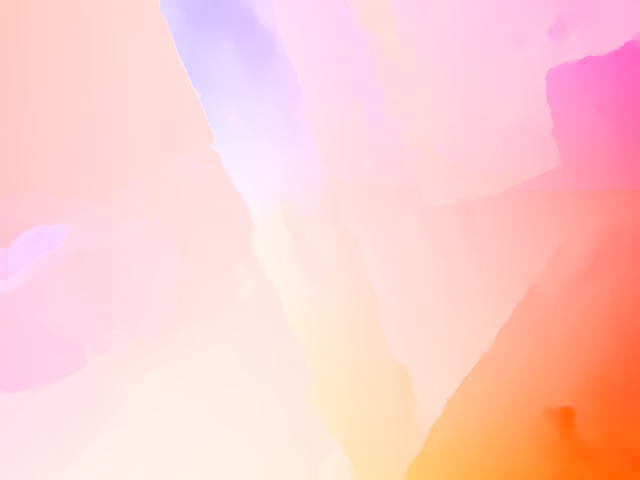}
        \put (2,65) {\footnotesize \textcolor{purple}{\textbf{\texttt{EPE: 1.281 }}}}
        \end{overpic} & 
        \begin{overpic}[width=0.155\linewidth,frame]{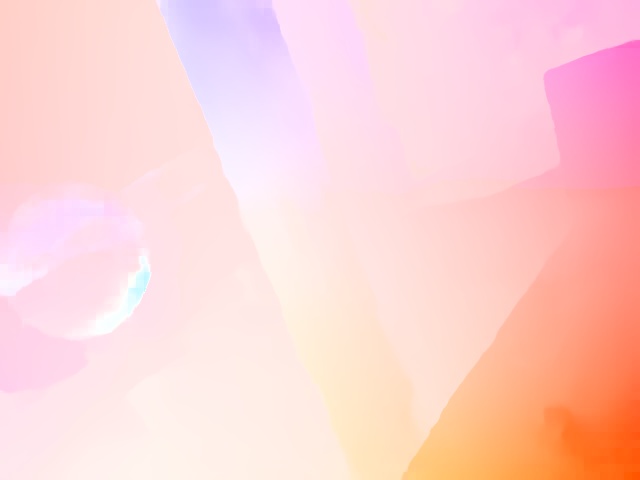}
        \put (2,65) {\footnotesize \textcolor{purple}{\textbf{\texttt{EPE: 1.336 }}}}
        \end{overpic} &
        \begin{overpic}[width=0.155\linewidth,frame]{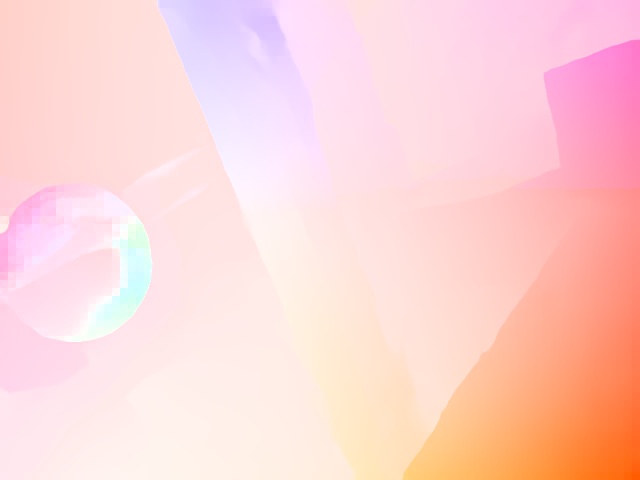}
        \put (2,65) {\footnotesize \textcolor{purple}{\textbf{\texttt{EPE: 1.435 }}}}
        \end{overpic} & 
        \begin{overpic}[width=0.155\linewidth,frame]{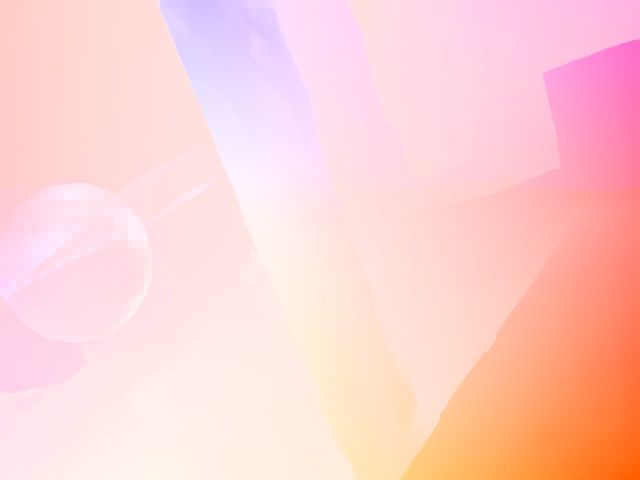}
        \put (2,65) {\footnotesize \textcolor{purple}{\textbf{\texttt{EPE: 1.129 }}}}
        \end{overpic} & 
        \begin{overpic}[width=0.155\linewidth,frame]{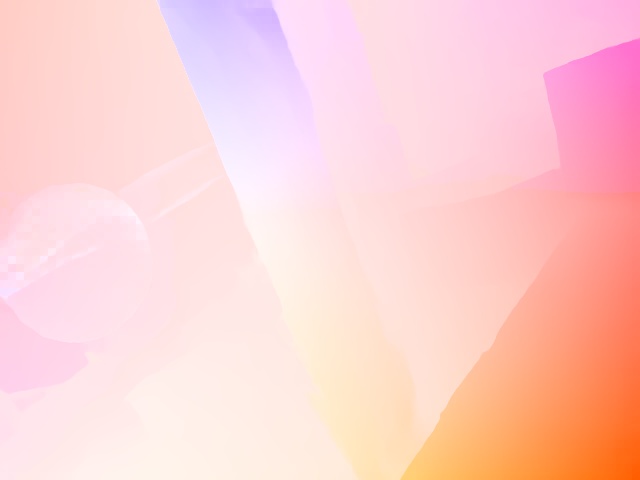}
        \put (2,65) {\footnotesize \textcolor{purple}{\textbf{\texttt{EPE: 1.112 }}}}
        \end{overpic} \\

        \begin{overpic}[width=0.155\linewidth,frame]{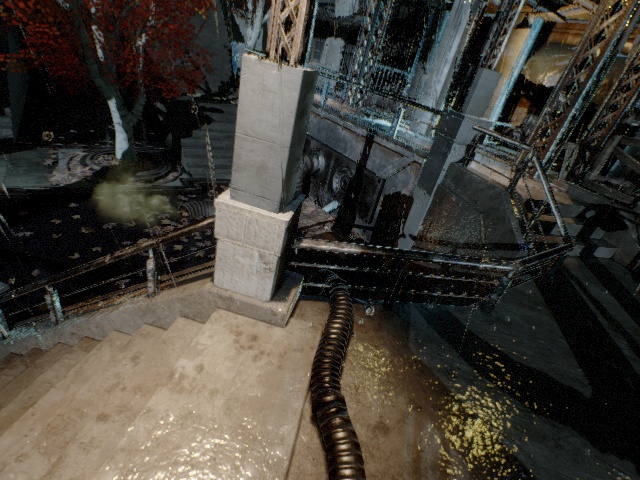}
        \end{overpic} & 
        \begin{overpic}[width=0.155\linewidth,frame]{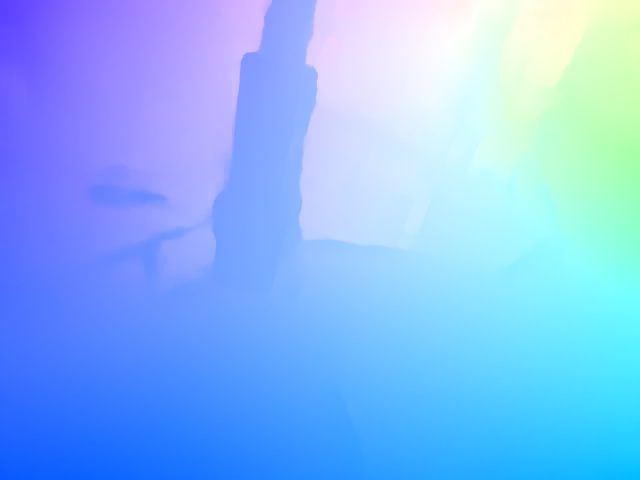}
        \put (2,65) {\footnotesize \textcolor{purple}{\textbf{\texttt{EPE: 0.416 }}}}
        \end{overpic} & 
        \begin{overpic}[width=0.155\linewidth,frame]{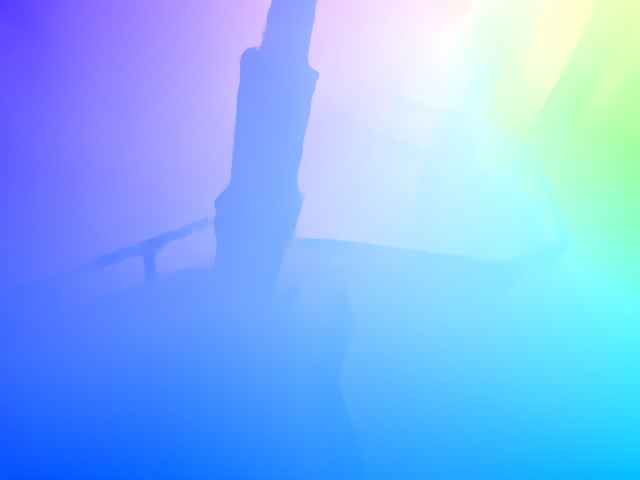}
        \put (2,65) {\footnotesize \textcolor{purple}{\textbf{\texttt{EPE: 0.369 }}}}
        \end{overpic} &
        \begin{overpic}[width=0.155\linewidth,frame]{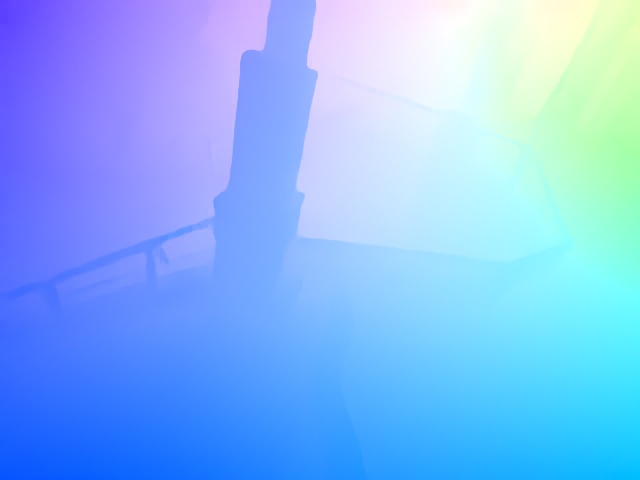}
        \put (2,65) {\footnotesize \textcolor{purple}{\textbf{\texttt{EPE: 0.336 }}}}
        \end{overpic} & 
        \begin{overpic}[width=0.155\linewidth,frame]{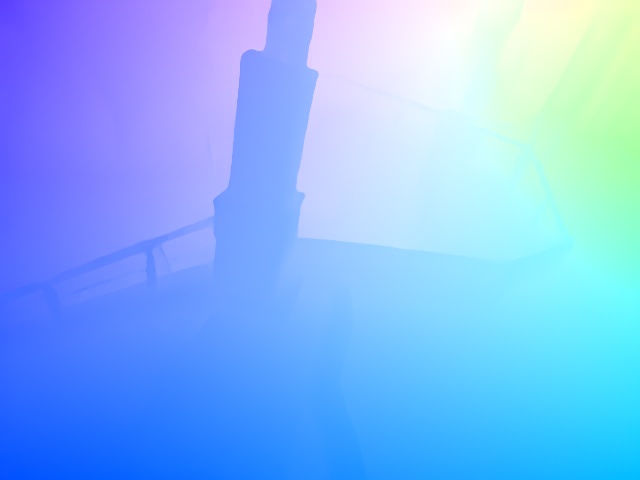}
        \put (2,65) {\footnotesize \textcolor{purple}{\textbf{\texttt{EPE: 0.330 }}}}
        \end{overpic} & 
        \begin{overpic}[width=0.155\linewidth,frame]{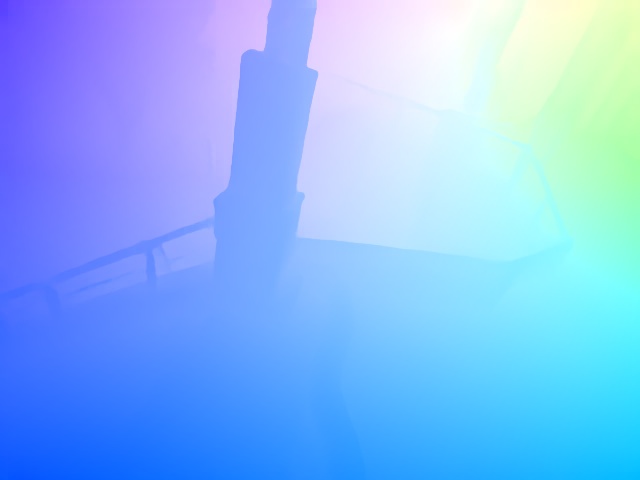}
        \put (2,65) {\footnotesize \textcolor{purple}{\textbf{\texttt{EPE: 0.328 }}}}
        \end{overpic} \\

        \begin{overpic}[width=0.155\linewidth,frame]{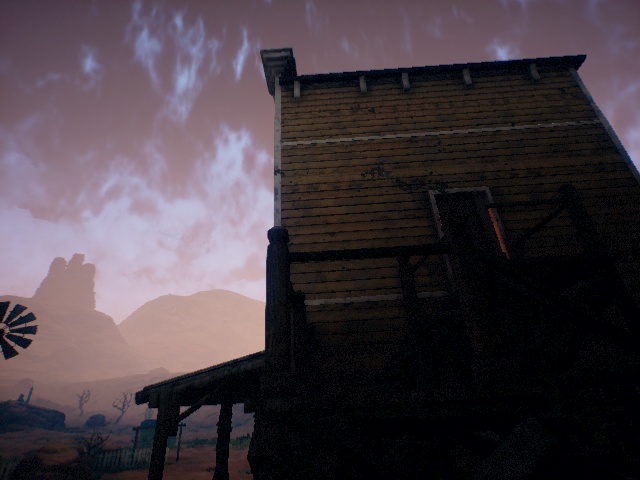}
        \end{overpic} & 
        \begin{overpic}[width=0.155\linewidth,frame]{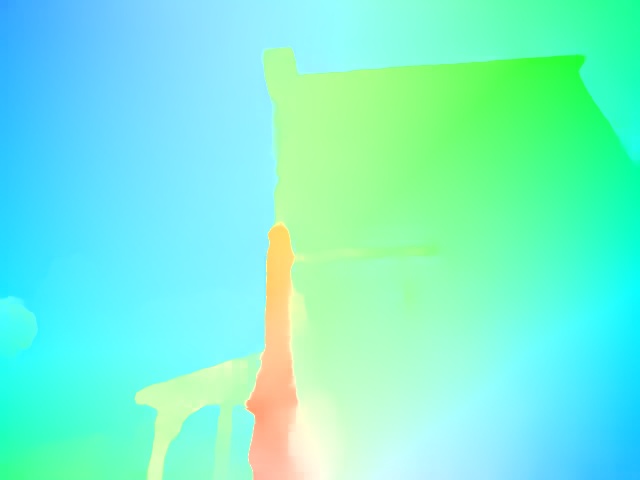}
        \put (2,65) {\footnotesize \textcolor{purple}{\textbf{\texttt{EPE: 0.454 }}}}
        \end{overpic} & 
        \begin{overpic}[width=0.155\linewidth,frame]{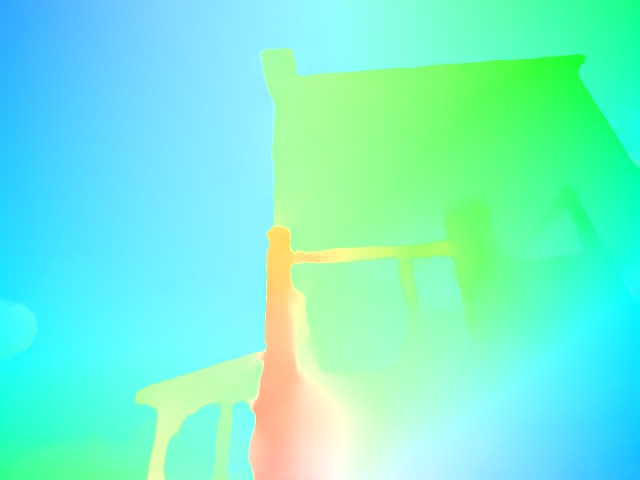}
        \put (2,65) {\footnotesize \textcolor{purple}{\textbf{\texttt{EPE: 0.432 }}}}
        \end{overpic} &
        \begin{overpic}[width=0.155\linewidth,frame]{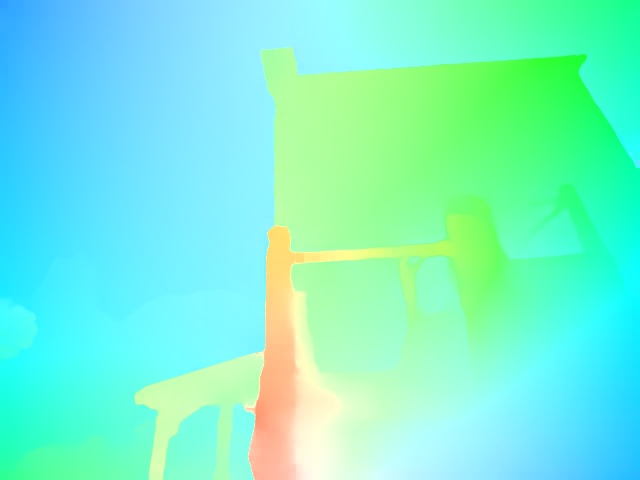}
        \put (2,65) {\footnotesize \textcolor{purple}{\textbf{\texttt{EPE: 0.408 }}}}
        \end{overpic} & 
        \begin{overpic}[width=0.155\linewidth,frame]{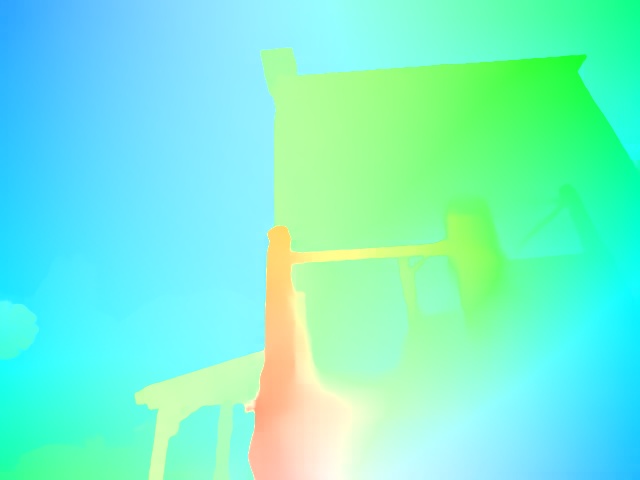}
        \put (2,65) {\footnotesize \textcolor{purple}{\textbf{\texttt{EPE: 0.383 }}}}
        \end{overpic} & 
        \begin{overpic}[width=0.155\linewidth,frame]{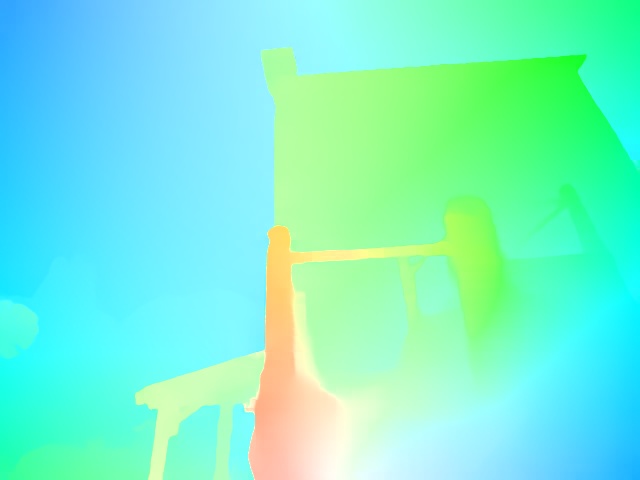}
        \put (2,65) {\footnotesize \textcolor{purple}{\textbf{\texttt{EPE: 0.377 }}}}
        \end{overpic} \vspace{-0.3cm}\\

        \footnotesize $\mathbf{I}_0$ & \footnotesize SEA-RAFT (S) & \footnotesize + $\mathbf{\Phi}_{0,1}$ & \footnotesize + $\mathbf{D}_{0,1}$ & \footnotesize + $\texttt{BaseNet}$ & \bf\footnotesize \net{} (T) \\
    \end{tabular}\vspace{-0.3cm}
    \captionof{figure}{\textbf{Qualitative results on TartanAir.} From left to right: first frame, flow predicted by SEA-RAFT (S) and ablated versions of \net{} (T).} 
    \label{fig:tartan}\vspace{-0.1cm}
\end{center}

\vspace{0.1cm}
\normalsize{In this document, we report additional qualitative results concerning the experiments reported in the main paper. 
Specifically, Figures \ref{fig:tartan} and \ref{fig:2012} provide a visual comparison between SEA-RAFT (S) and the different versions of \net{} (T) ablated in Table 1 of the main paper, respectively on TartanAir and KITTI 2012. Figures \ref{fig:sintel} and \ref{fig:2015} show further comparisons between SEA-RAFT (S) and \net{} (T), respectively on Sintel and KITTI 2015, while Figure \ref{fig:spring} reports qualitative results on Spring by the same models, trained following ``C$\rightarrow$T$\rightarrow$TSKH" schedule. Table \ref{tab:spring_gen} shows a further, quantitative comparison between SEA-RAFT and \net{} variants trained according to ``C$\rightarrow$T" schedule, highlighting how the former model suffers of generalization issues more severely with respect to ours. This is also shown qualitatively in Figure \ref{fig:spring_gen}, as well as in the teaser in the main paper.
Finally, Table \ref{tab:optical_flow_first_layer} reports detailed results on the LayeredFlow dataset, highlighting how \net{} achieves large improvements on most metrics for \textit{Transparent} and \textit{Reflective} regions thanks to the strong priors injected by the depth foundation model. This comes at the expense of some marginal drops on \textit{Diffuse} regions, mostly inherited from the baseline SEA-RAFT models. }\vspace{-1cm}
}
]

\begin{figure*}[t]
    \centering
    \renewcommand{\tabcolsep}{1pt}
    \begin{tabular}{ccc}

    \begin{overpic}[width=0.32\linewidth,frame]{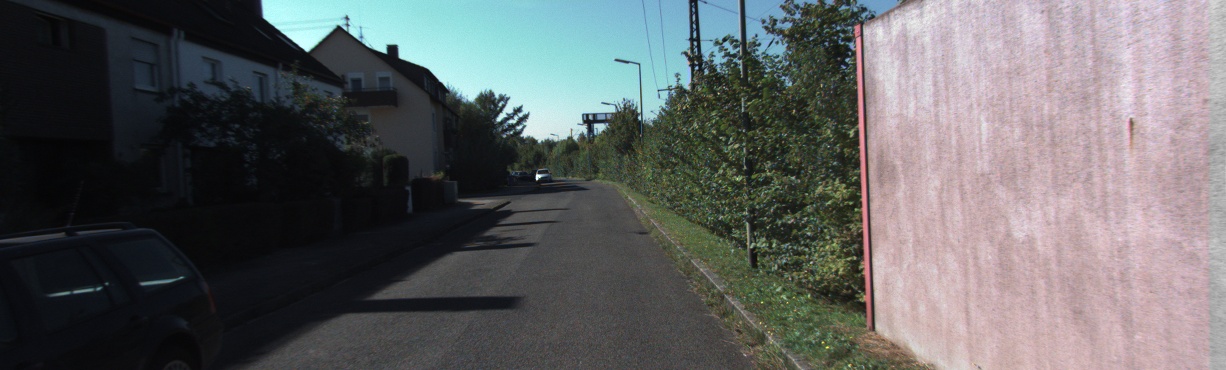}
    \end{overpic} & 
    \begin{overpic}[width=0.32\linewidth,frame]{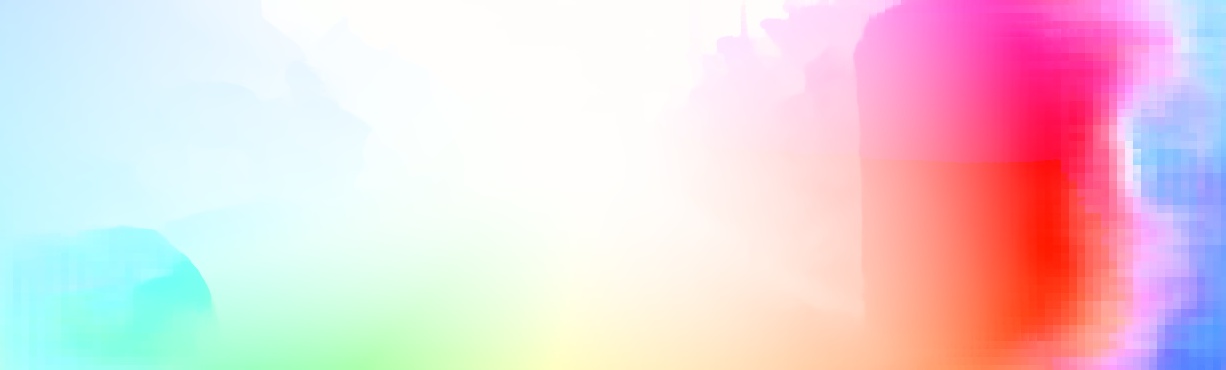}
    \put (2,25) {\footnotesize \textcolor{purple}{\textbf{\texttt{EPE: 11.746 }}}}
    \end{overpic} & 
    \begin{overpic}[width=0.32\linewidth,frame]{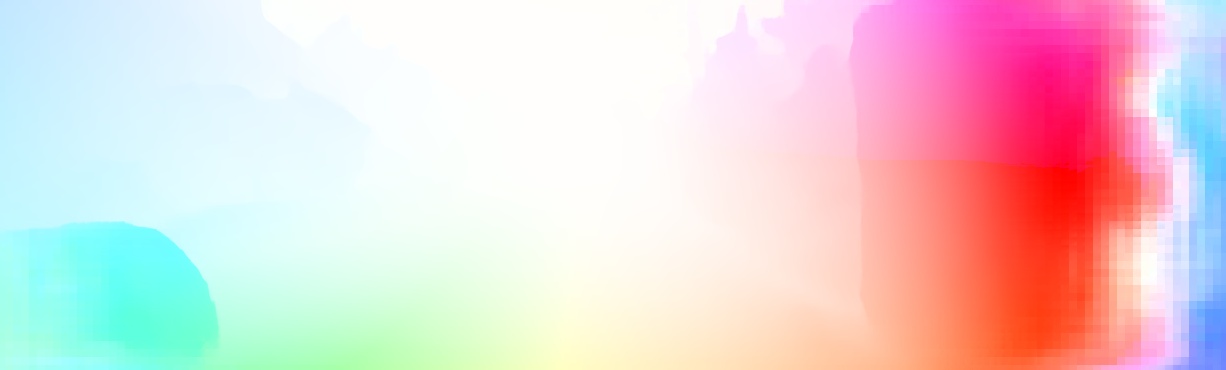}
    \put (2,25) {\footnotesize \textcolor{purple}{\textbf{\texttt{EPE: 10.255 }}}}
    \end{overpic} \vspace{-0.2cm}\\
    \footnotesize $\mathbf{I}_0$ & \footnotesize SEA-RAFT (S) & \footnotesize + $\mathbf{\Phi}_{0,1}$ \\ 
    \begin{overpic}[width=0.32\linewidth,frame]{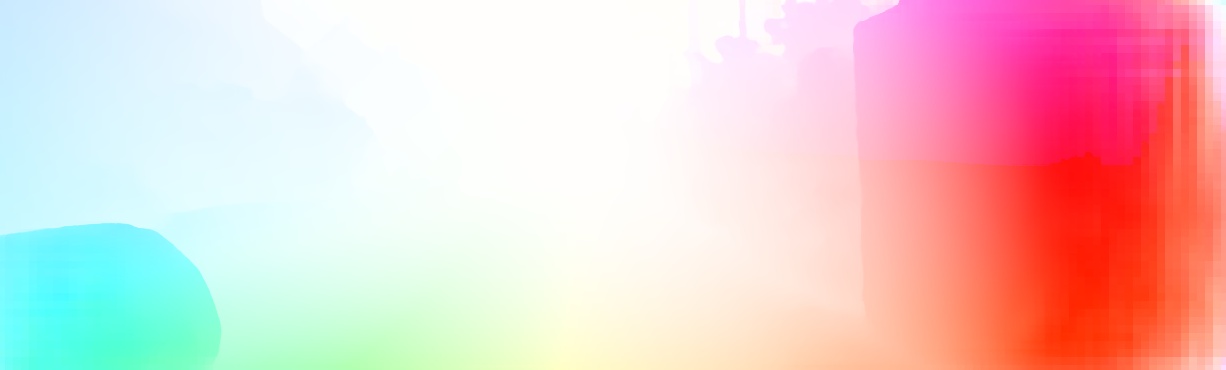}
    \put (2,25) {\footnotesize \textcolor{purple}{\textbf{\texttt{EPE: 6.674 }}}}
    \end{overpic} & 
    \begin{overpic}[width=0.32\linewidth,frame]{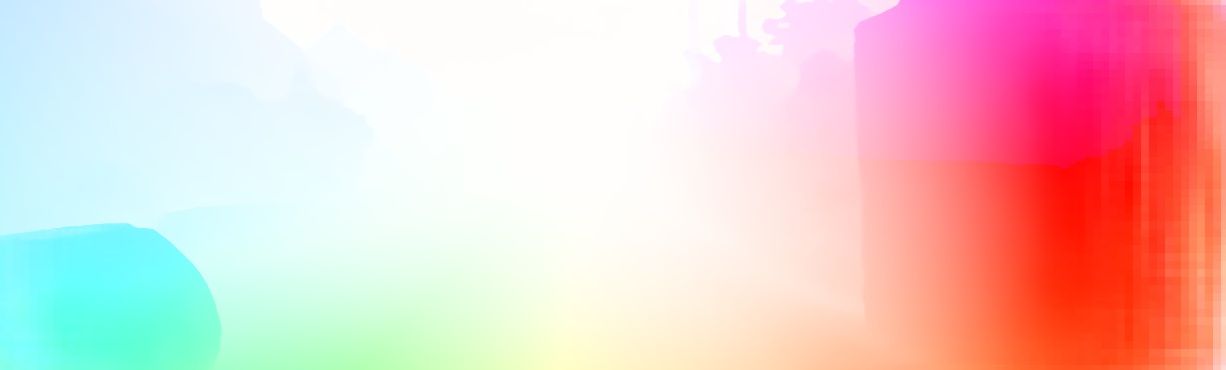}
    \put (2,25) {\footnotesize \textcolor{purple}{\textbf{\texttt{EPE: 7.154 }}}}
    \end{overpic} & 
    \begin{overpic}[width=0.32\linewidth,frame]{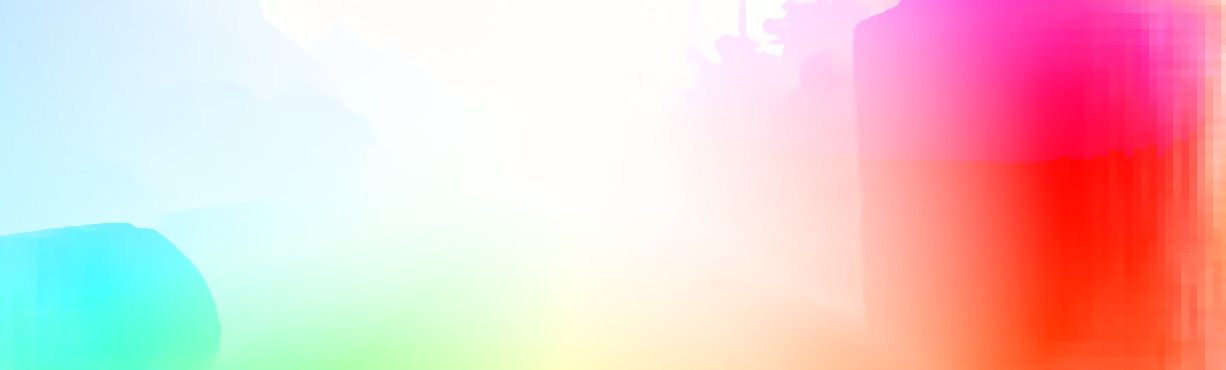}
    \put (2,25) {\footnotesize \textcolor{purple}{\textbf{\texttt{EPE: 6.873 }}}}
    \end{overpic} \vspace{-0.2cm}\\
    \footnotesize + $\mathbf{D}_{0,1}$ & \footnotesize + $\texttt{BaseNet}$ & \bf\footnotesize \net{} (T) \vspace{0.3cm}\\

    \begin{overpic}[width=0.32\linewidth,frame]{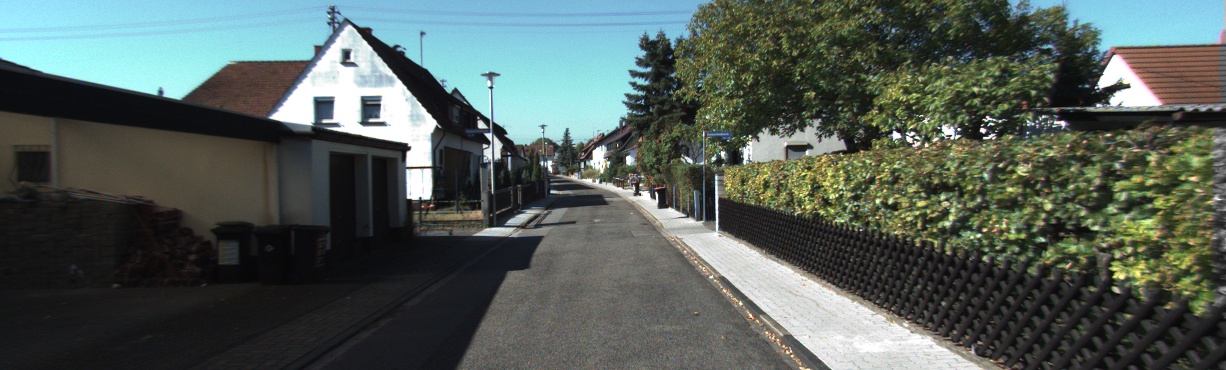}
    \end{overpic} & 
    \begin{overpic}[width=0.32\linewidth,frame]{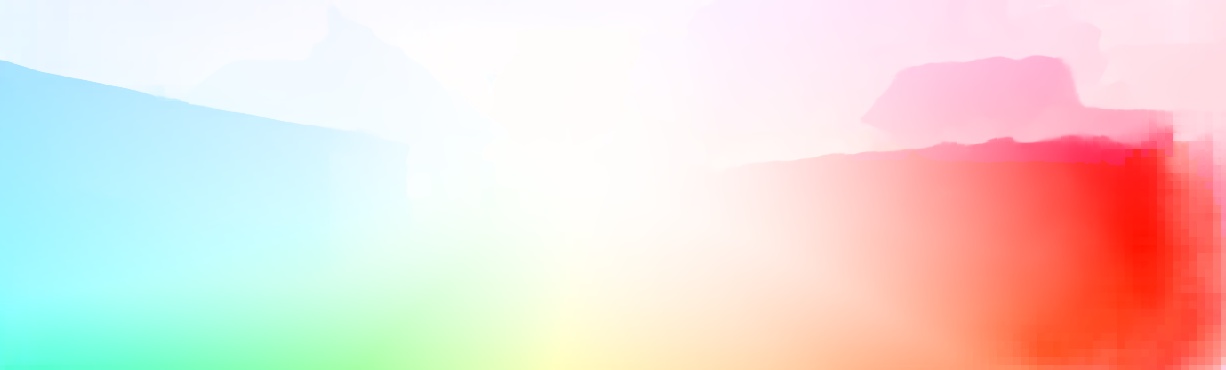}
    \put (2,25) {\footnotesize \textcolor{purple}{\textbf{\texttt{EPE: 3.794 }}}}
    \end{overpic} & 
    \begin{overpic}[width=0.32\linewidth,frame]{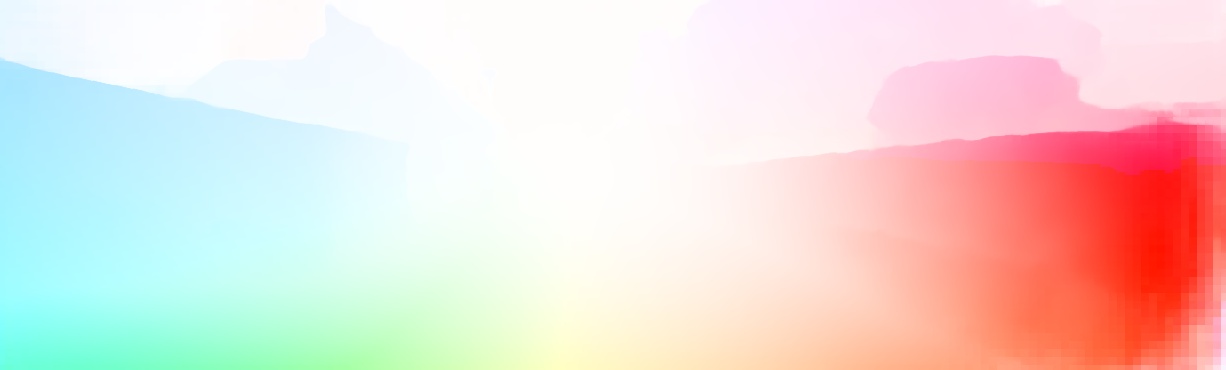}
    \put (2,25) {\footnotesize \textcolor{purple}{\textbf{\texttt{EPE: 2.999 }}}}
    \end{overpic} \vspace{-0.2cm}\\
    \footnotesize $\mathbf{I}_0$ & \footnotesize SEA-RAFT (S) & \footnotesize + $\mathbf{\Phi}_{0,1}$ \\ 
    \begin{overpic}[width=0.32\linewidth,frame]{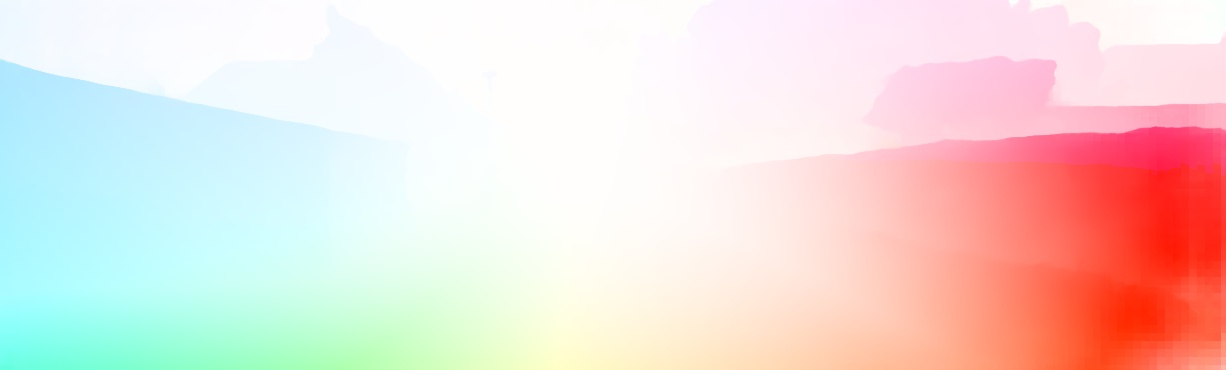}
    \put (2,25) {\footnotesize \textcolor{purple}{\textbf{\texttt{EPE: 2.233 }}}}
    \end{overpic} & 
    \begin{overpic}[width=0.32\linewidth,frame]{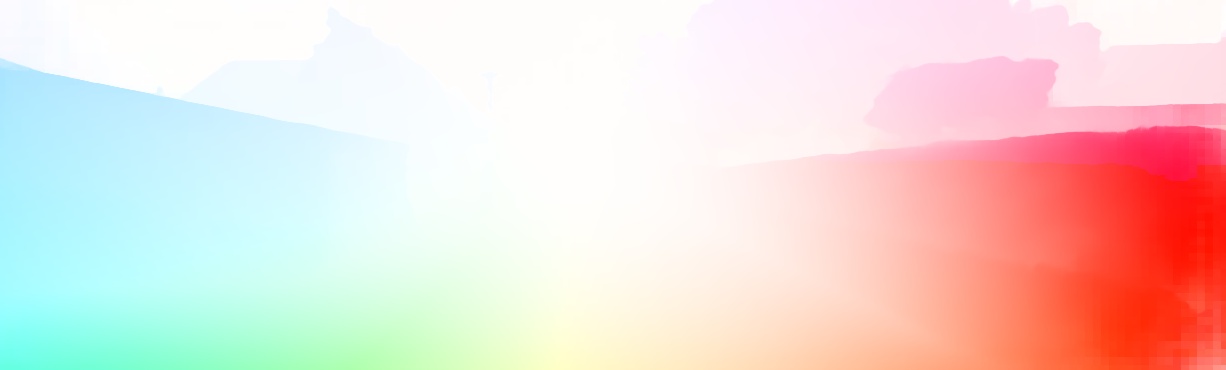}
    \put (2,25) {\footnotesize \textcolor{purple}{\textbf{\texttt{EPE: 2.073 }}}}
    \end{overpic} & 
    \begin{overpic}[width=0.32\linewidth,frame]{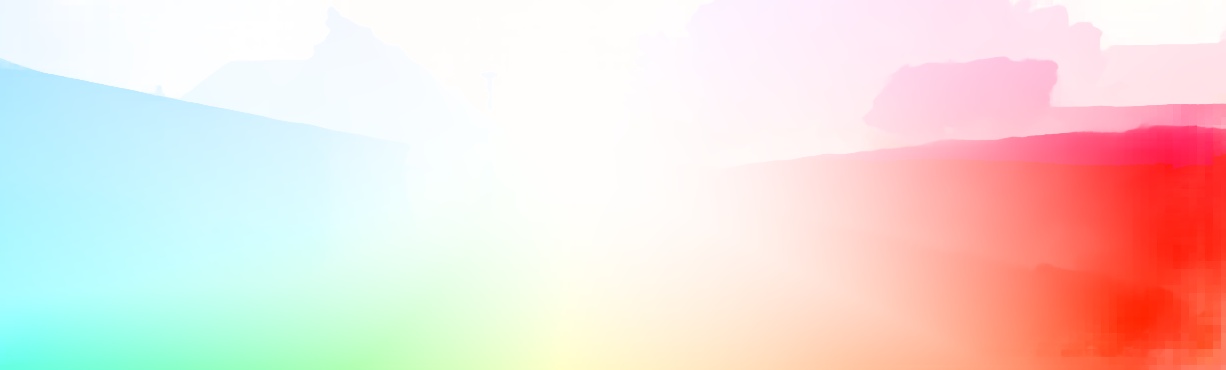}
    \put (2,25) {\footnotesize \textcolor{purple}{\textbf{\texttt{EPE: 1.927 }}}}
    \end{overpic} \vspace{-0.2cm}\\
    \footnotesize + $\mathbf{D}_{0,1}$ & \footnotesize + $\texttt{BaseNet}$ & \bf\footnotesize \net{} (T) \vspace{0.3cm}\\

    \begin{overpic}[width=0.32\linewidth,frame]{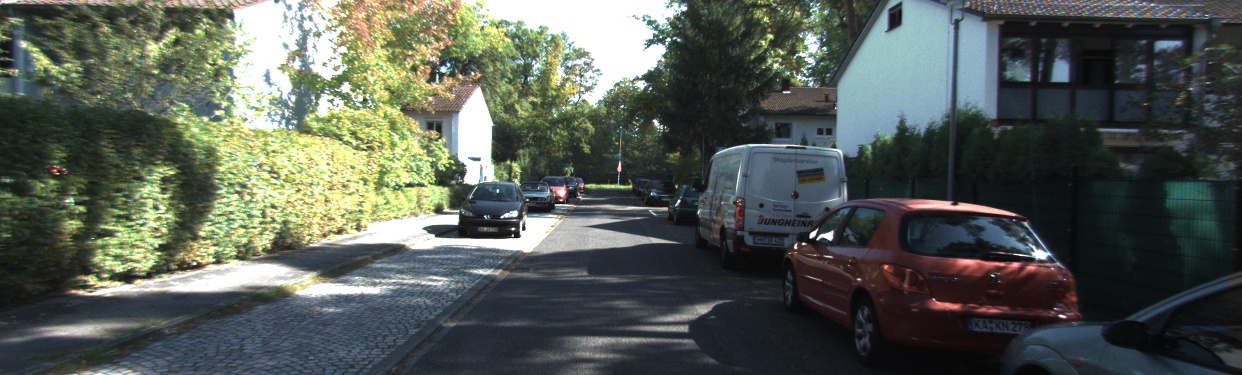}
    \end{overpic} & 
    \begin{overpic}[width=0.32\linewidth,frame]{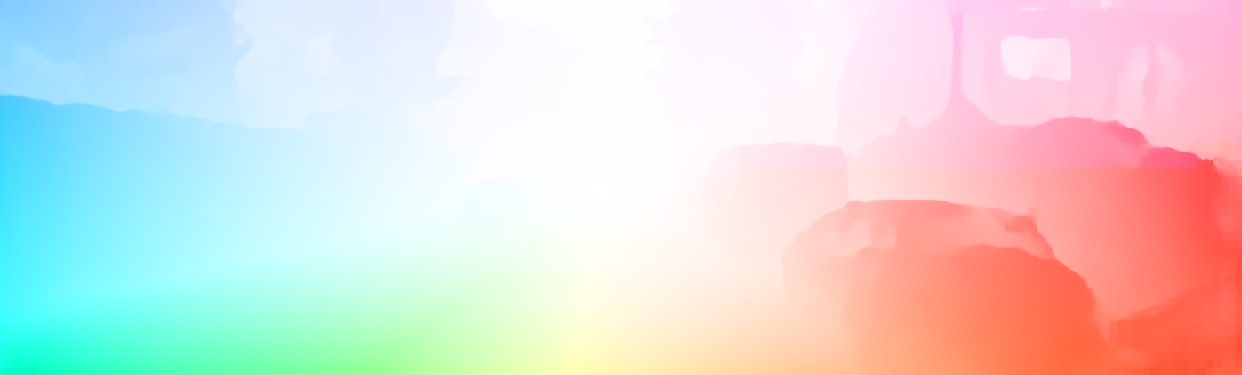}
    \put (2,25) {\footnotesize \textcolor{purple}{\textbf{\texttt{EPE: 1.920 }}}}
    \end{overpic} & 
    \begin{overpic}[width=0.32\linewidth,frame]{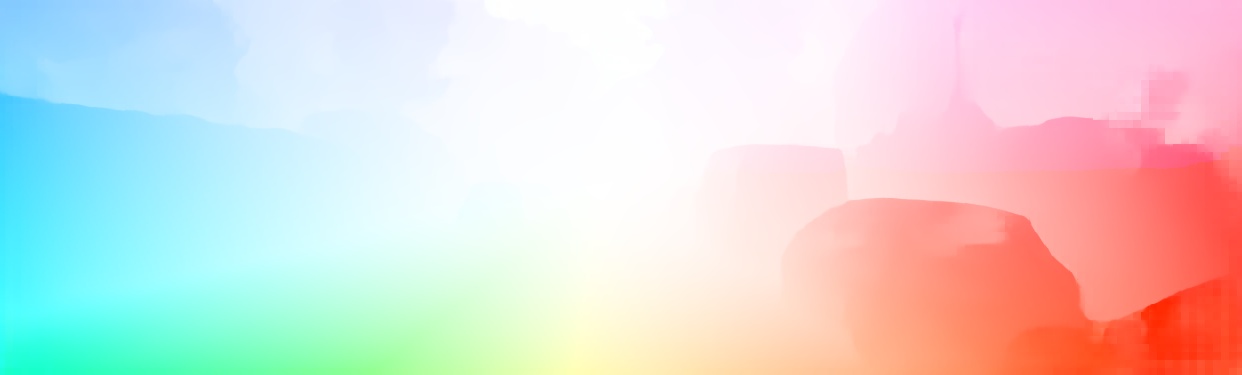}
    \put (2,25) {\footnotesize \textcolor{purple}{\textbf{\texttt{EPE: 1.619 }}}}
    \end{overpic} \vspace{-0.2cm}\\
    \footnotesize $\mathbf{I}_0$ & \footnotesize SEA-RAFT (S) & \footnotesize + $\mathbf{\Phi}_{0,1}$ \\ 
    \begin{overpic}[width=0.32\linewidth,frame]{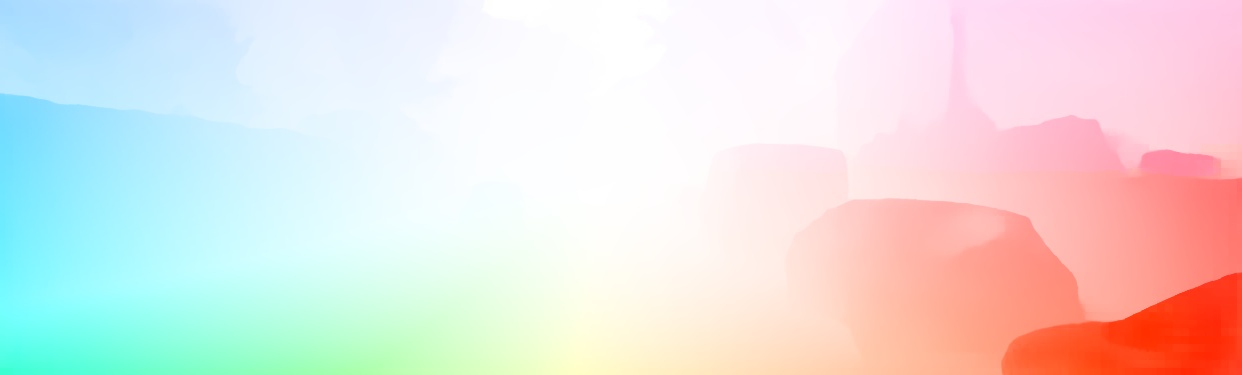}
    \put (2,25) {\footnotesize \textcolor{purple}{\textbf{\texttt{EPE: 1.392 }}}}
    \end{overpic} & 
    \begin{overpic}[width=0.32\linewidth,frame]{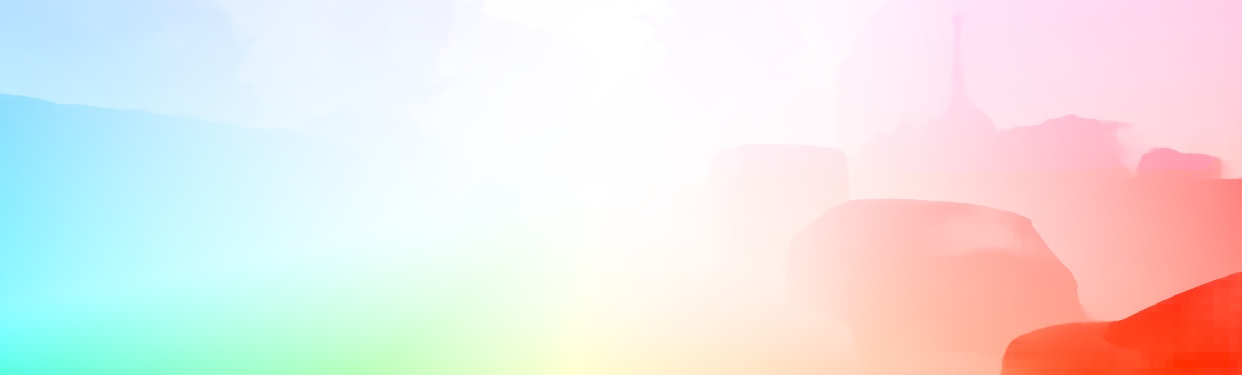}
    \put (2,25) {\footnotesize \textcolor{purple}{\textbf{\texttt{EPE: 1.176 }}}}
    \end{overpic} & 
    \begin{overpic}[width=0.32\linewidth,frame]{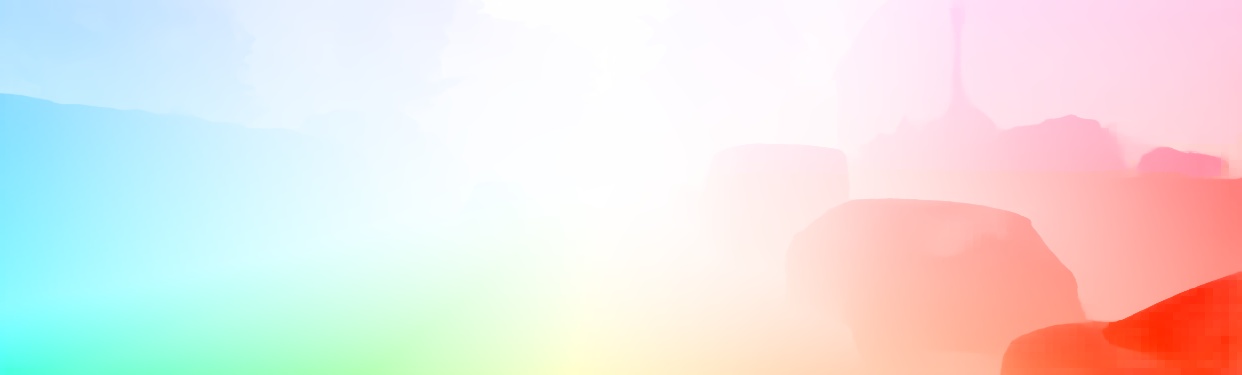}
    \put (2,25) {\footnotesize \textcolor{purple}{\textbf{\texttt{EPE: 1.214 }}}}
    \end{overpic} \vspace{-0.2cm}\\
    \footnotesize + $\mathbf{D}_{0,1}$ & \footnotesize + $\texttt{BaseNet}$ & \bf\footnotesize \net{} (T) \\
    
    \end{tabular}
    
    \caption{\textbf{Qualitative results on KITTI 2012.} From left to right, on two rows: first frame, flow predicted by SEA-RAFT (S) and ablated versions of \net{} (T).}
    \label{fig:2012}
\end{figure*}

\begin{figure*}[t]
    \centering
    \renewcommand{\tabcolsep}{1pt}
    \begin{tabular}{ccc}

    \begin{overpic}[width=0.32\linewidth,frame]{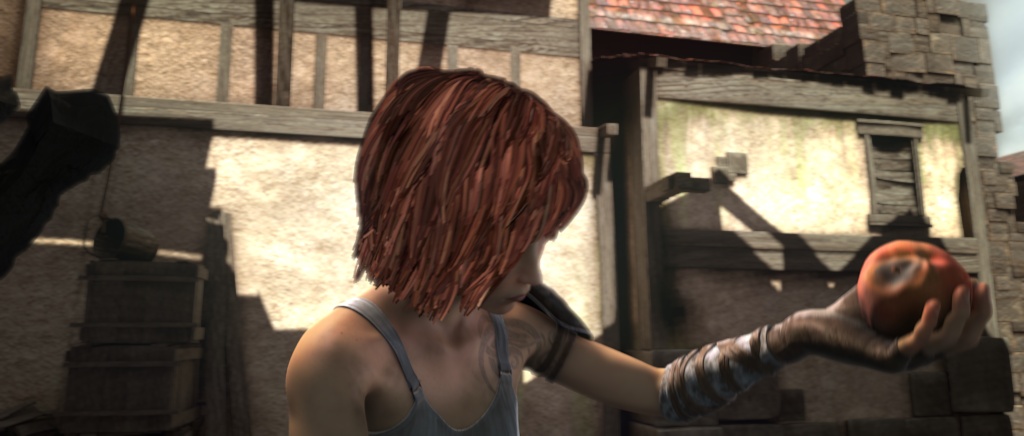}
    \end{overpic} & 
    \begin{overpic}[width=0.32\linewidth,frame]{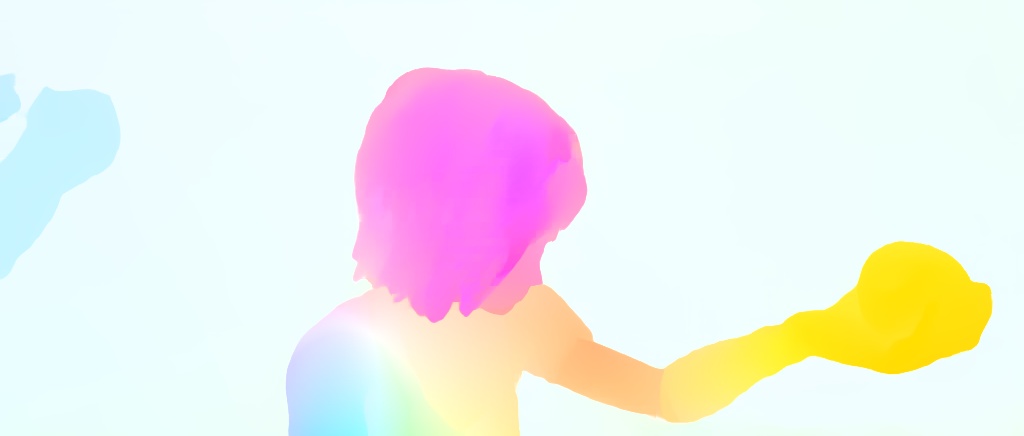}
    \put (2,37) {\small \textcolor{purple}{\textbf{\texttt{EPE: 0.301 }}}}
    \end{overpic} & 
    \begin{overpic}[width=0.32\linewidth,frame]{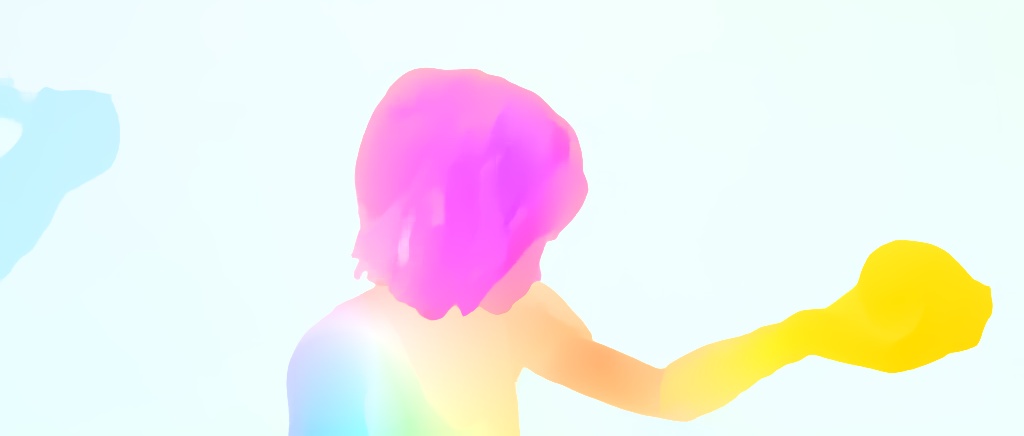}
    \put (2,37) {\small \textcolor{purple}{\textbf{\texttt{EPE: 0.291 }}}}
    \end{overpic} \\
    \begin{overpic}[width=0.32\linewidth,frame]{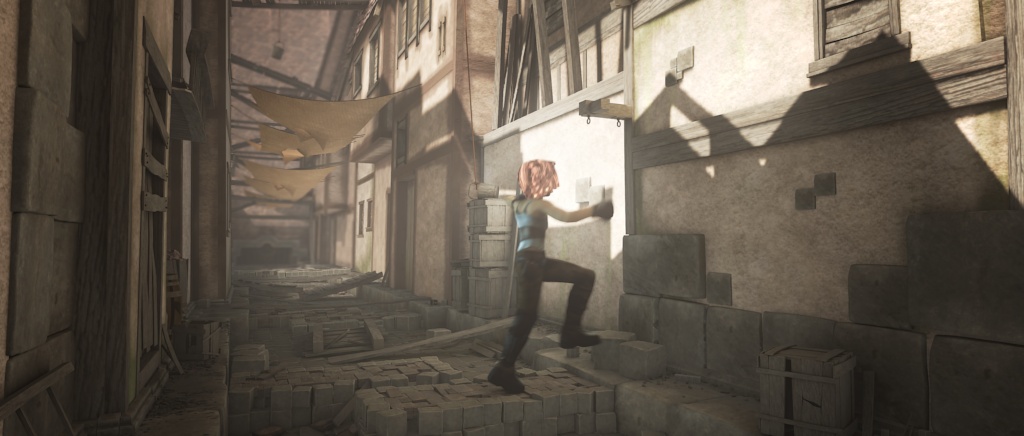}
    \end{overpic} & 
    \begin{overpic}[width=0.32\linewidth,frame]{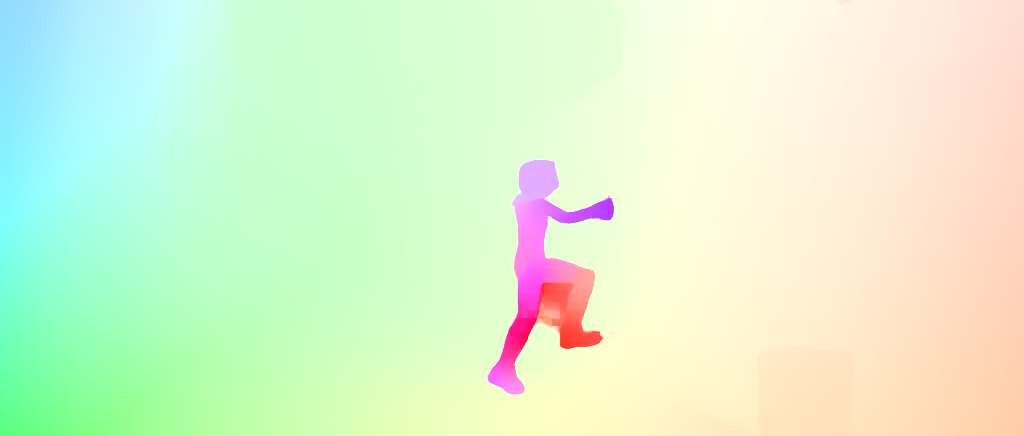}
    \put (2,37) {\small \textcolor{purple}{\textbf{\texttt{EPE: 0.214 }}}}
    \end{overpic} & 
    \begin{overpic}[width=0.32\linewidth,frame]{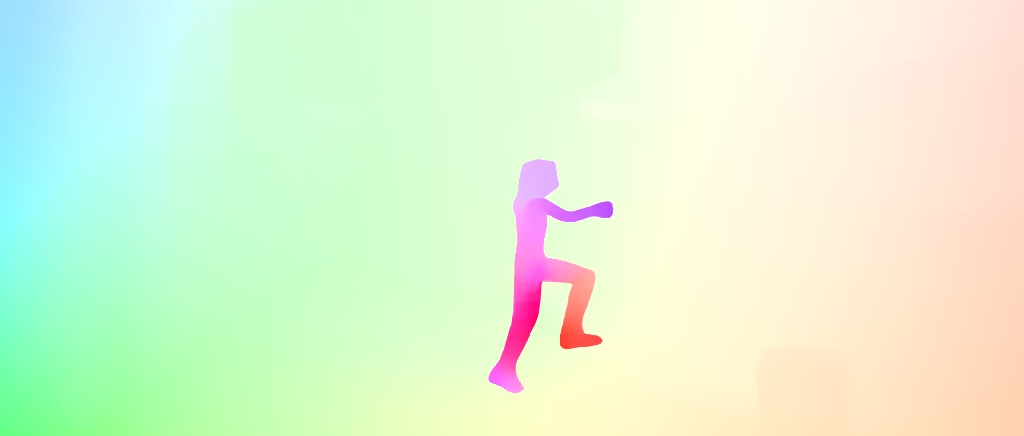}
    \put (2,37) {\small \textcolor{purple}{\textbf{\texttt{EPE: 0.168 }}}}
    \end{overpic} \\
    \begin{overpic}[width=0.32\linewidth,frame]{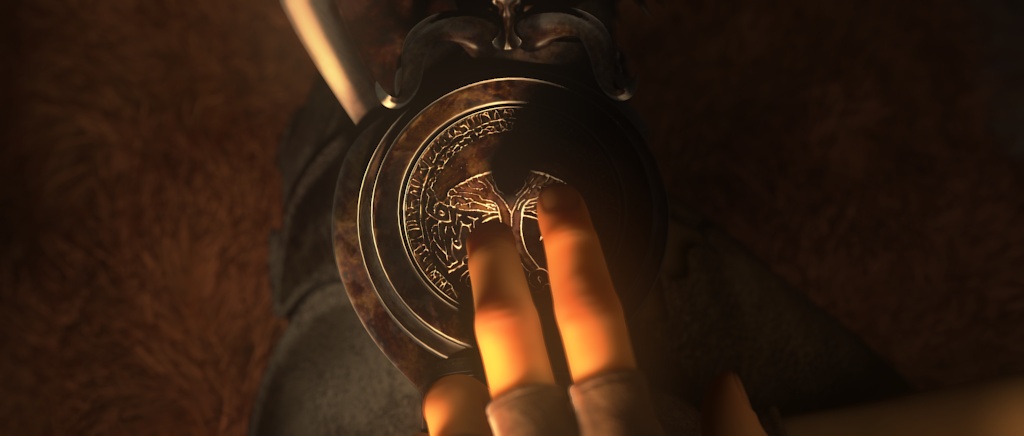}
    \end{overpic} & 
    \begin{overpic}[width=0.32\linewidth,frame]{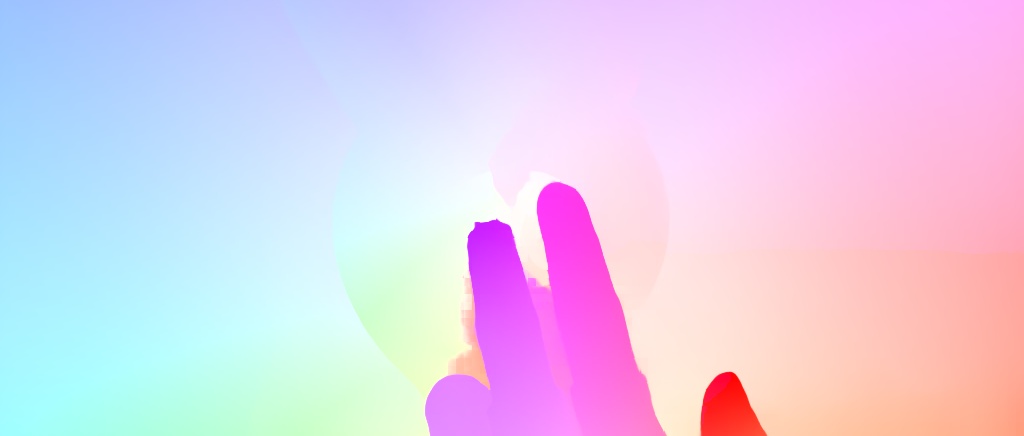}
    \put (2,37) {\small \textcolor{purple}{\textbf{\texttt{EPE: 0.210 }}}}
    \end{overpic} & 
    \begin{overpic}[width=0.32\linewidth,frame]{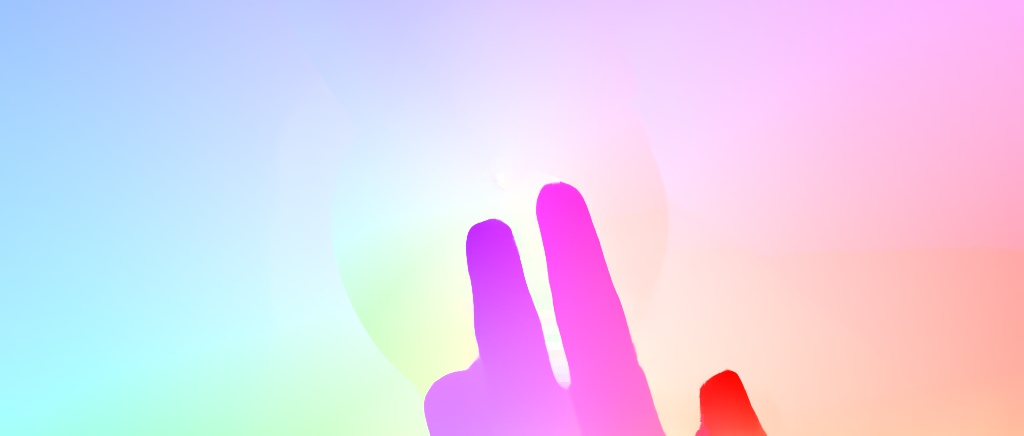}
    \put (2,37) {\small \textcolor{purple}{\textbf{\texttt{EPE: 0.193 }}}}
    \end{overpic} \vspace{-0.2cm}\\

    \footnotesize $\mathbf{I}_0$ & \footnotesize SEA-RAFT (S) & \bf\footnotesize \net{} (T) \vspace{-0.2cm}\\
    \end{tabular}
    
    \caption{\textbf{Qualitative results on Sintel.} From left to right: first frame, flow by SEA-RAFT (S) and FlowSeek (T).}
    \label{fig:sintel}
\end{figure*}

\begin{figure*}[t]
    \centering
    \renewcommand{\tabcolsep}{1pt}
    \begin{tabular}{ccc}

    \begin{overpic}[width=0.32\linewidth,frame]{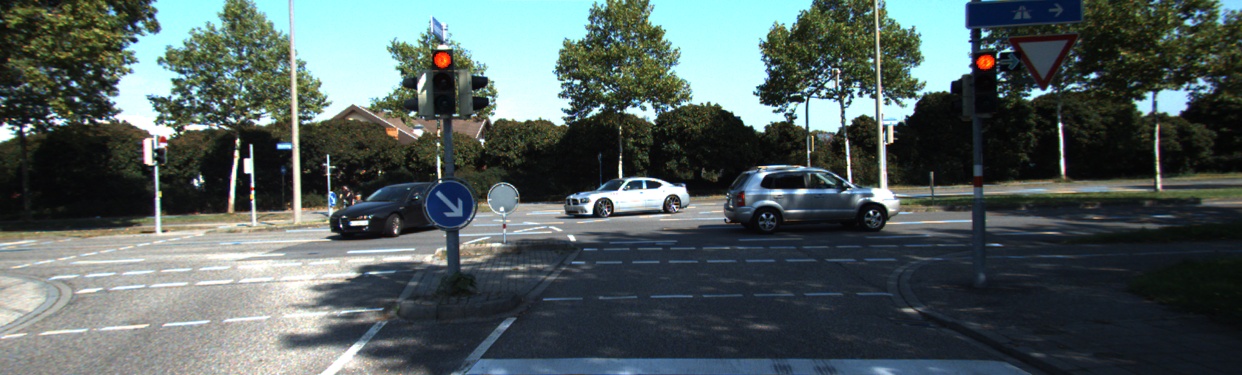}
    \end{overpic} & 
    \begin{overpic}[width=0.32\linewidth,frame]{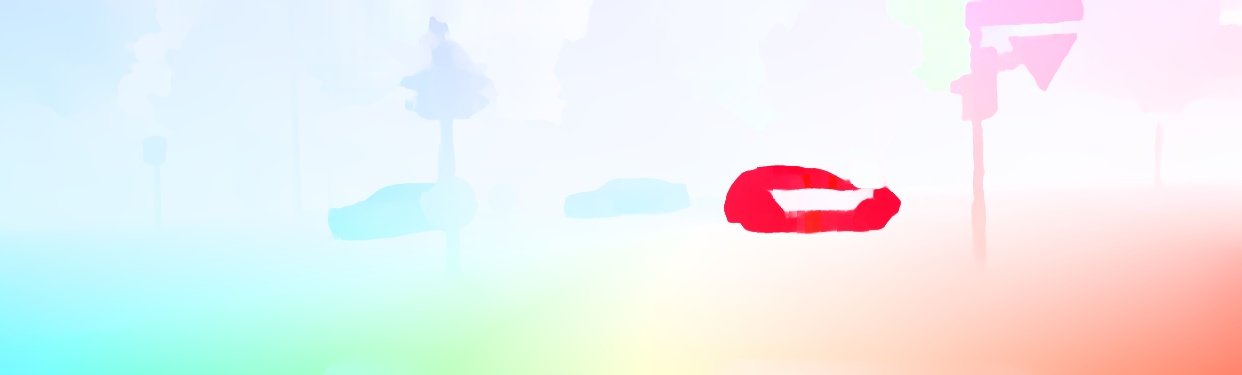}
    \put (2,25) {\small \textcolor{purple}{\textbf{\texttt{EPE: 1.266 }}}}
    \end{overpic} & 
    \begin{overpic}[width=0.32\linewidth,frame]{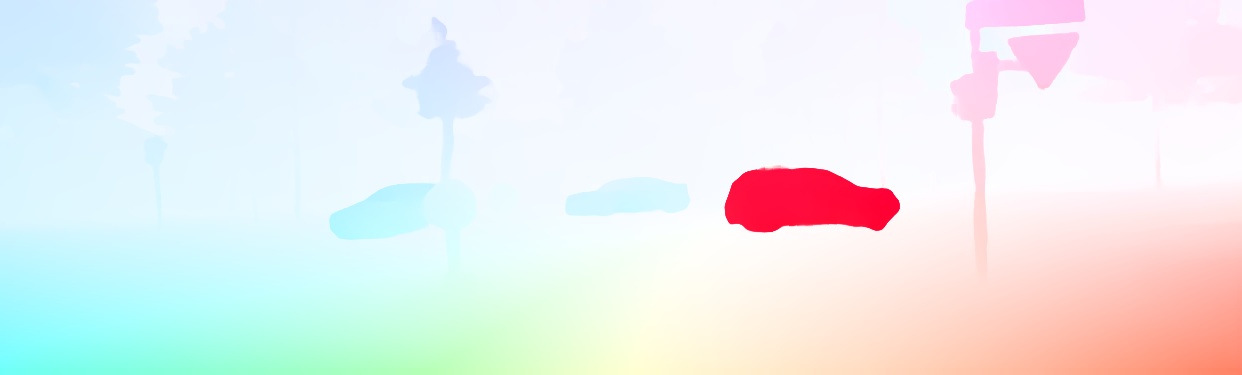}
    \put (2,25) {\small \textcolor{purple}{\textbf{\texttt{EPE: 0.427 }}}}
    \end{overpic} \\

    \begin{overpic}[width=0.32\linewidth,frame]{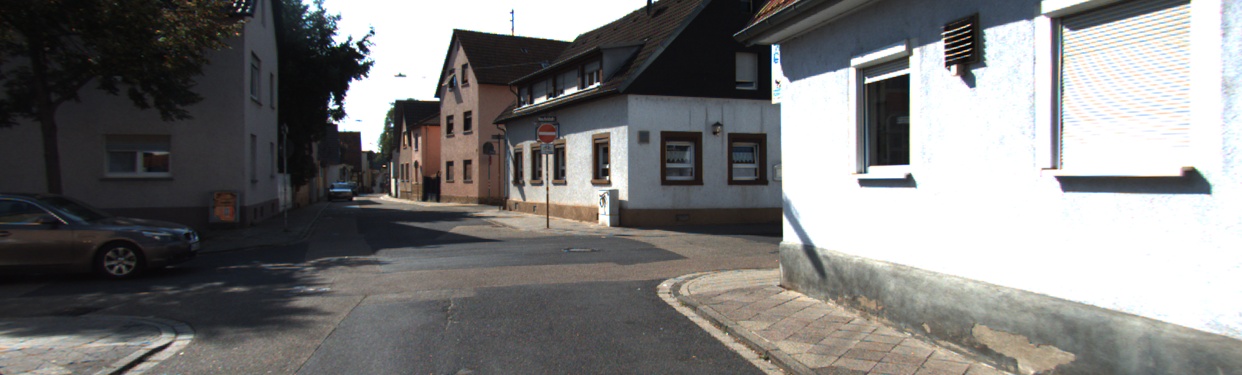}
    \end{overpic} & 
    \begin{overpic}[width=0.32\linewidth,frame]{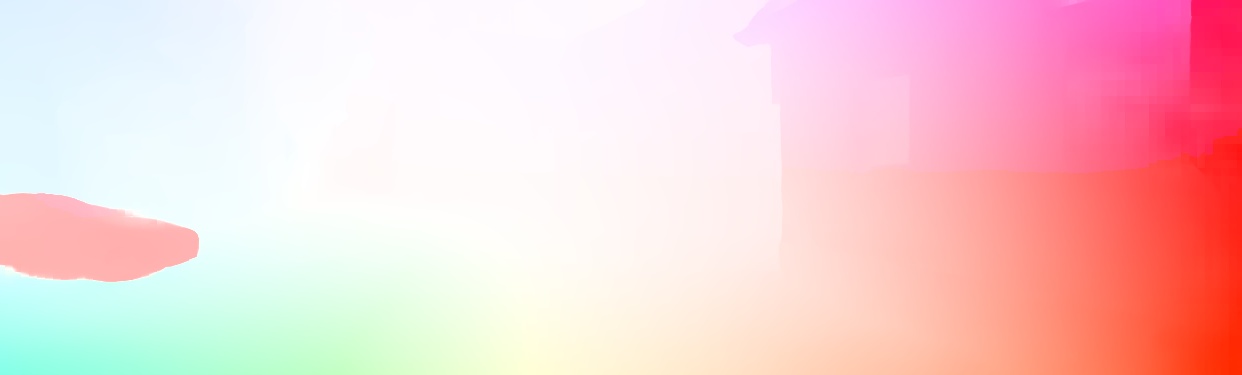}
    \put (2,25) {\small \textcolor{purple}{\textbf{\texttt{EPE: 0.760 }}}}
    \end{overpic} & 
    \begin{overpic}[width=0.32\linewidth,frame]{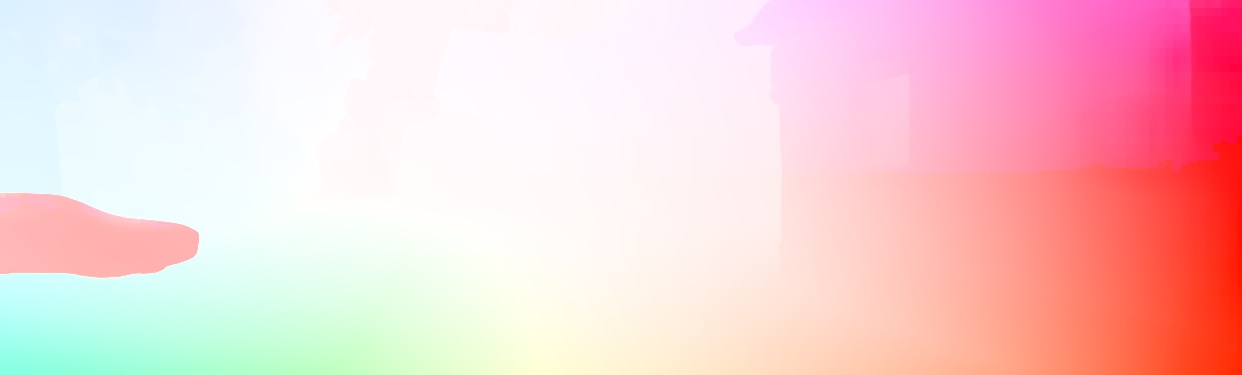}
    \put (2,25) {\small \textcolor{purple}{\textbf{\texttt{EPE: 0.661 }}}}
    \end{overpic} \vspace{-0.2cm}\\
    
    \begin{overpic}[width=0.32\linewidth,frame]{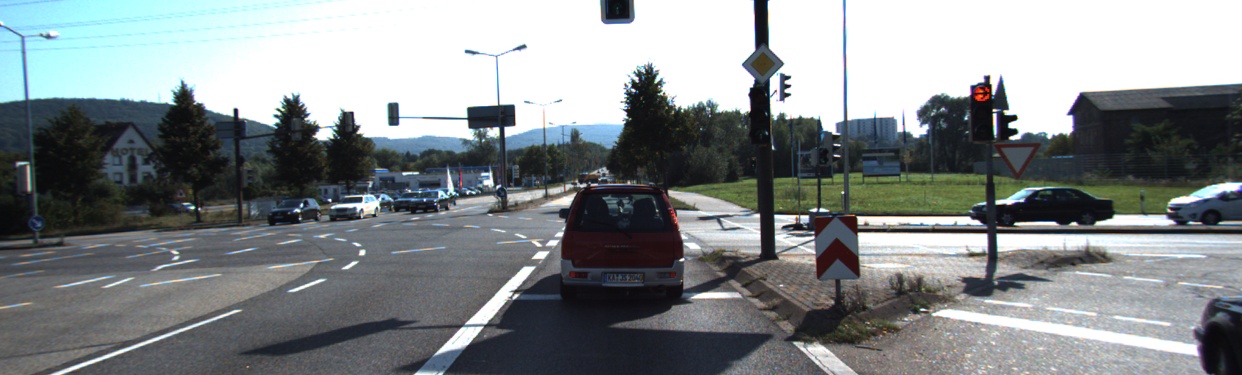}
    \end{overpic} & 
    \begin{overpic}[width=0.32\linewidth,frame]{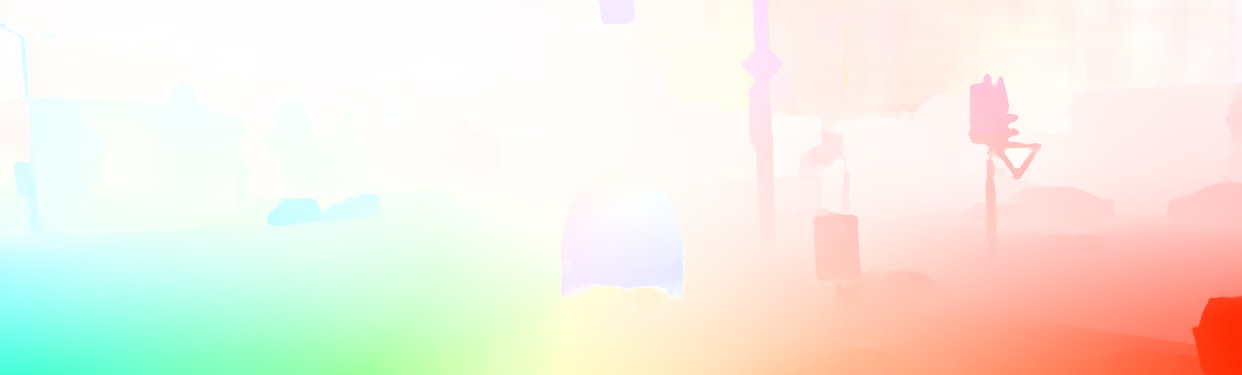}
    \put (2,25) {\small \textcolor{purple}{\textbf{\texttt{EPE: 0.448 }}}}
    \end{overpic} & 
    \begin{overpic}[width=0.32\linewidth,frame]{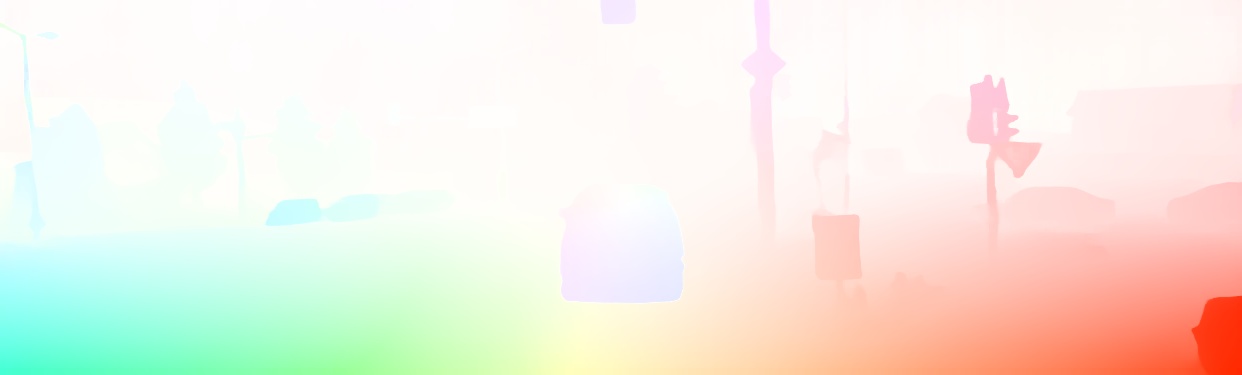}
    \put (2,25) {\small \textcolor{purple}{\textbf{\texttt{EPE: 0.418 }}}}
    \end{overpic} \\

    \begin{overpic}[width=0.32\linewidth,frame]{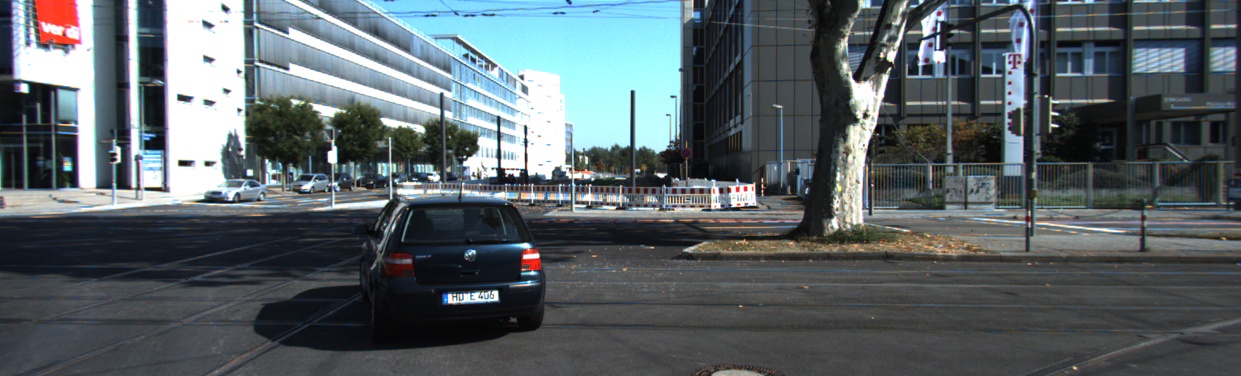}
    \end{overpic} & 
    \begin{overpic}[width=0.32\linewidth,frame]{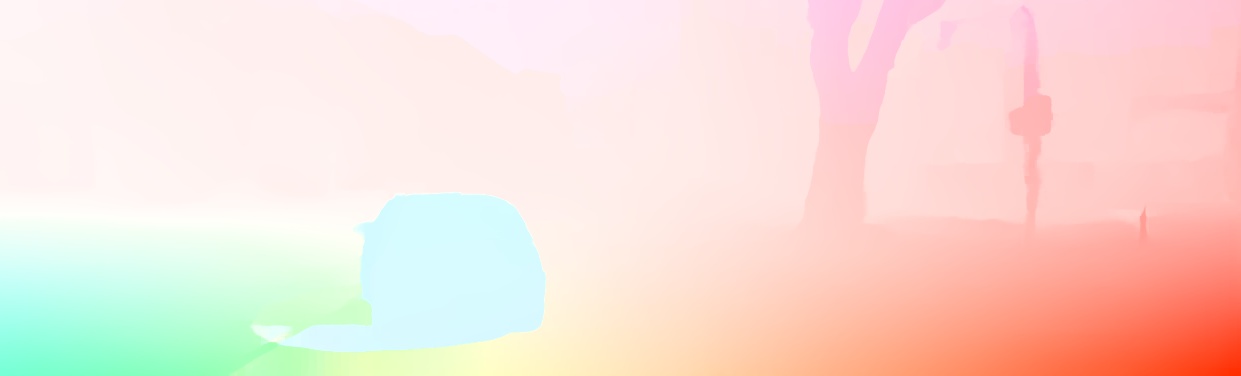}
    \put (2,25) {\small \textcolor{purple}{\textbf{\texttt{EPE: 0.682 }}}}
    \end{overpic} & 
    \begin{overpic}[width=0.32\linewidth,frame]{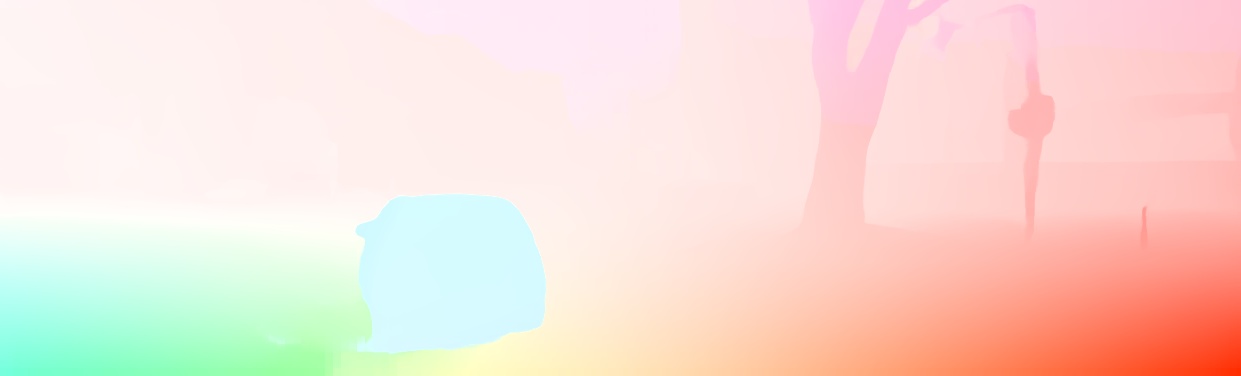}
    \put (2,25) {\small \textcolor{purple}{\textbf{\texttt{EPE: 0.533 }}}}
    \end{overpic} \vspace{-0.2cm}\\

    \footnotesize $\mathbf{I}_0$ & \footnotesize SEA-RAFT (S) & \bf\footnotesize \net{} (T) \vspace{-0.2cm}\\
    \end{tabular}
    
    \caption{\textbf{Qualitative results on KITTI 2015.} From left to right: first frame, flow by SEA-RAFT (S) and FlowSeek (T).}
    \label{fig:2015}
\end{figure*}

\begin{figure*}[t]
    \centering
    \renewcommand{\tabcolsep}{1pt}
    \begin{tabular}{ccc}

    \begin{overpic}[width=0.32\linewidth,frame]{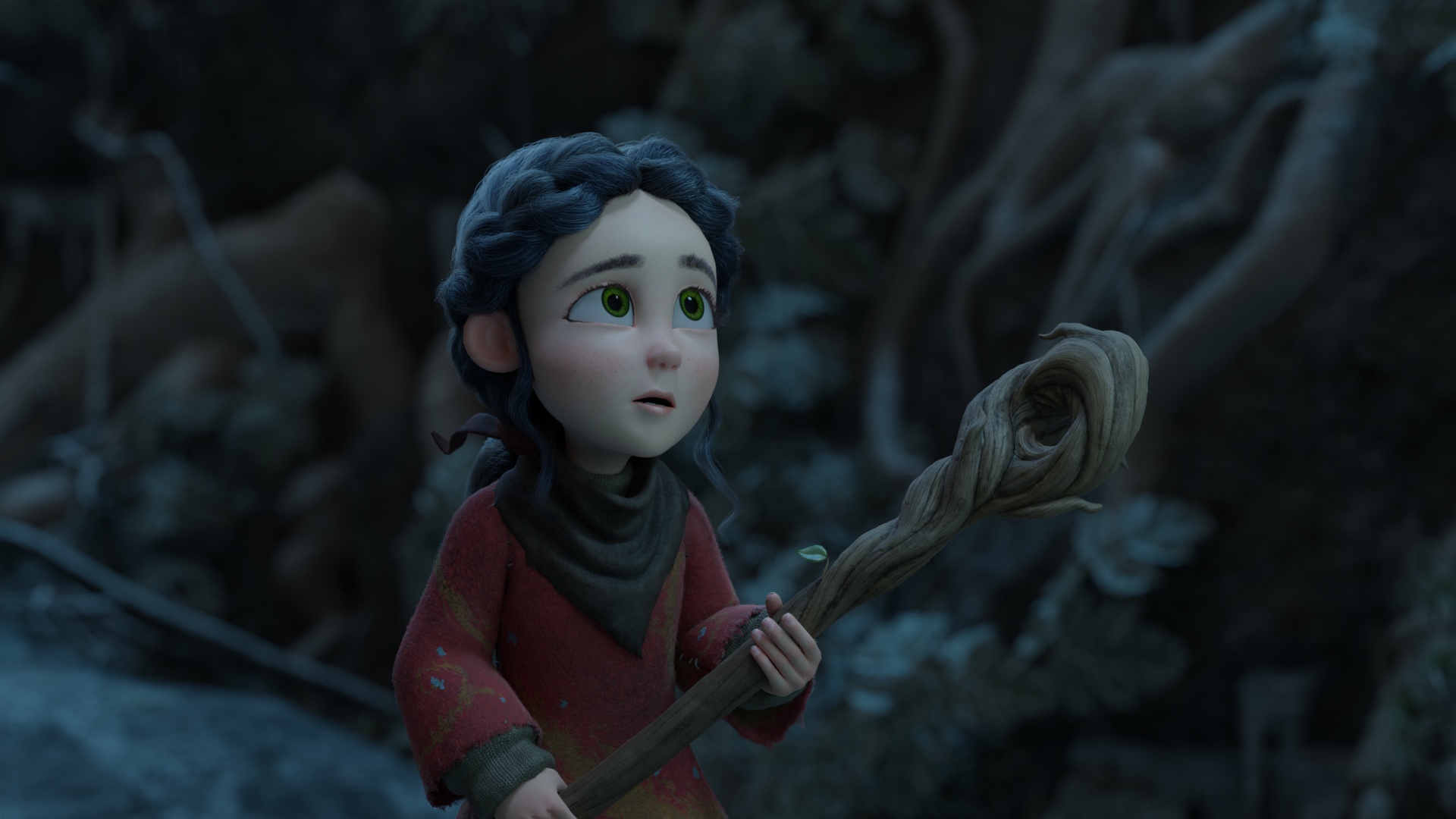}
    \end{overpic} & 
    \begin{overpic}[width=0.32\linewidth,frame]{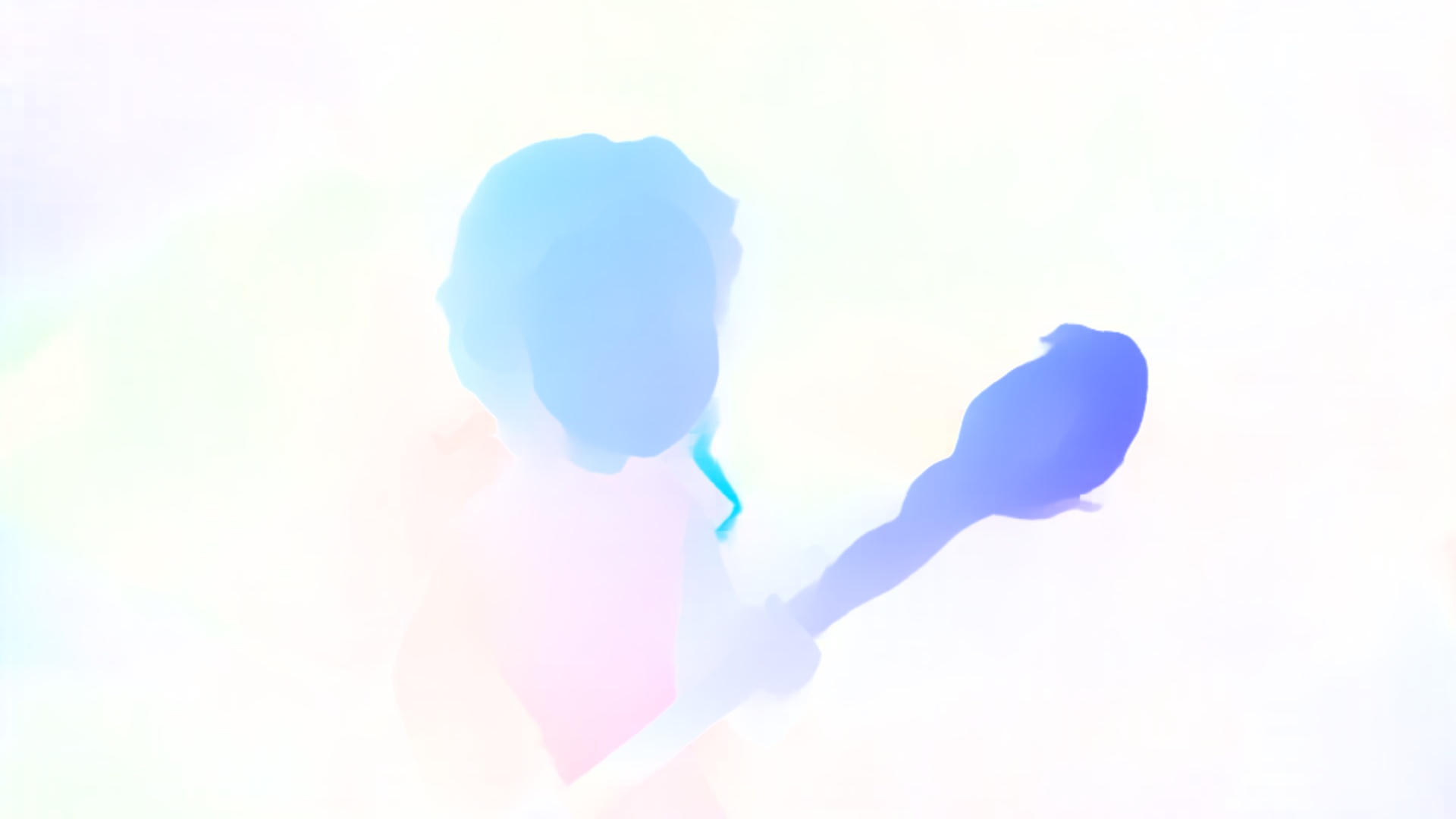}
    \put (2,52) {\small \textcolor{purple}{\textbf{\texttt{EPE: 0.058 }}}}
    \end{overpic} & 
    \begin{overpic}[width=0.32\linewidth,frame]{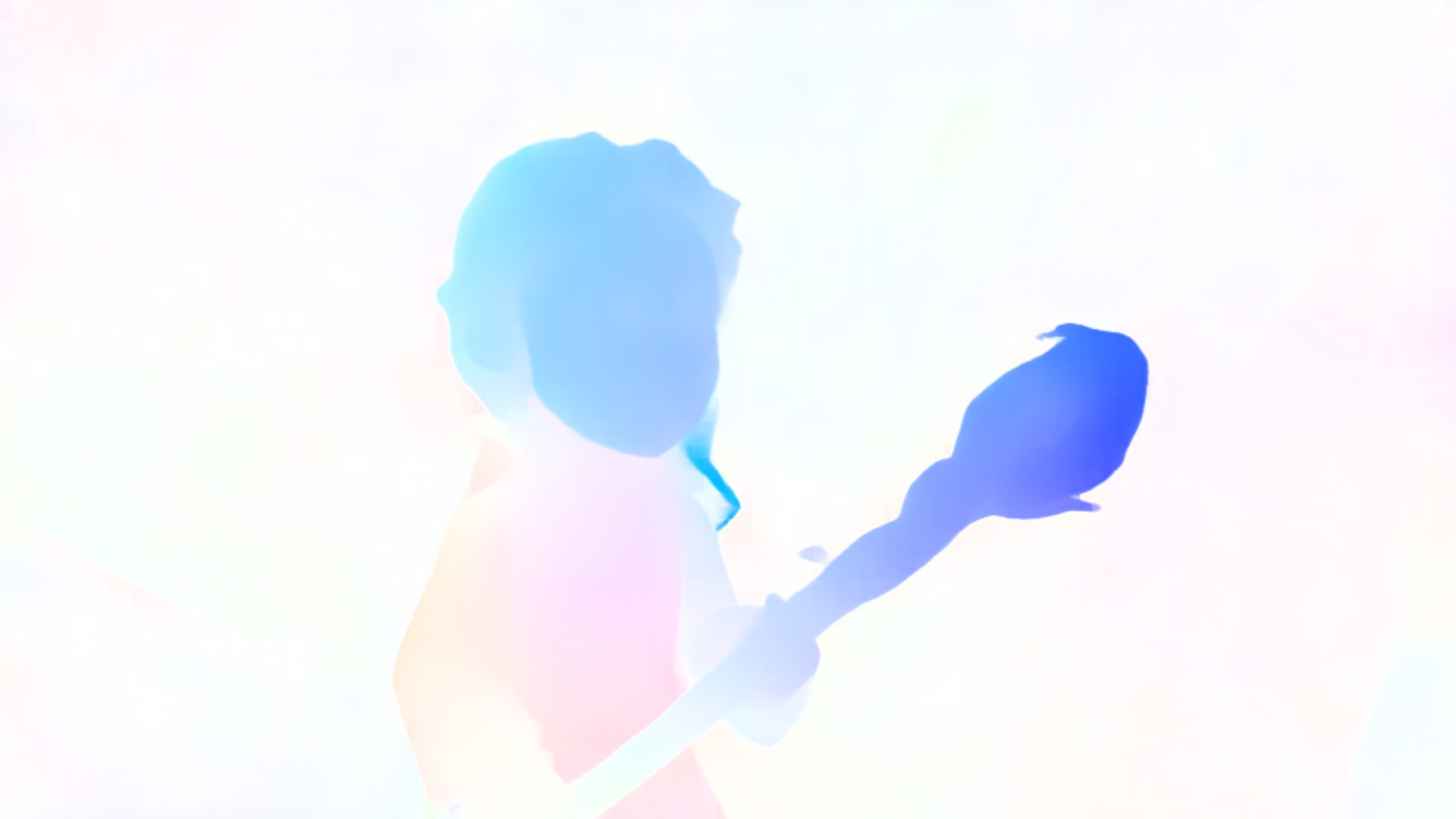}
    \put (2,52) {\small \textcolor{purple}{\textbf{\texttt{EPE: 0.046 
    }}}}
    \end{overpic} \\
    
    \begin{overpic}[width=0.32\linewidth,frame]{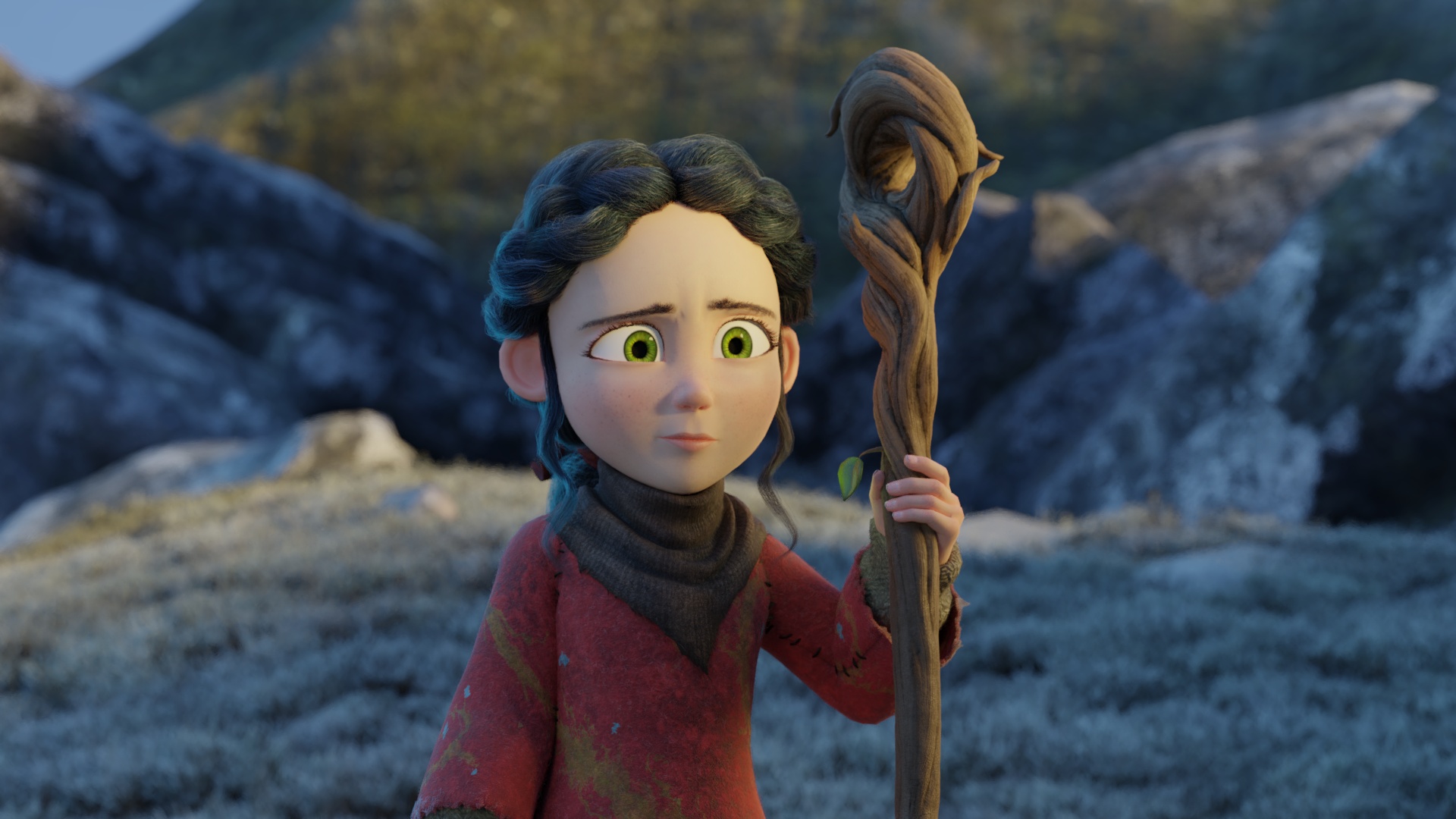}
    \end{overpic} & 
    \begin{overpic}[width=0.32\linewidth,frame]{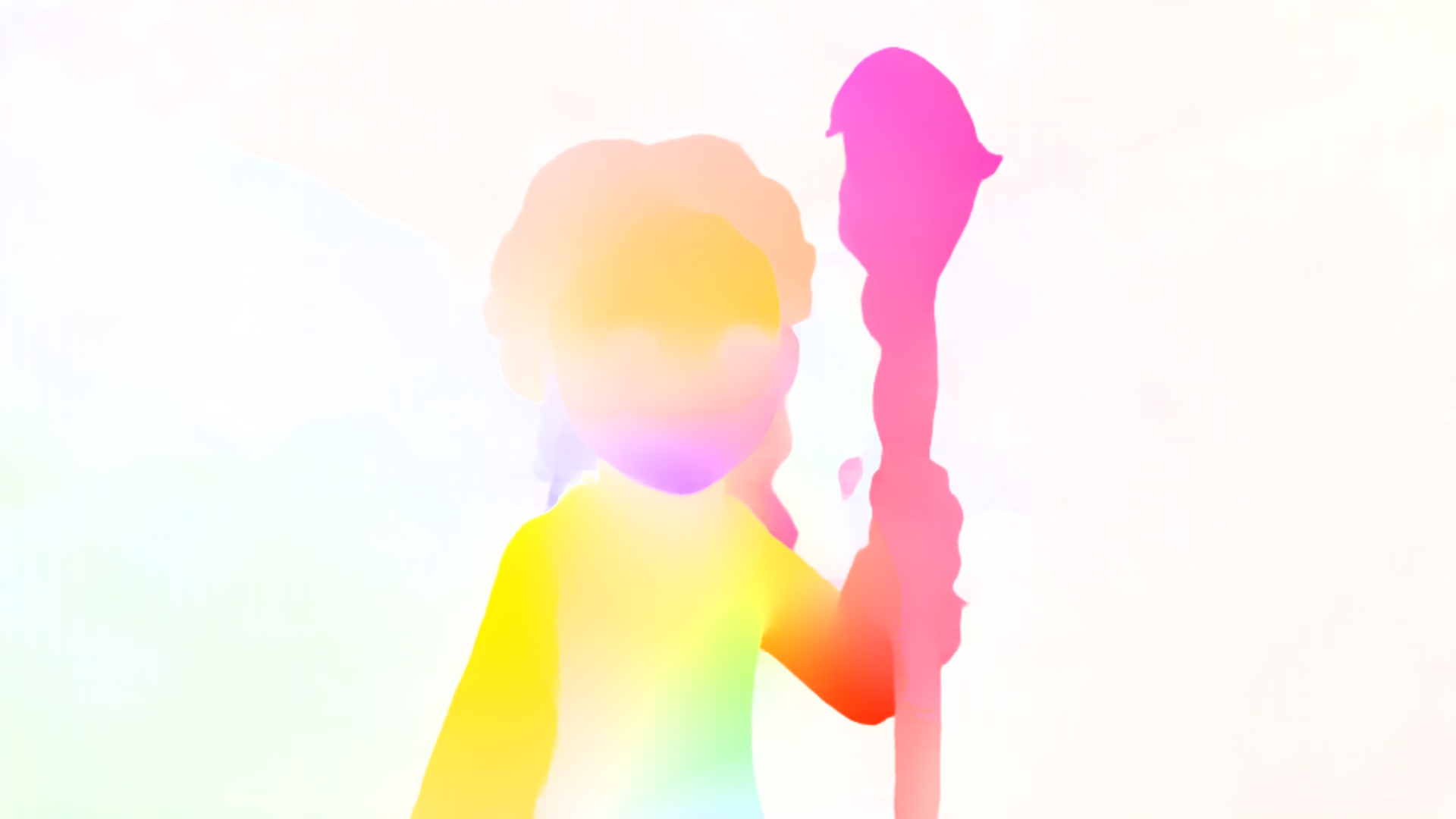}
    \put (2,52) {\small \textcolor{purple}{\textbf{\texttt{EPE: 0.084 }}}}
    \end{overpic} & 
    \begin{overpic}[width=0.32\linewidth,frame]{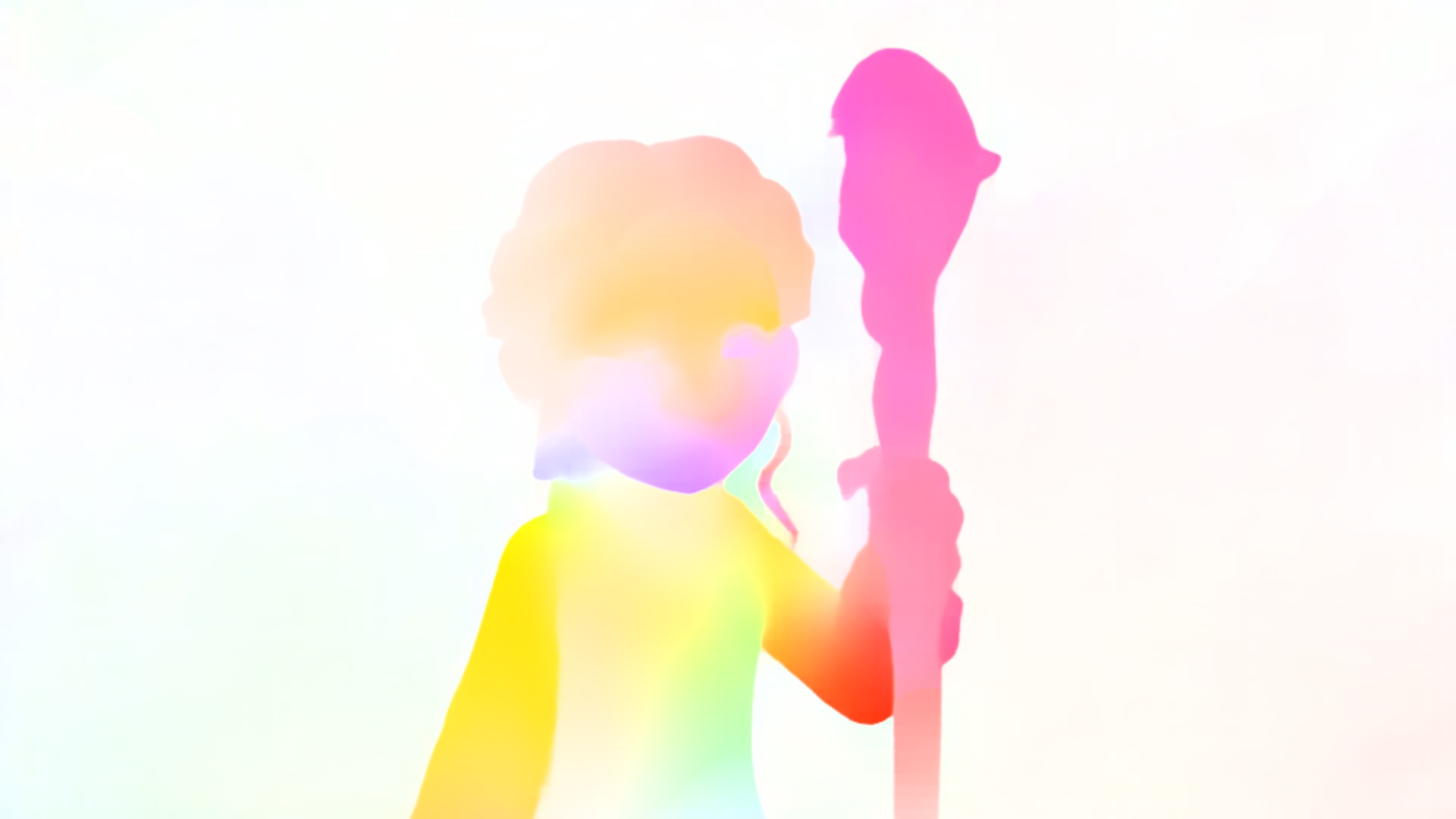}
    \put (2,52) {\small \textcolor{purple}{\textbf{\texttt{EPE: 0.074 
    }}}}
    \end{overpic} \vspace{-0.2cm}\\

    \footnotesize $\mathbf{I}_0$ & \footnotesize SEA-RAFT (S) & \bf\footnotesize \net{} (T) \vspace{-0.2cm}\\
    \end{tabular}
    
    \caption{\textbf{Qualitative results on Spring  -- ``C$\rightarrow$T$\rightarrow$TSKH" schedule after TartanAir pretraining.} From left to right: first frame, flow by SEA-RAFT (S) and FlowSeek (T).}
    \label{fig:spring}
\end{figure*}

\begin{table*}[b]
    \centering
    \resizebox{1.0\linewidth}{!}{
    \renewcommand{\tabcolsep}{10pt}
    \begin{tabular}{cccc}
        \hspace{2cm} &
        \begin{tabular}{llrrrrrr}
    \toprule
    \multirow{2}{*}{Extra Data} & \multirow{2}{*}{Method} &\multicolumn{2}{c}{Spring (train)} \\
    \cmidrule(l{0.5ex}r{0.5ex}){3-4} 
    & 
    & 1px$\downarrow$ & EPE$\downarrow$ 
    \\ 
        \midrule 
        \multirow{8}{*}{Tartan}
        & {SEA-RAFT (S)}  & 4.574 & 0.597 \\
        & {SEA-RAFT (M)} & 4.815 & 0.580 \\ 
        & {SEA-RAFT (L)} & 5.988 & 0.850 \\ 
        & \bf \net{} (T)  & 4.070 & 0.416 \\ 
        & \bf \net{} (S)  & 3.999 & 0.410 \\ 
        & \bf \net{} (M)  & 4.094 & 0.424\\ 
        & \bf \net{} (L)  & 4.067 & 0.422 \\ 
    \bottomrule
    \end{tabular}
    & 
        \begin{tabular}{llrrrrrr}
    \toprule
    \multirow{2}{*}{Extra Data} & \multirow{2}{*}{Method} &\multicolumn{2}{c}{Spring (train)} \\
    \cmidrule(l{0.5ex}r{0.5ex}){3-4} 
    & 
    & 1px$\downarrow$ & EPE$\downarrow$ 
    \\ 
        \midrule 
        \multirow{8}{*}{Tartan}
        & {SEA-RAFT (S)}  & 4.161 & 0.410 \\
        & {SEA-RAFT (M)} & \trd{3.888} & \snd{0.406} \\ 
        & {SEA-RAFT (L)} & \snd 3.842 & 0.426 \\ 
        & \bf \net{} (T)  & 4.111 & 0.410 \\ 
        & \bf \net{} (S)  & 4.058 & \snd 0.406 \\ 
        & \bf \net{} (M)  & 3.941 & 0.419 \\ 
        & \bf \net{} (L)  & \fst 3.838 & \fst 0.402 \\ 
    \bottomrule
    \end{tabular} & \hspace{2cm} \\
    \end{tabular}
    }\vspace{-0.3cm}
    \caption{
    \textbf{Zero-Shot Generalization -- Spring.} Models trained with ``C $\rightarrow$ T" (left) and ``C $\rightarrow$ T $\rightarrow$ TSKH" (right) schedule. }
    \label{tab:spring_gen}
    \vspace{-0.3cm}
\end{table*}

\begin{figure*}[t]
    \centering
    \renewcommand{\tabcolsep}{1pt}
    \begin{tabular}{ccc}

    \begin{overpic}[width=0.32\linewidth,frame]{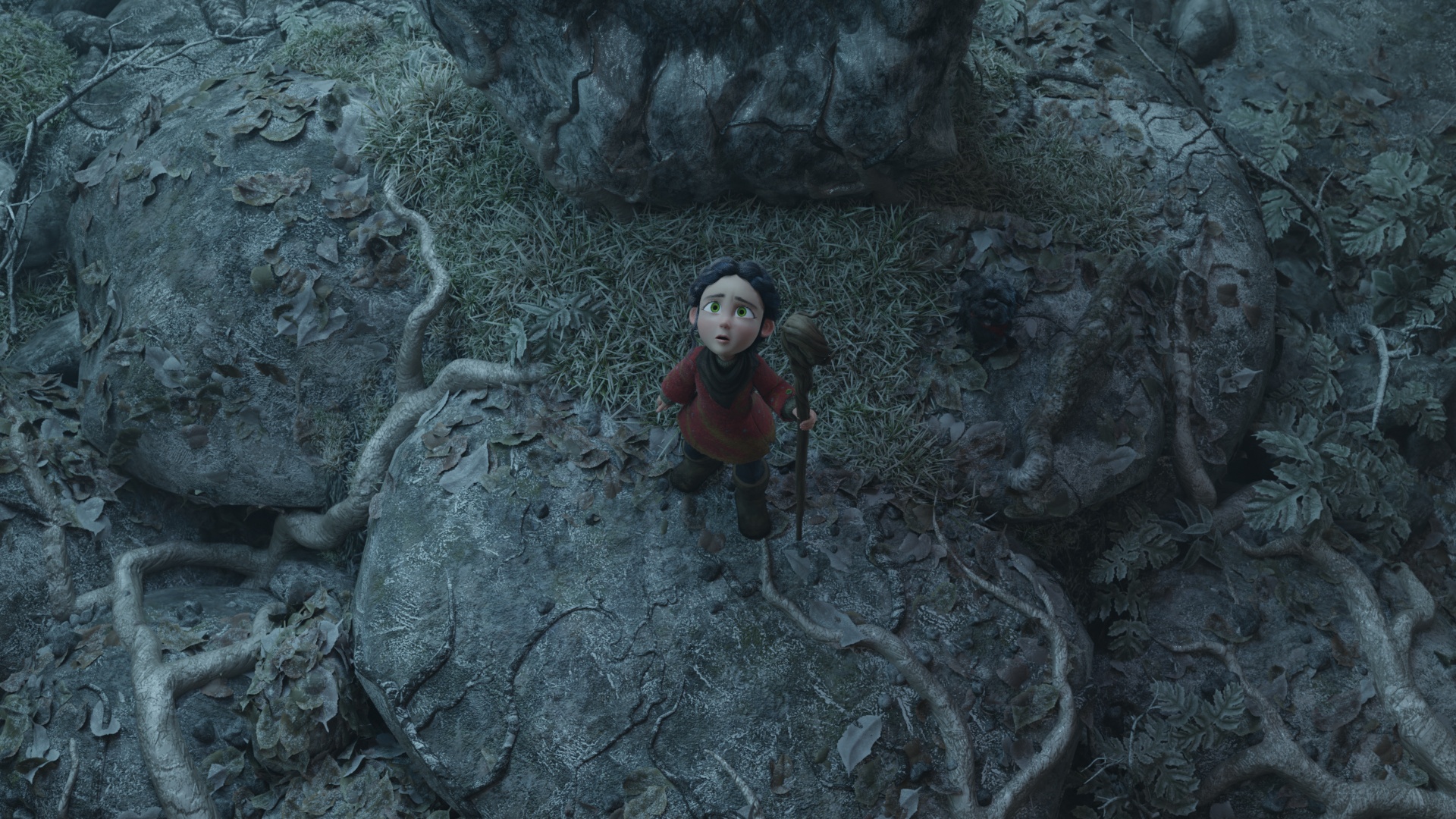}
    \end{overpic} & 
    \begin{overpic}[width=0.32\linewidth,frame]{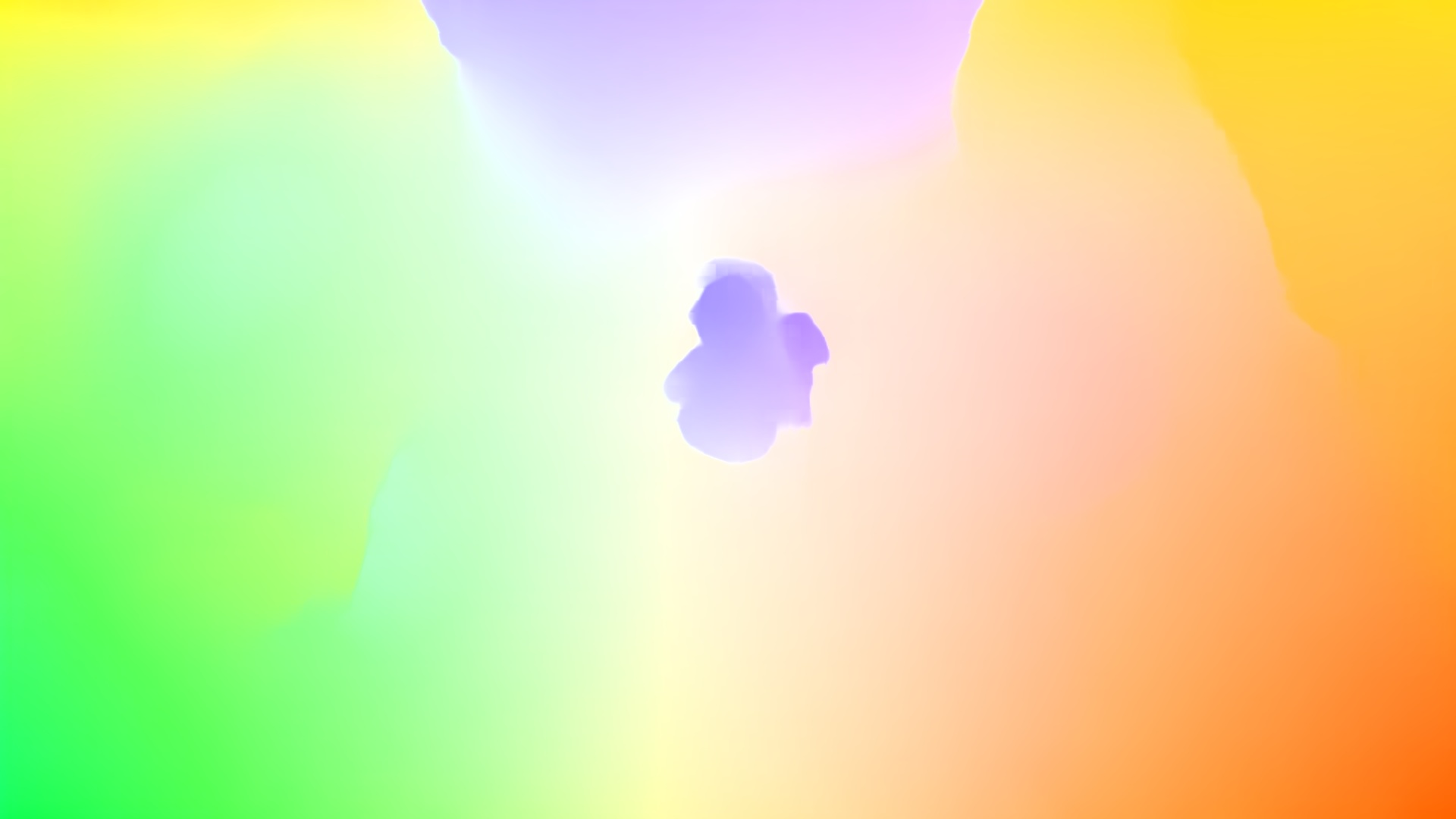}
    \put (2,52) {\small \textcolor{purple}{\textbf{\texttt{EPE: 0.146 }}}}
    \end{overpic} & 
    \begin{overpic}[width=0.32\linewidth,frame]{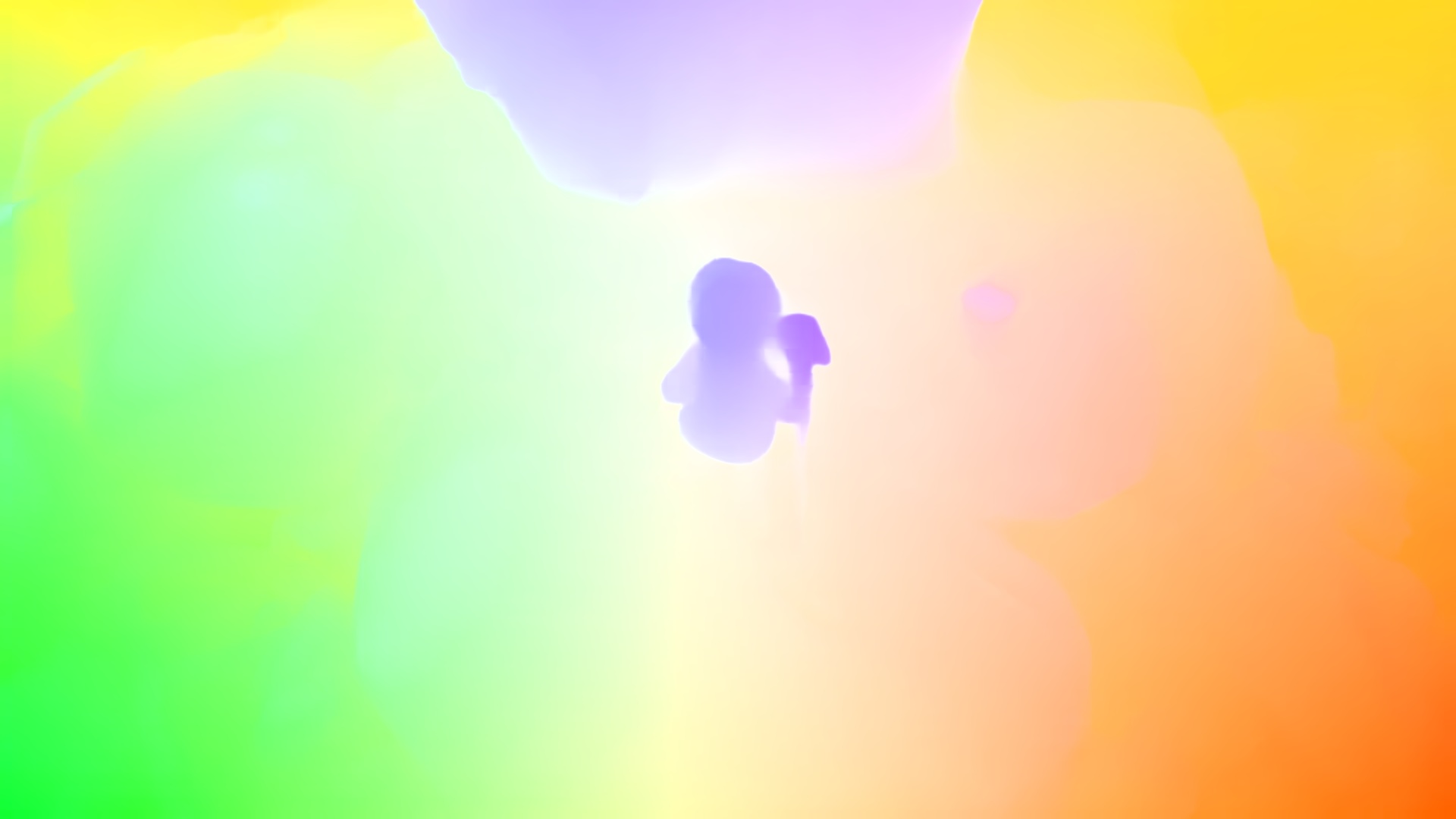}
    \put (2,52) {\small \textcolor{purple}{\textbf{\texttt{EPE: 0.108 
    }}}}
    \end{overpic} \\
    \begin{overpic}[width=0.32\linewidth,frame]{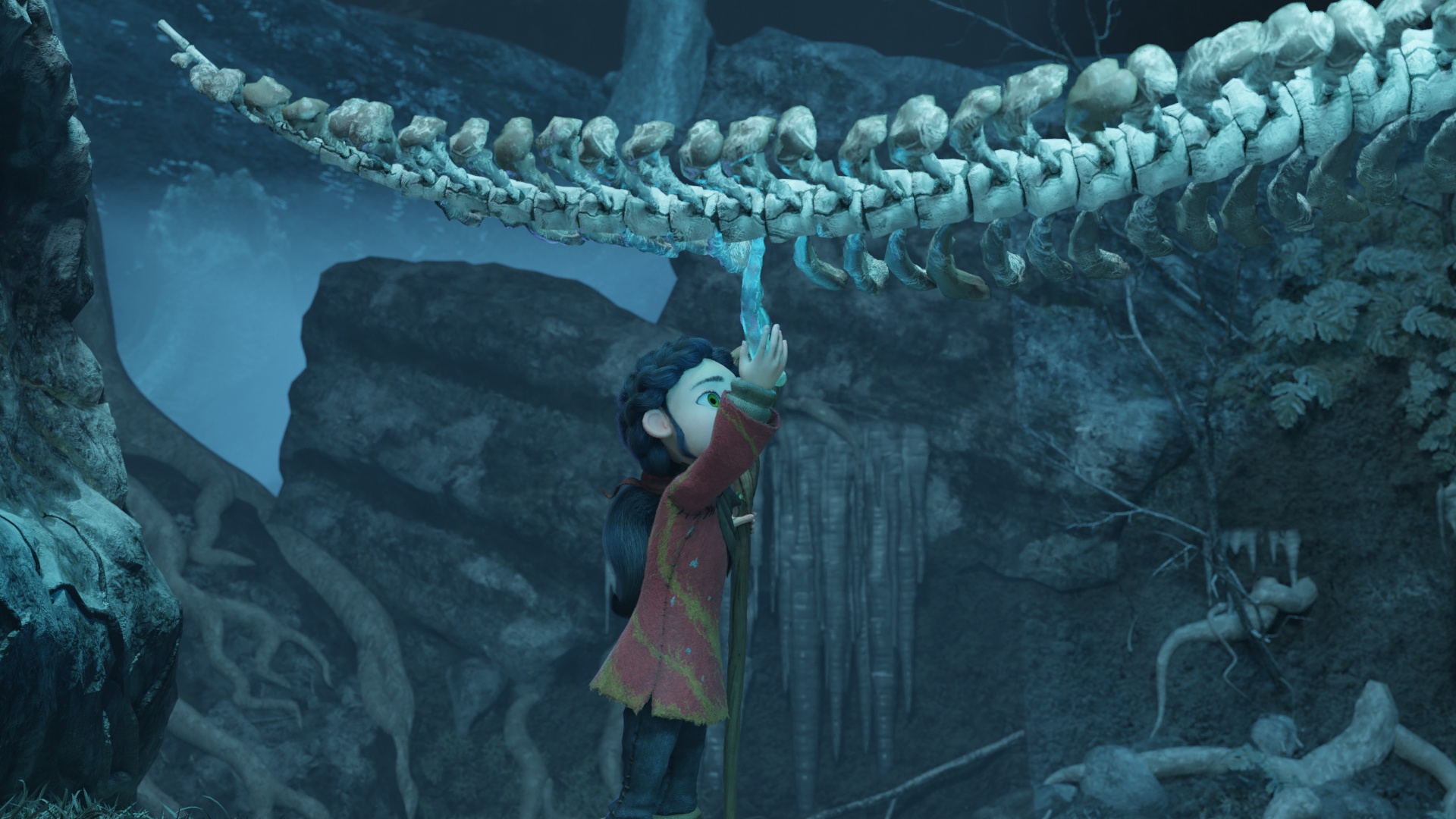}
    \end{overpic} & 
    \begin{overpic}[width=0.32\linewidth,frame]{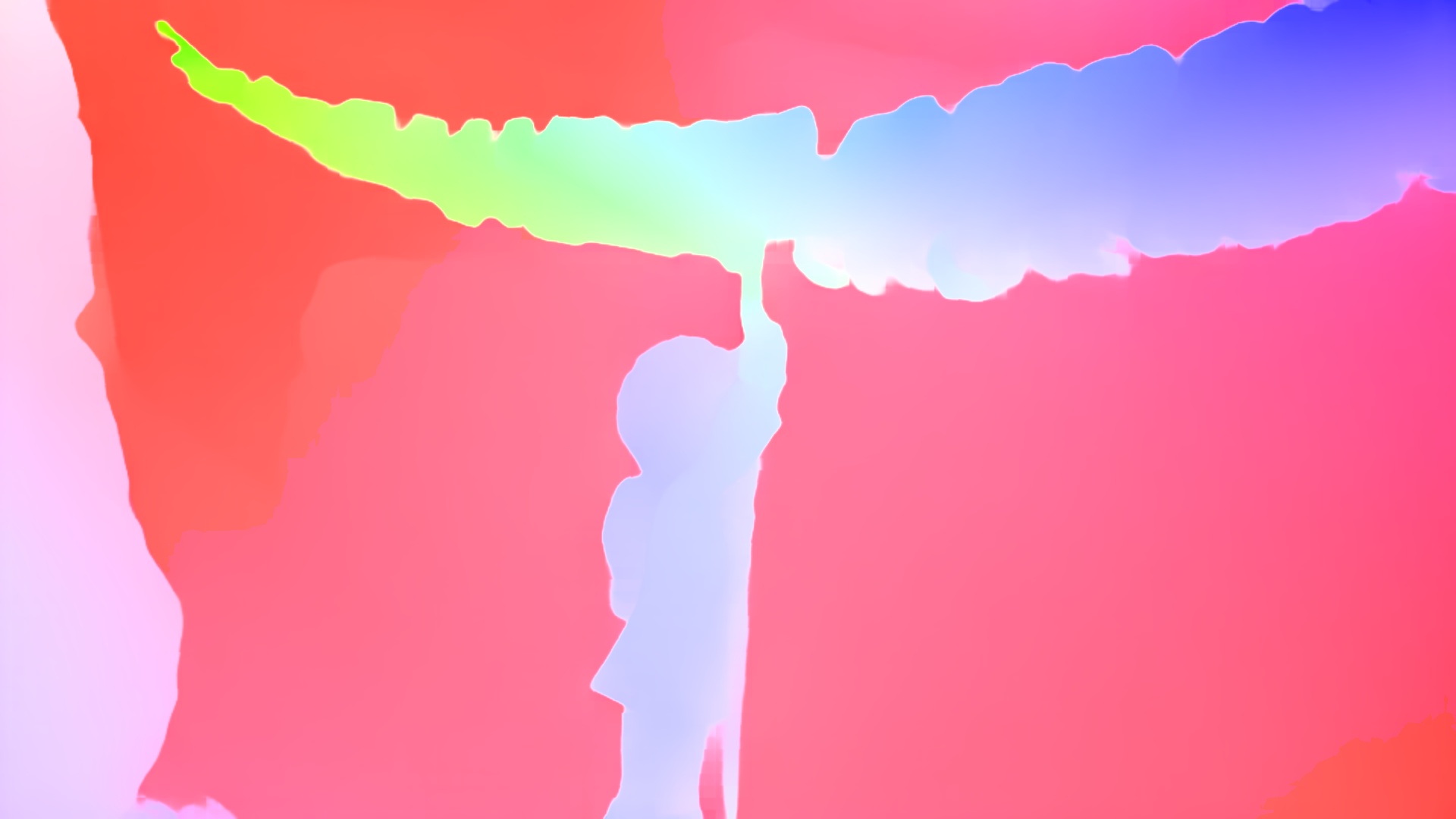}
    \put (2,52) {\small \textcolor{purple}{\textbf{\texttt{EPE: 0.292 }}}}
    \end{overpic} & 
    \begin{overpic}[width=0.32\linewidth,frame]{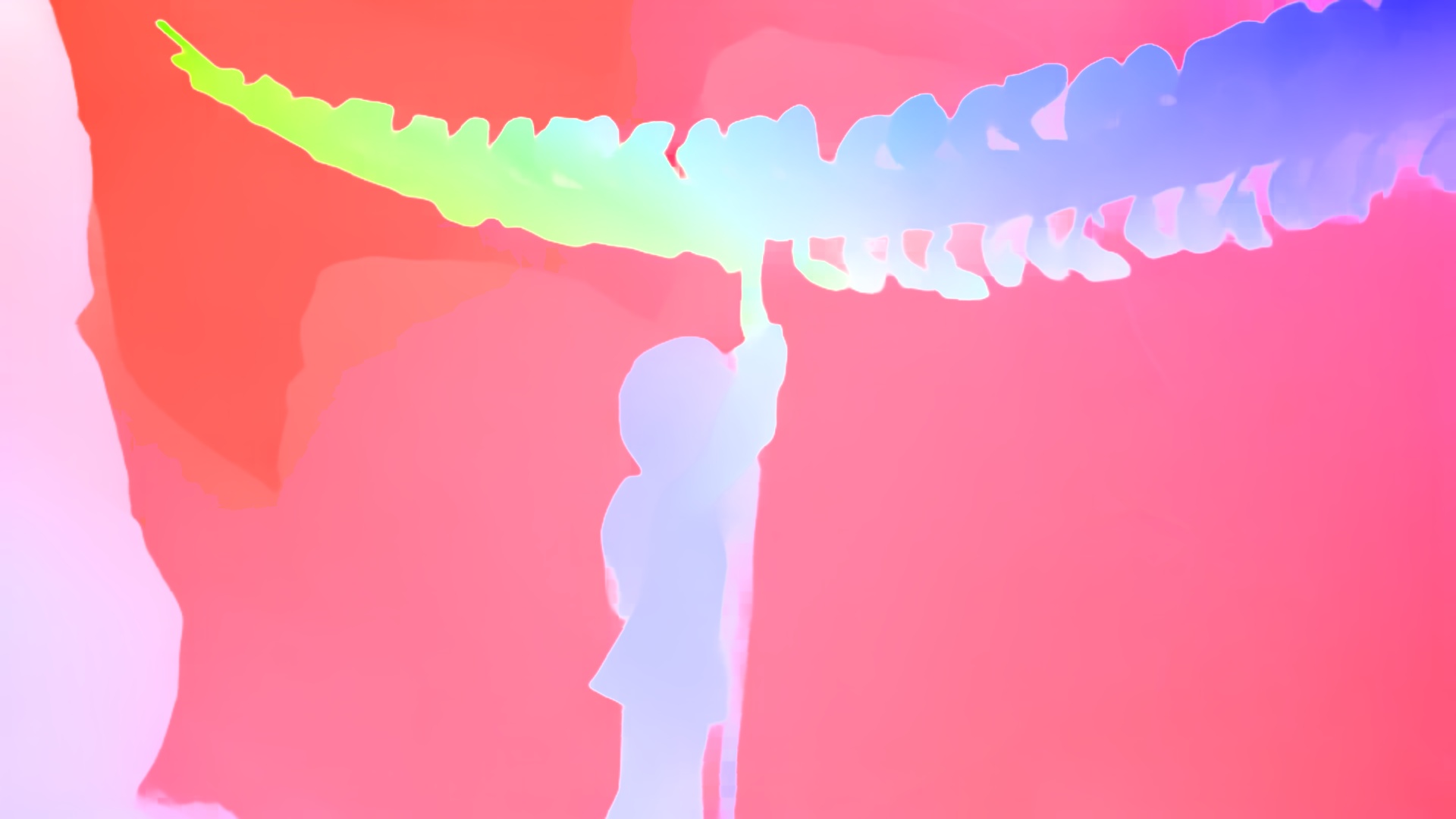}
    \put (2,52) {\small \textcolor{purple}{\textbf{\texttt{EPE: 0.241 
    }}}}
    \end{overpic} \\

    \begin{overpic}[width=0.32\linewidth,frame]{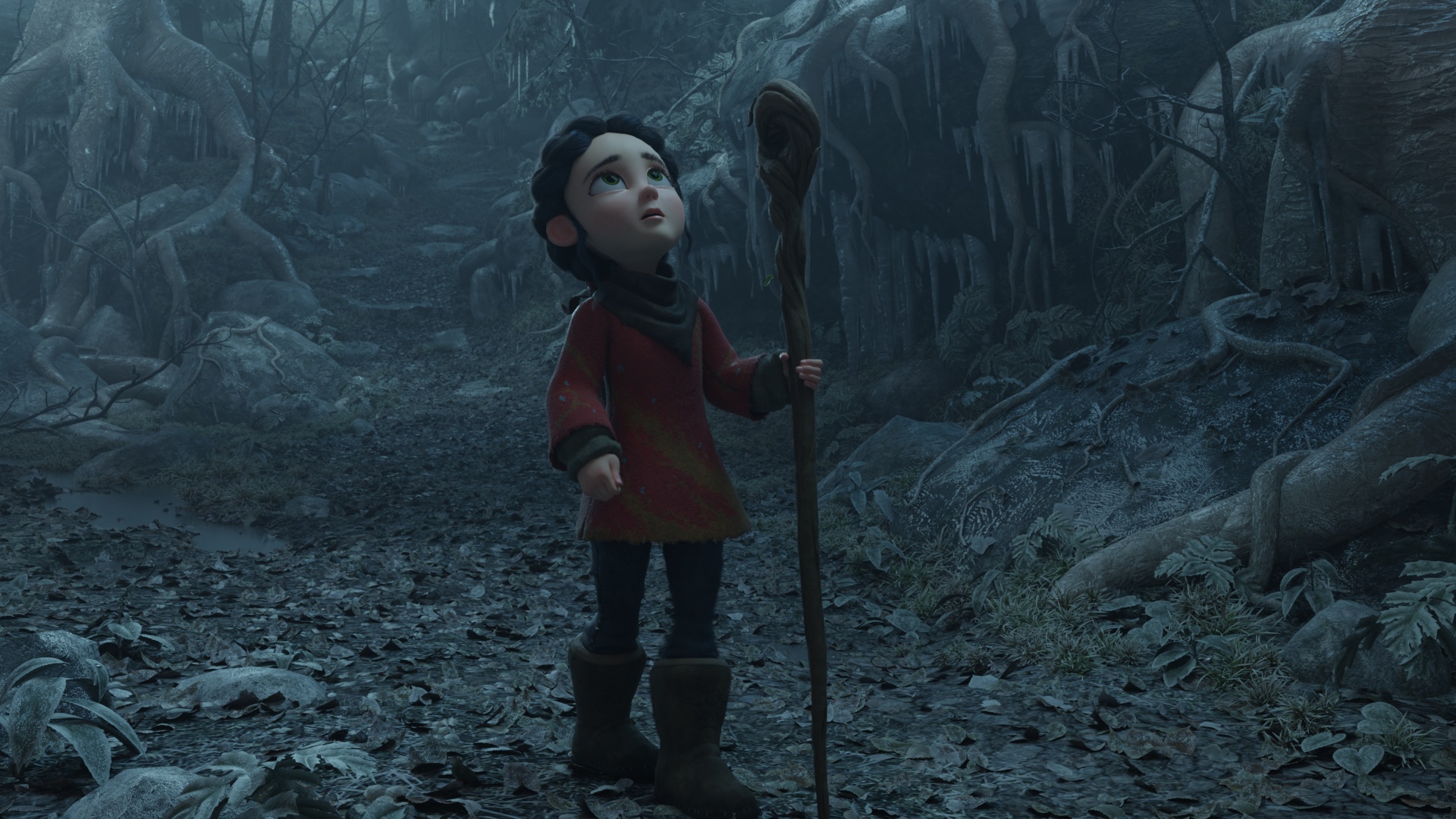}
    \end{overpic} & 
    \begin{overpic}[width=0.32\linewidth,frame]{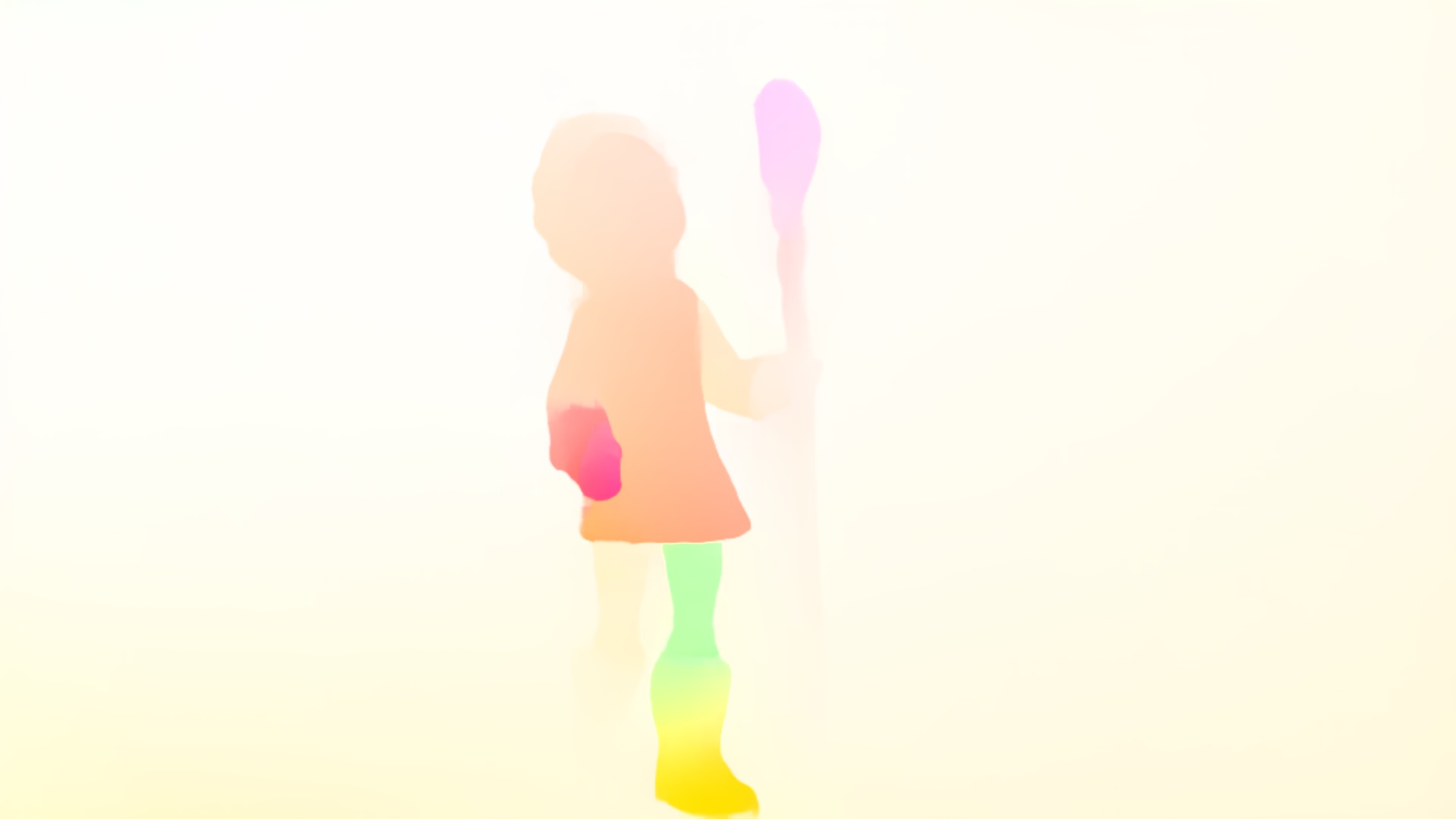}
    \put (2,52) {\small \textcolor{purple}{\textbf{\texttt{EPE: 0.112 }}}}
    \end{overpic} & 
    \begin{overpic}[width=0.32\linewidth,frame]{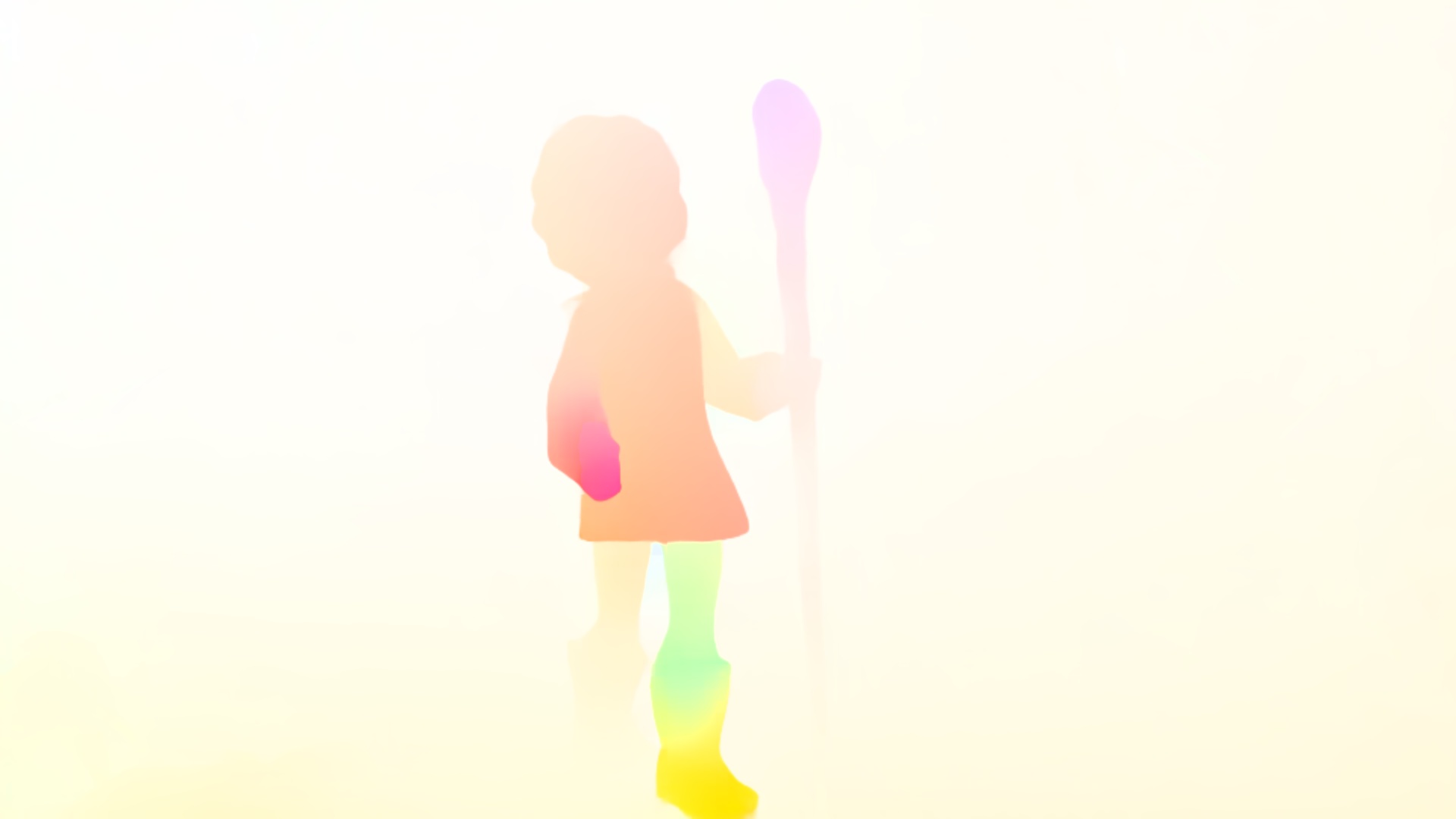}
    \put (2,52) {\small \textcolor{purple}{\textbf{\texttt{EPE: 0.097 
    }}}}
    \end{overpic} \\

    \begin{overpic}[width=0.32\linewidth,frame]{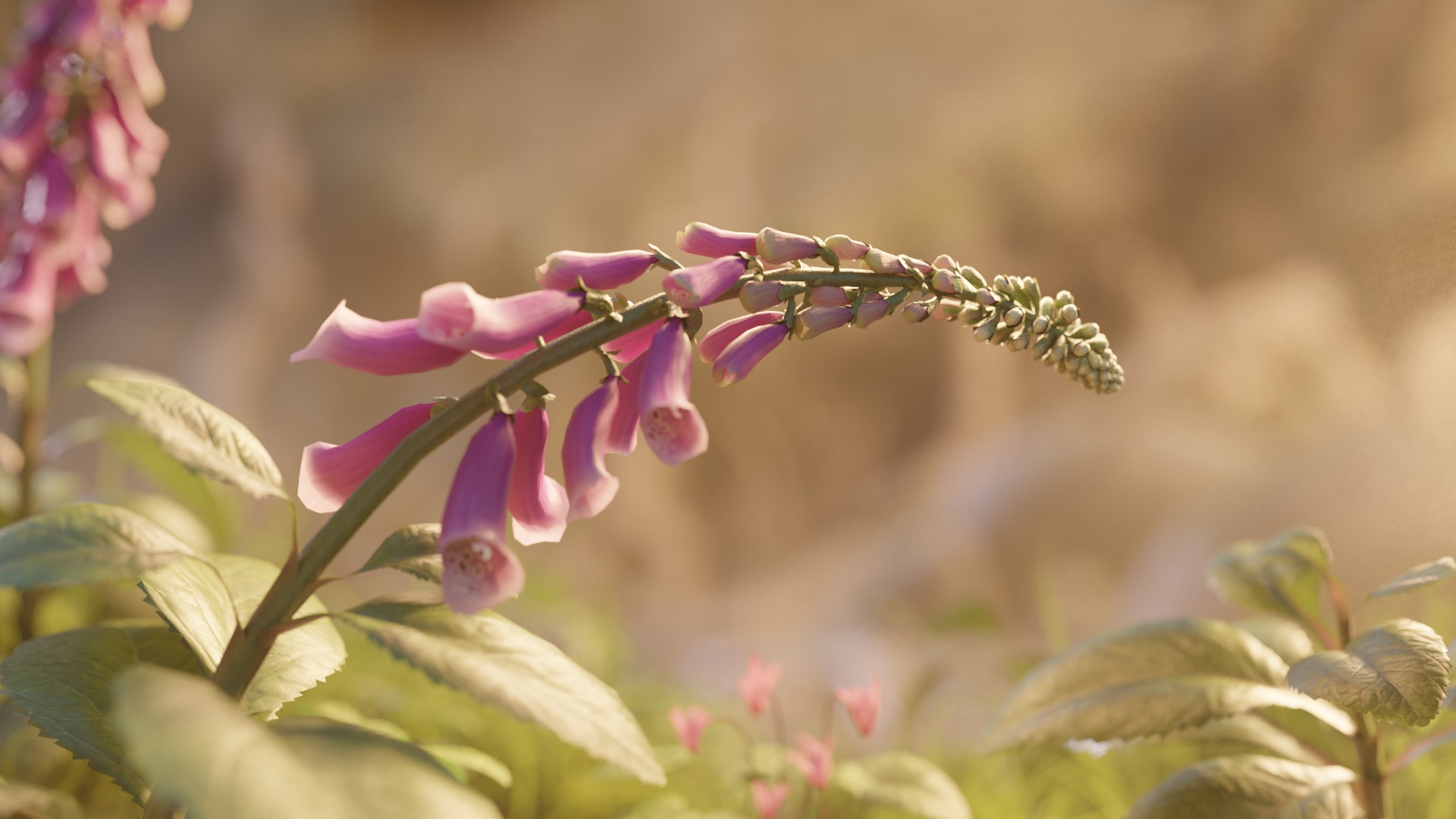}
    \end{overpic} & 
    \begin{overpic}[width=0.32\linewidth,frame]{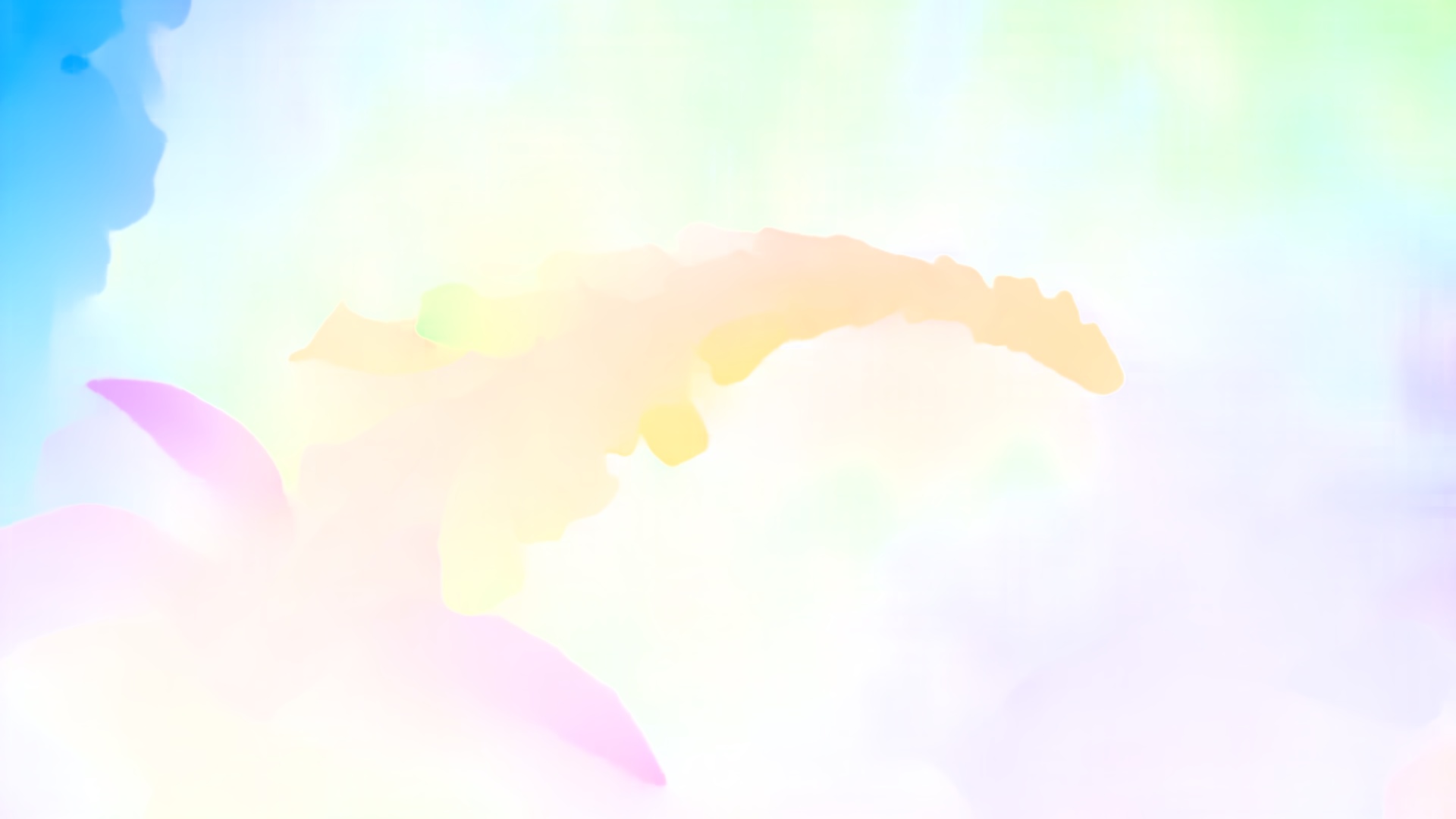}
    \put (2,52) {\small \textcolor{purple}{\textbf{\texttt{EPE: 0.327 }}}}
    \end{overpic} & 
    \begin{overpic}[width=0.32\linewidth,frame]{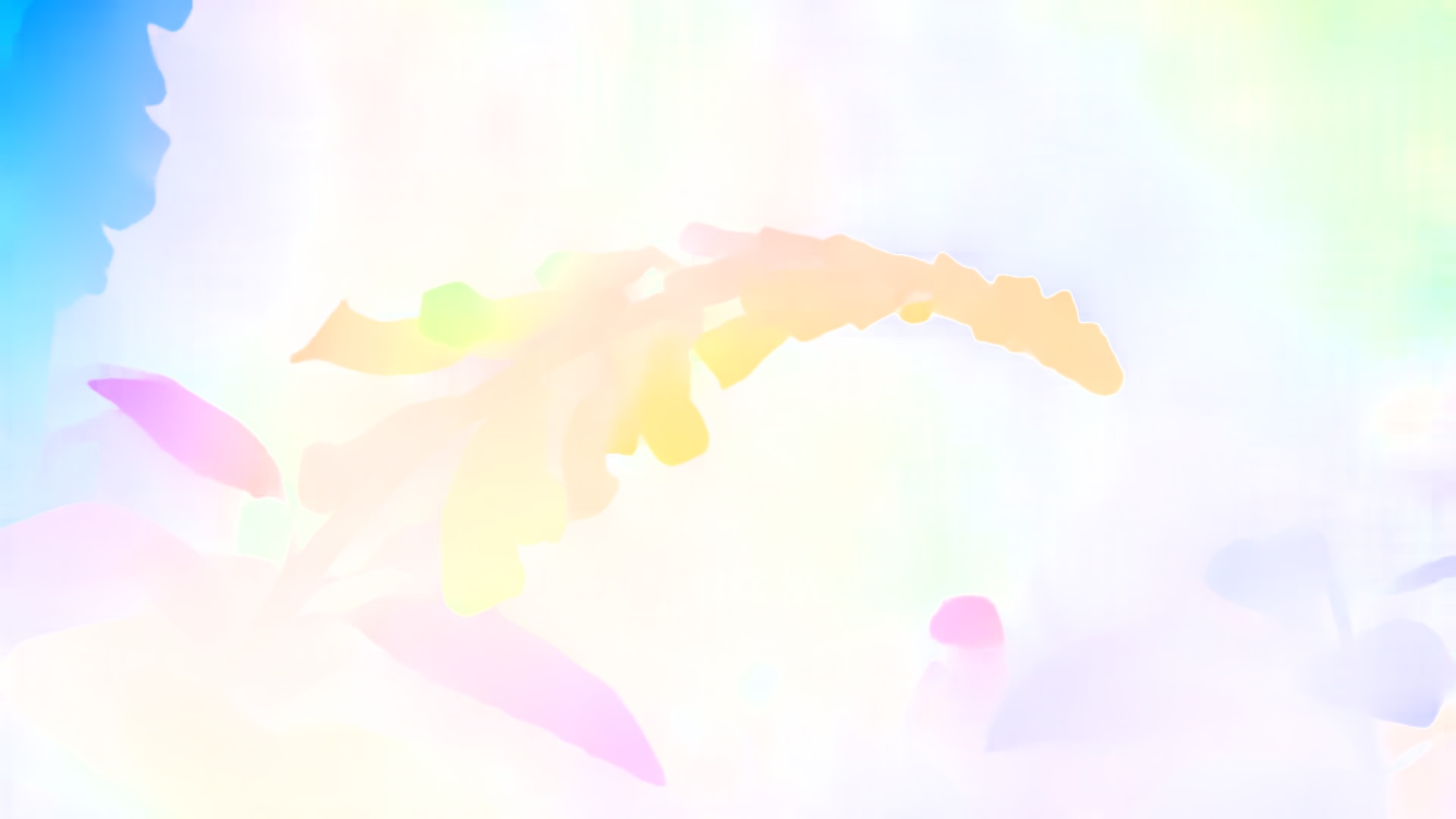}
    \put (2,52) {\small \textcolor{purple}{\textbf{\texttt{EPE: 0.240 
    }}}}
    \end{overpic} \vspace{-0.2cm}\\

    \footnotesize $\mathbf{I}_0$ & \footnotesize SEA-RAFT (S) & \bf\footnotesize \net{} (T) \vspace{-0.2cm}\\
    \end{tabular}

    \caption{\textbf{Qualitative results on Spring -- ``C$\rightarrow$T" schedule after TartanAir pretraining.} From left to right: first frame, flow by SEA-RAFT (S) and FlowSeek (T).}
    \label{fig:spring_gen}
\end{figure*}

\begin{table*}[b]
  \centering
  \resizebox{\linewidth}{!}{
  \begin{tabular}{l r rrrr r rrrr r rrrr r rrrr}
    \toprule
    \multirow{2}{*}{Method} && \multicolumn{4}{c}{All} && \multicolumn{4}{c}{Transparent} && \multicolumn{4}{c}{Reflective} && \multicolumn{4}{c}{Diffuse} \\
    \cmidrule{2-21}
    && EPE$\downarrow$ & 1px$\downarrow$ & 3px$\downarrow$ & 5px$\downarrow$ 
    && EPE$\downarrow$ & 1px$\downarrow$ & 3px$\downarrow$ & 5px$\downarrow$ 
    && EPE$\downarrow$ & 1px$\downarrow$ & 3px$\downarrow$ & 5px$\downarrow$
    && EPE$\downarrow$ & 1px$\downarrow$ & 3px$\downarrow$ & 5px$\downarrow$\\
    \midrule
    FlowNet-C && 9.71 & 89.07 & 61.51 & 43.93 && 11.08 & 89.23 & 62.43 & 45.05 && 6.38 & 88.25 & 58.36 & 40.03 && 8.53 & 89.08 & 54.13 & 35.35 \\
    FlowNet2 && 10.07 & 77.56 & 54.22 & 42.13 &&
                11.46 & 78.20 & 56.15 & 44.38 &&
                6.70 & 75.39 & 46.69 & 33.33 &&
                7.66 & 72.44 & 43.21 & 29.41 \\
    PWC-Net && 9.49 & 74.93 & 50.47 & 39.05
                && \trd 10.90 & 76.47 & 52.59 & 41.43
                && 5.99 & 69.84 & 42.99 & 29.82
                && 6.91 & 61.85 & 34.77 & 25.30 \\
    GMA && 9.77 & 72.46 & 46.93 & 36.97
            && 12.01 & 75.48 & 50.07 & 40.24
            && 4.48 & 60.85 & 35.42 & 24.26
            && \snd 2.26 & 54.20 & 25.56 & 17.98 \\
    SKFlow && 9.86 & 72.02 & 47.44 & 36.88
            && 12.00 & \trd 74.90 & 50.84 & 40.14
            && 4.78 & 60.89 & 35.21 & 24.40
            && 3.23 & 54.90 & \trd 23.23 & 17.18 \\
    CRAFT && 10.36 & 72.34 & 47.54 & 37.00
            && 12.65 & \snd 74.75 & 50.96 & 40.47
            && 4.65 & 64.12 & 35.10 & 23.06
            && 3.30 & \trd 53.08 & 23.95 & 18.89\\
    GMFlow && \trd 9.09 & 81.99 & 51.79 & 37.75
            && 10.93 & 83.01 & 53.87 & 40.06
            && 5.20 & 80.02 & 44.73 & 28.64
            && 5.01 & 66.91 & 35.13 & 24.79 \\
    GMFlow+ && 9.46 & 82.71 & 53.14 & 39.70
            && 11.31 & 83.21 & 54.91 & 42.10
            && 6.04 & 81.61 & 46.57 & 29.95
            && 5.71 & 75.97 & 41.43 & 27.85\\
    FlowFormer && 10.20 & 73.59 & 48.97 & 38.56
            && 12.51 & 76.91 & 52.56 & 42.27
            && 5.00 & 61.03 & 36.18 & 24.76
            && \fst 2.17 & \snd 52.89 & \snd 22.90 & \snd 14.12 \\ 
    RAFT && 9.38 & 71.98 & 46.46 & 36.15
            && 11.31 & \fst 74.65 & 49.69 & 39.34
            && 5.57 & 61.53 & 35.73 & 24.57
            && \trd 2.62 & 56.72 & \fst 19.44 & \fst 14.05 \\
    \midrule 
    SEA-RAFT (S) && 10.05 & 71.48 & 46.90 & 36.32 && 12.07 & 75.80 & 51.14 & 40.40 && 3.59 & 49.67 & 24.79 & 15.48 && 4.56 & 55.00 & 36.77 & 22.60 \\
    SEA-RAFT (M) && 10.17 & 69.73 & 45.94 & 34.78 && 12.33 & 75.43 & 50.93 & 39.20 && 3.17 & 40.17 & 19.17 & 11.75 && 3.75 & 54.90 & 40.83 & 24.17 \\
    SEA-RAFT (L) && 10.99 & \trd 69.46 & 45.59 & 34.78 && 13.33 & 75.53 & 50.55 & 39.13 && 3.17 & \trd 37.99 & 19.12 & 11.99 && 3.69 & 53.75 & 38.96 & 25.52 \\
    \bf \net{} (T) && \trd 9.09  & 70.82  & 43.74 & 32.36 && 11.31  & 76.30  & 49.53 & 36.96 && \snd 2.31 & 42.12  & \trd 13.49 & \trd 8.44 && 2.70 & 59.06  & 30.42 & 20.62 \\
    \bf \net{} (S) && 9.16 & 69.99 & \trd 43.67 & \snd 31.90 && 11.50 & 75.96 & \trd 49.41 & \snd 36.62 && \fst 1.79 & 38.44 & \snd 13.46 & \fst 7.38 && 2.69 & 59.69 & 32.71 & 20.10 \\
    \bf \net{} (M) && \fst 8.30 & \fst 68.85 & \snd 41.81 & \trd 32.09 && \snd 10.17 & 75.38 & \snd 47.58 & \trd 36.74 && 2.74 & \snd 35.07 & \fst 12.34 & \snd 8.27 && 2.81 & \fst 51.15 & 23.12 & \trd 16.98 \\
    \bf \net{} (L) && \fst 8.30 & \snd 68.98 & \fst 41.49 & \fst 31.64 && \fst 10.15 & 75.56 & \fst 46.78 & \fst 36.02 && \trd 2.39 & \fst 34.49 & 14.09 & 8.90 && 3.20 & 54.69 & 27.50 & 20.31 \\
    
    \bottomrule
  \end{tabular}
  }
  \vspace{-3mm}
  \caption{\textbf{Zero-shot generalization -- LayeredFlow (train) first layer evaluation.} Images are down-sampled by 8.}\vspace{-0.3cm}
  \label{tab:optical_flow_first_layer}
\end{table*}

\end{document}